# Multirobot rendezvous with visibility sensors in nonconvex environments

Anurag Ganguli Jorge Cortés Francesco Bullo

***Abstract*— This paper presents a coordination algorithm for mobile autonomous robots. Relying upon distributed sensing the robots achieve rendezvous, that is, they move to a common location. Each robot is a point mass moving in a nonconvex environment according to an omnidirectional kinematic model. Each robot is equipped with line-of-sight limited-range sensors, i.e., a robot can measure the relative position of any object (robots or environment boundary) if and only if the object is within a given distance and there are no obstacles in-between. The algorithm is designed using the notions of robust visibility, connectivity-preserving constraint sets, and proximity graphs. Simulations illustrate the theoretical results on the correctness of the proposed algorithm, and its performance in asynchronous setups and with sensor measurement and control errors.**

## I. INTRODUCTION

Multi-agent robotic systems have been receiving increasing attention in recent times. This is due in no small part to the remarkable advances made in recent years in the development of small, agile, relatively inexpensive sensor nodes with mobile and networking capabilities. These systems are ultimately intended for a wide variety of purposes such as search and rescue, exploration, environmental monitoring, location-aware computing, and the maintaining of structures. In this paper we design algorithms to steer a group of robots to a common location, or rendezvous, in a nonconvex environment.

The rendezvous problem is a fundamental motion coordination problem for collections of robots. In its essence, it is the most basic formation control problem and can be used as a building block for more sophisticated behaviors. It is related to the classic consensus problem in distributed algorithms. This problem is, here, tackled in a fully distributed manner, i.e., our robots do not have any global knowledge about the position of all other robots, do not have any global knowledge about the environment, and do not share a common reference frame. The information that is available to an individual robot is only what is provided by a local "visibility sensor." In other words, a robot can measure the relative position of a second robot if and only if the robots are visible to each other in the nonconvex environment and lie within a given distance of each other.

The literature on multirobot systems is very extensive. Examples include the survey in [1] and the special issue [2] of the IEEE Transaction on Robotics and Automation. Our multi-robot model is inspired by the literature on networks of mobile interacting robots: an early contribution is the model proposed in [3] consisting of a group of identical "distributed anonymous mobile robots" characterized as follows. Each robot completes repeatedly a cycle of operations: it senses the relative position of all other robots, elaborates this information, and moves. In this early work, the robots share a common clock. A related model is presented in [4], where the robots evolve asynchronously, have limited visibility, and share a common reference frame. For these types of multirobot systems, the "multi-agent rendezvous" problem and the first "circumcenter algorithm" have been introduced in [5]. This circumcenter algorithm has been extended to various asynchronous strategies in [6], [4], where rendezvous is referred to as the "gathering" problem. The circumcenter algorithm has been extended beyond planar problems to arbitrary dimensions in [7], where its robustness properties are also characterized. *None of these previous works considers the problem in nonconvex environments with line-of-sight visibility sensors.*

We conclude the literature review by mentioning that formation control and rendezvous problems have been widely investigated with different assumptions on the inter-agent sensing. For example, a control law for groups with *time-dependent* sensing topology is proposed in [8]; this and similar works, however, depend upon a critical assumption of connectivity of the inter-agent sensing graph. This assumption is imposed without a model for when two robots can detect and measure each other's relative position. In this paper, we consider *position-dependent* graphs and, extending to visibility sensors a key idea in [5], we show how to constrain the robots' motion to maintain connectivity of the inter-agent sensing graph.

Next, we describe the essential details of our model. We consider a group of robotic agents moving in a nonconvex environment without holes. We assume each robot is modeled as a point mass. We assume that each robot is equipped with an *omnidirectional limited-range visibility sensor*; the nomenclature is adopted from [9, Section 11.5]. Such a sensor is a device (or combination of devices) that determines within its line of sight and its sensing range the following quantities: (i) the relative position of other robots, and (ii) the relative position of the boundary of environment. By omnidirectional we mean that the field-of-vision for the sensor is $2\pi$ radians. Examples of visibility sensors are scanning laser range finders

Submitted to http://www.arxiv.org on November 5, 2006. An early version of this work appeared in the IEEE Conf. on Decision and Control and European Control Conference, in Seville, Spain, 2005.

Anurag Ganguli is with the Coordinated Science Laboratory, University of Illinois at Urbana-Champaign, and with the Center for Control, Dynamical Systems and Computation, University of California, Santa Barbara, CA 93106, USA, aganguli@uiuc.edu

Jorge Cortés is with the Department of Applied Mathematics and Statistics, University of California, Santa Cruz, CA 95064, USA, jcortes@ucsc.edu

Francesco Bullo is with the Center for Control, Dynamical Systems and Computation, University of California, Santa Barbara, CA 93106, USA, bullo@engineering.ucsb.edu

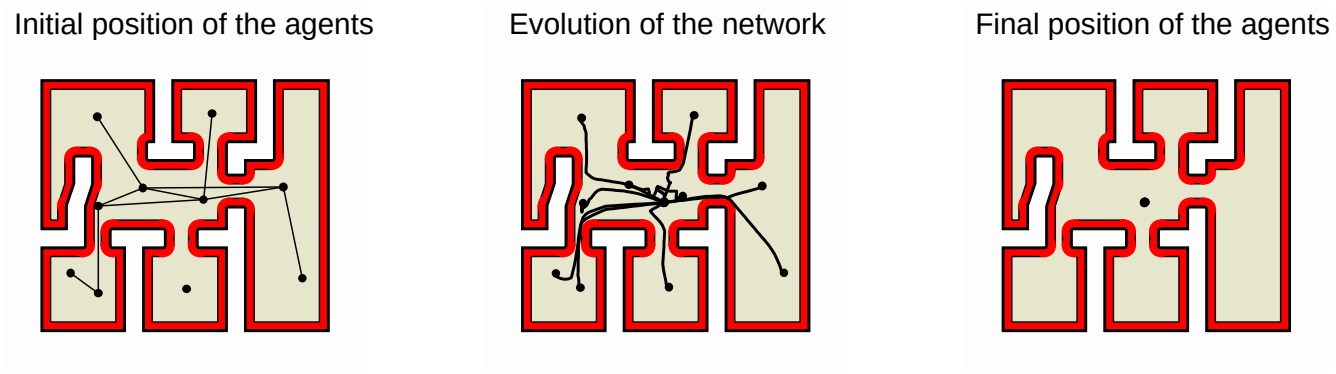

Fig. 1. Execution of the `Perimeter Minimizing Algorithm` described in Section V-A on a group of robots distributed in a polygon, $Q$, shaped like a typical floor plan. The graph shown in the left-most figure is the $r$-range visibility graph $\mathcal{G}_{r\text{-vis},Q_\epsilon}$ (see Section II).

with accurate distance measurements at high angular density,[1] time-of-flight range cameras,[2] and optical depth sensors based on structured light systems, e.g., see [10]. The range data obtained from the sensors can be processed to obtain a geometric representation of the area visible from a robot, e.g., see [11]. We do not directly address in this work issues related to feature extraction from range data. We assume that the algorithm regulating the robots' motion is memoryless, i.e., we consider static feedback laws. Given this model, the goal is to design a discrete-time algorithm which ensures that the robots converge to a common location within the environment. See Figure 1 for a graphical description of a simulation.

This paper's main contribution is a novel provably correct algorithm that achieves rendezvous in a nonconvex planar environment among robots with omnidirectional range-limited visibility sensors. Rendezvous is achieved among all robots if the inter-agent sensing graph is connected at the initial time or becomes connected at any time during the evolution.

The technical approach contains a number of novel contributions and proceeds as follows. First, we review a few useful geometric notions [12], such as robust visibility [13] and proximity graphs [14], and introduce various novel visibility graphs. Second, to maintain connectivity during the system evolution, we design novel constraint sets that (i) ensure that the visibility between two robots is preserved, and (ii) are upper semicontinuous or closed maps of the robots' positions. Third, based on a discussion on visibility graphs, we define a new proximity graph, called the locally-cliqueless visibility graph, which contains fewer edges than the visibility graph, and has the same connected components. This construction is useful in the connectivity maintenance problem. Fourth, we provide a careful analysis of the algorithm we propose. As novel Lyapunov function, we consider the perimeter of the relative convex hull of the robot positions. The main theorem is proved via our recent version of the LaSalle Invariance Principle for set-valued maps [7]. Fifth and final, extensive simulations validate our theoretical results and establish the convergence of our algorithm under more realistic assumptions than the ones adopted in the theoretical analysis: our algorithm performance is still adequate assuming asynchronous agent operation, noise errors in sensing and control, or finite-size disk robots.

[1] E.g., the Hokuyo URG-04LX, see `http://www.hokuyo-aut.jp`, and the Sick S2000, see `http://www.sick.com`.

[2] E.g., the SwissRanger SR-3000, see `http://www.swissranger.ch`.

The paper is organized as follows. Section II contains some useful geometric notions. In Section III we model robots with visibility sensors and we introduce the rendezvous and connectivity maintenance problems. In Section IV we introduce constraint sets and locally-cliqueless visibility graphs. In Section V we propose the `Perimeter Minimizing Algorithm` for the rendezvous problem. Numerical simulations are presented in Section VI. Additional analysis results and all proofs are presented in Appendices I-III.

## II. Geometric notions

In this section we introduce some useful geometric notions. We begin by reviewing some standard notation. Let $\mathbb{Z}_{\geq 0}$, $\mathbb{R}$, $\mathbb{R}_{\geq 0}$, and $\mathbb{R}_{>0}$ denote the sets of nonnegative integer, real, nonnegative real, and positive real numbers, respectively. For $p \in \mathbb{R}^2$ and $r \in \mathbb{R}_{>0}$, let $B(p, r)$ denote the *closed ball* centered at $p$ of radius $r$. Given a bounded set $X \subset \mathbb{R}^2$, let $\mathrm{co}(X)$ denote the convex hull of $X$, and let $\mathrm{CC}(X)$ denote the *circumcenter* of $X$, i.e., the center of the smallest-radius circle enclosing $X$. For $p, q \in \mathbb{R}^2$, let $]p, q[= \{\lambda p + (1-\lambda)q \mid 0 < \lambda < 1\}$ and $[p, q] = \{\lambda p + (1-\lambda)q \mid 0 \leq \lambda \leq 1\}$ denote the *open* and *closed segment* with extreme points $p$ and $q$, respectively. Let $|X|$ denote the cardinality of a finite set $X$ in $\mathbb{R}^2$. Given a set of points $X \subset \mathbb{R}^2$, and another point $p \in \mathbb{R}^2$, let $\mathrm{dist}(p, X)$ denote the Euclidean distance of $p$ to the set $X$. The diameter $\mathrm{diam}(X)$ of a compact set $X$ is the maximum distance between any two points in $X$.

Now, let us turn our attention to the type of environments we are interested in. Given any compact and connected subset $Q$ of $\mathbb{R}^2$, let $\partial Q$ denote its boundary. A point $q$ of $\partial Q$ is *strictly concave* if for all $\epsilon > 0$ there exists $q_1$ and $q_2$ in $B(q, \epsilon) \cap \partial Q$ such that the open interval $]q_1, q_2[$ is outside $Q$. A *strict concavity* of $\partial Q$ is either an isolated strictly concave point or a connected set of strictly concave points. According to this definition, a strict concavity is either an isolated point (e.g., points $r_1$ and $r_2$ in Figure 2) or an arc (e.g., arc $a_1$ in Figure 2). Also, any strictly concave point belongs to exactly one strict concavity.

**Definition II.1 (Allowable environments)** *A set $Q \subset \mathbb{R}^2$ is* allowable *if*

(i) *$Q$ is compact and simply connected;*
(ii) *$\partial Q$ is continuously differentiable except on a finite number of points;*
(iii) *$\partial Q$ has a finite number of strict concavities.*

Recall that, roughly speaking, a set is simply connected if it is connected and it contains no hole. A particular case of the environment described above is a polygonal environment, the concavities being the reflex vertices[3] of the environment.

At almost all strictly concave points $v$, one can define the tangent to $\partial Q$. (Here, the wording "almost all" points means all except for a finite number.) At all such points $v$, the *internal tangent half-plane* $H_Q(v)$ is the half-plane whose boundary is tangent to $\partial Q$ at $v$ and whose interior does not contain any points of the concavity; see Figure 2.

[3] A vertex of a polygon is reflex if its interior angle is strictly greater than $\pi$.

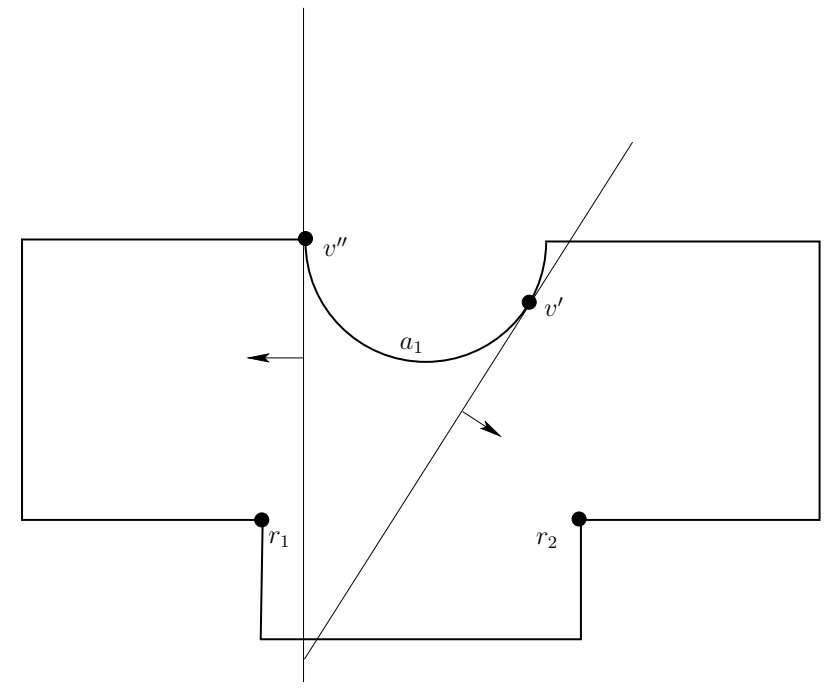

Fig. 2. An allowable environment $Q$: the closed arc $a_1$ and the isolated points $r_1, r_2$ are strict concavities. $v'$ is a point on $a_1$ where the slope of $\partial Q$ is defined. $H_Q(v')$ is the half-plane with the tangent to $\partial Q$ at $v'$ as the boundary and the interior in the direction of the arrow. $v''$ is a point on $a_1$ where the slope of $\partial Q$ is not well-defined. In this case, we define the tangent to be the one shown in the plot. $H_Q(v'')$ is the half-plane with the tangent to $\partial Q$ at $v''$ as the boundary and the interior in the direction of the arrow.

A point $q \in Q$ is *visible* from $p \in Q$ if $[p, q] \subset Q$. The *visibility set* $\mathcal{V}(p) \subset Q$ from $p \in Q$ is the set of points in $Q$ visible from $p$. This notion can be extended as follows (see [13]):

**Definition II.2 (Robust visibility)** *Take $\epsilon > 0$ and $Q \subset \mathbb{R}^2$.*

(i) *The point $q \in Q$ is* $\epsilon$-robustly visible *from the point $p \in Q$ if $\cup_{q' \in [p,q]} B(q', \epsilon) \subset Q$.*
(ii) *The* $\epsilon$-robust visibility set $\mathcal{V}(p, \epsilon) \subset Q$ *from $p \in Q$ is the set of points in $Q$ that are $\epsilon$-robustly visible from $p$.*
(iii) *The* $\epsilon$-contraction $Q_\epsilon$ *of the set $Q$ is the set $\{p \in Q \mid ||p - q|| \geq \epsilon$ for all $q \in \partial Q\}$.*

These notions are illustrated in Figure 3. Loosely speaking, two points $p, q$ are mutually $0$-robustly visible if and only if they are mutually visible. We present the following properties without proof in the interest of brevity.

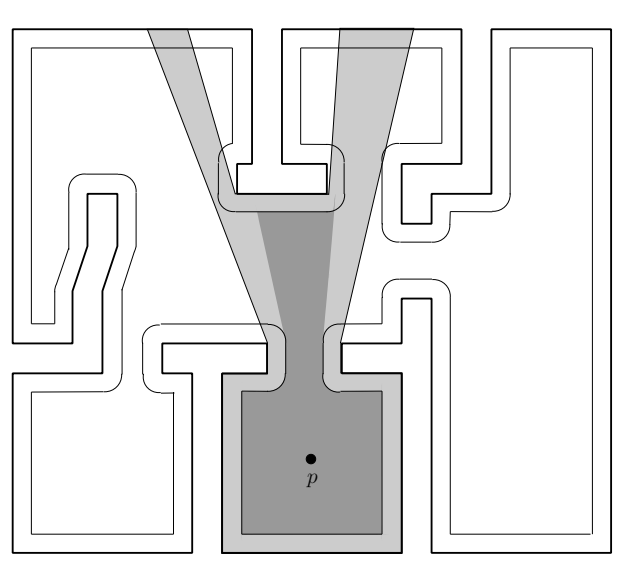

Fig. 3. Robust visibility notions. $Q$ is the outer polygonal environment; the $\epsilon$-contraction $Q_\epsilon$ is the region with the curved boundary and containing the point $p$; the visibility set $\mathcal{V}(p)$ is the region shaded in light gray; the $\epsilon$-robust visibility set $\mathcal{V}(p, \epsilon)$ is the region shaded in darker gray. Note that the isolated concavities of $Q$ give rise to strictly concave arcs in $Q_\epsilon$.

**Lemma II.3** *Given an allowable environment $Q$ and $\epsilon > 0$, the following statements hold:*

(i) *$q \in Q$ is $\epsilon$-robustly visible from $p \in Q$ if and only if $[p, q] \subset Q_\epsilon$;*
(ii) *if $\epsilon$ is sufficiently small, then $Q_\epsilon$ is allowable;*
(iii) *all strict concavities of $\partial Q_\epsilon$ have non-zero length and are continuously differentiable.*

**Remarks II.4** (i) In light of Lemma II.3(ii), in what follows we assume that $\epsilon$ is small enough for $Q_\epsilon$ to be connected and therefore allowable.
(ii) Robust visibility is a useful concept in many practically meaningful ways. For example, according to this notion, points are visible only if they are at least at a distance $\epsilon$ from the boundary. This is useful when an object is arbitrarily close to the boundary and is indistinguishable from the boundary itself. Additionally, the parameter $\epsilon$ might be thought of as a measure of the physical size of the robot. Thus confining the robots to the $\epsilon$-robust visibility set guarantees free movement of the robot in the environment. Indeed, the notion of $\epsilon$-contraction is related to the classical work on motion planning in [15], see also [9]. □

We now define some graphs which will be useful in describing the interactions between robots.

**Definition II.5 (Proximity graphs)** *A* proximity graph *is a graph whose nodes are a set of points $\mathcal{P} = \{p_1, \dots, p_n\}$ and whose edges are a function of $\mathcal{P}$. Given $\mathcal{P} \subset Q$, $\epsilon > 0$ and $r > 0$, define:*

(i) *The* visibility graph $\mathcal{G}_{\text{vis},Q}$ *at $\mathcal{P}$ is the graph with node set $\mathcal{P}$ and with edge set $\mathcal{E}_{\text{vis},Q}(\mathcal{P})$ defined by: $(p_i, p_j) \in \mathcal{E}_{\text{vis},Q}(\mathcal{P})$ if and only if $[p_i, p_j] \subset Q$.*
(ii) *The* $\epsilon$-robust visibility graph $\mathcal{G}_{\text{vis},Q_\epsilon}$ *is the visibility graph at $\mathcal{P}$ for $Q_\epsilon$.*
(iii) *The* $r$-disk graph $\mathcal{G}_{r\text{-disk}}$ *at $\mathcal{P}$ is the graph with node set $\mathcal{P}$ and with edge set $\mathcal{E}_{r\text{-disk}}(\mathcal{P})$ defined by: $(p_i, p_j) \in \mathcal{E}_{r\text{-disk}}(\mathcal{P})$ if and only if $\|p_i - p_j\| \leq r$.*
(iv) *The* $r$-range visibility graph $\mathcal{G}_{r\text{-vis},\, Q}$ *at $\mathcal{P}$ is the graph with node set $\mathcal{P}$ and with edge set $\mathcal{E}_{r\text{-disk}}(\mathcal{P}) \cap \mathcal{E}_{\text{vis},Q}(\mathcal{P})$.*
(v) *A* Euclidean Minimum Spanning Tree $\mathcal{G}_{\text{EMST},\mathcal{G}}$ *at $\mathcal{P}$ of a proximity graph $\mathcal{G}$ is a minimum-length spanning tree of $\mathcal{G}(\mathcal{P})$ whose edge $(p_i, p_j)$ has length $\|p_i - p_j\|$. If $\mathcal{G}(\mathcal{P})$ is not connected, then $\mathcal{G}_{\text{EMST},\mathcal{G}}(P)$ is the union of Euclidean Minimum Spanning Trees of its connected components.*

In other words, two points $p, q$ are neighbors in the $r$-range visibility graph, for instance, if and only if they are mutually visible and separated by a distance less than or equal to $r$. Example graphs are shown in Figure 4. General properties of proximity graphs are defined in [14], [7]. For simplicity, when $\mathcal{G}$ is the complete graph, we denote the Euclidean Minimum Spanning Tree of $\mathcal{G}$ by $\mathcal{G}_{\text{EMST}}$.

We say that two proximity graphs $\mathcal{G}_1$ and $\mathcal{G}_2$ *have the same connected components* if, for all sets of points $\mathcal{P}$, the graphs $\mathcal{G}_1(\mathcal{P})$ and $\mathcal{G}_2(\mathcal{P})$ have the same number of connected components consisting of the same vertices.

**Definition II.6 (Neighbors set)** *Given a set of points $\mathcal{P} = \{p_1, \dots, p_n\}$ and a proximity graph $\mathcal{G}$, we let $\mathcal{N}_i(\mathcal{G}, \mathcal{P}) = \mathcal{N}_{i,\mathcal{G}}(\mathcal{P})$ denote the* set of neighbors including itself *of $p_i$. In* *other words, if $\{p_{i_1}, \dots, p_{i_m}\}$ are the neighbors of $p_i$ in $\mathcal{G}$ at $\mathcal{P}$, then $\mathcal{N}_i(\mathcal{G}, \mathcal{P}) = \mathcal{N}_{i,\mathcal{G}}(\mathcal{P}) = \{p_{i_1}, \dots, p_{i_m}\} \cup \{p_i\}$.*

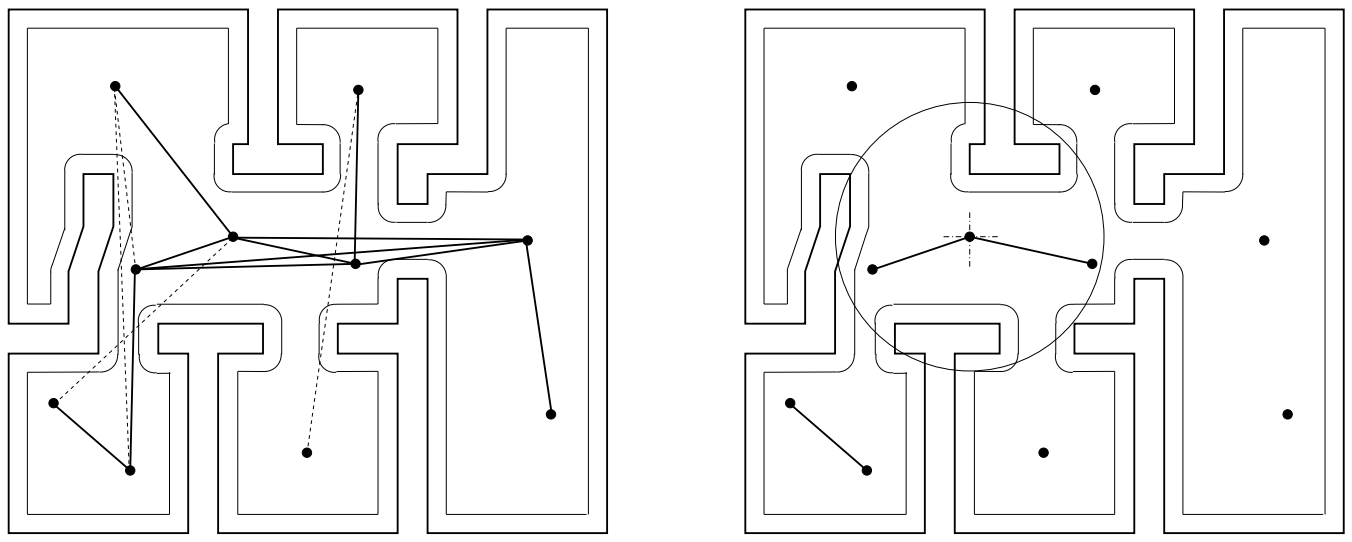

Fig. 4. The figure on the left shows the visibility graph (whose edges are the solid lines as well as the dashed lines) and the $\epsilon$-robust visibility graph (whose edges are the solid lines alone) of a set of points in a nonconvex polygon. The figure on the right shows the $r$-range $\epsilon$-robust visibility graph. The disk in the figure shows the sensing range for one of the agents.

**Definition II.7 (Relative convex hull)** *Take an allowable environment $Q$.*

(i) *$X \subseteq Q$ is* relatively convex *if the shortest path inside $Q$ connecting any two points of $X$ is contained in $X$.*
(ii) *The* relative convex hull $\mathrm{rco}(X, Q)$ *of $X \subset Q$ is the smallest*[4] *relatively convex subset of $Q$ that contains $X$.*
(iii) *If $X$ is a finite set of points, then a* vertex *of* $\mathrm{rco}(X, Q)$ *is a point $p \in X$ with the property that* $\mathrm{rco}(X \setminus \{p\}, Q)$ *is a strict subset of* $\mathrm{rco}(X, Q)$*. The set of vertices of* $\mathrm{rco}(X, Q)$ *is denoted by* $\mathrm{Ve}(\mathrm{rco}(X, Q))$.

The relative convex hull of an example set of points and its vertices are shown in Figure 5. In what follows we will need the notion of perimeter of certain sets, and in particular, of the relative convex hull of a collection of points.

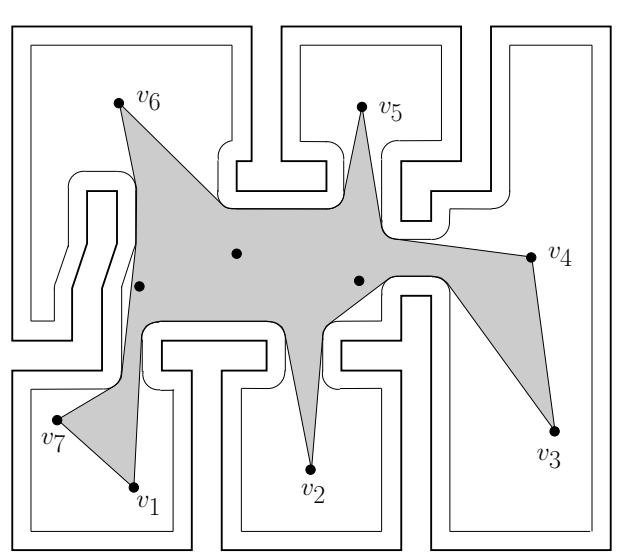

Fig. 5. Relative convex hull $\mathrm{rco}(X, Q_\epsilon)$ of a set of points $X$ (solid disks) inside a the $\epsilon$-contraction of an allowable set $Q$. The set of vertices $\mathrm{Ve}(\mathrm{rco}(X, Q_\epsilon))$ is the set $\{v_1, \dots, v_7\}$.

**Definition II.8 (Perimeter)** *Take an allowable environment $Q$ and a closed subset $X \subset Q$.*

(i) *If $X$ has measurable boundary $\partial X$ and is equal to the closure of its interior, then* $\mathrm{perimeter}(X)$ *is the length of $\partial X$.*
(ii) *If* $\mathrm{rco}(X, Q)$ *is not equal to the closure of its interior, then* $\mathrm{perimeter}(\mathrm{rco}(X, Q))$ *is the length of the shortest measurable curve inside $Q$ enclosing $X$.*

[4]That is, $\mathrm{rco}(X, Q)$ is the intersection of all relatively convex subsets of $Q$ that contain $X$.

**Remarks II.9** (i) If $\mathrm{rco}(X, Q)$ is equal to the closure of its interior, then its boundary is the shortest measurable curve inside $Q$ enclosing $X$ (i.e., the two definitions of perimeter are equivalent). On the other hand, if $\mathrm{rco}(X, Q)$ is a segment, then Definition II.8(ii) says that the perimeter of $\mathrm{rco}(X, Q)$ is twice its length.
(ii) The key property of Definition II.8 is that, if $X$ is a finite set of points in $Q$, then the perimeter of $\mathrm{rco}(X, Q)$ depends continuously on the points in $X$. □

## III. Synchronous robots with visibility sensors and the rendezvous and connectivity maintenance problems

In this section we model a group of $n$ robots with visibility sensors in a given allowable environment $Q$. We assume that $\epsilon$ is a known positive constant sufficiently small so that $Q_\epsilon$ is allowable. For $i \in \{1, \dots, n\}$, we model the $i$th robot as a point $p_i \in Q$ and we refer to Section VI for an extension to a disk model. We make the following modeling assumptions:

*Synchronized controlled motion model*: Robot $i$ moves at time $t \in \mathbb{Z}_{\geq 0}$ for a unit period of time, according to the discrete-time control system

$$p_i[t+1] = p_i[t] + u_i[t]. \tag{1}$$

We assume that there is a maximum step size $s_{\text{max}} > 0$ for any robot, that is, $\|u_i\| \leq s_{\text{max}}$. Note that the $n$ identical robots are *synchronized* in the sense that the calculation of $u[t]$ in equation (1) takes place at the same times $t$ for all robots.

*Sensing model*: Robot $i$ senses (i) the presence and the position of any other robot that is visible and within distance $r$ from $p_i$, and (ii) the subset of $\partial Q$ that is visible and within distance $(r + \epsilon)$ from $p_i$. This in turn implies that the robot can sense the subset of $\partial Q_\epsilon$ that is visible and within distance $r$ from $p_i$. It is convenient to define the *sensing region from position* $p_i$ to be $\mathcal{S}(p_i) = \mathcal{V}(p_i, \epsilon) \cap B(p_i, r)$. The range $r$ is the same for all robots.

Note that, by definition, two robots with visibility sensors detect each other's presence and relative position if and only if they are neighbors in the robust visibility graph $\mathcal{G}_{\text{vis}, Q_\epsilon}$.

**Remark III.1 (No common reference frame)** The model presented above assumes the ability of robots to sense absolute positions of other robots; this assumption is only made to keep the presentation as simple as possible. In this and subsequent remarks, we treat the more realistic setting in which the $n$ robots have $n$ distinct reference frames $\Sigma_1, \dots, \Sigma_n$. We let $\Sigma_0$ denote a fixed reference frame. Notation-wise, a point $q$, a vector $w$, and a set of points $S$ expressed with respect to frame $\Sigma_i$ are denoted by $q^i$, $w^i$ and $S^i$, respectively. For example, this means that $Q^i$ is the environment $Q$ as expressed in frame $\Sigma_i$. We assume that the origin of $\Sigma_i$ is $p_i$ and that the orientation of $\Sigma_i$ with respect to $\Sigma_0$ is $R_i^0 \in \mathrm{SO}(2)$. Therefore, changes of reference frames are described by the equations: $q^0 = R_i^0 q^i + p_i^0$, $w^0 = R_i^0 w^i$, and $S^0 = R_i^0 S^i + p_i^0$. If we let $\mathcal{V}_{Q^j}(p_i^j, \epsilon)$ denote the visibility set expressed in $\Sigma_j$, for $j \in \{0, 1, \dots, n\}$, then one can define

$$\mathcal{S}(p_i^j, Q^j) = \mathcal{V}_{Q^j}(p_i^j, \epsilon) \cap B(p_i^j, r),$$

and verify that $\mathcal{S}(p_i^0, Q^0) = R_i^0 \mathcal{S}(p_i^i, Q^i) + p_i^0$. Note that $p_i^i = 0$.

Finally, we can describe our motion and sensing model under the no common reference frame assumption. Robot $i$ moves according to

$$p_i^0[t+1] = p_i^0[t] + R_i^0[t] u_i[t], \tag{2}$$

and it senses the robot positions $p_j^i$ and the subset of $(\partial Q)^i$ that are within the sensing region $\mathcal{S}(p_i^i, Q^i)$. □

We now state the two control design problems addressed in this paper for groups of robots with visibility sensors.

**Problem III.2 (Rendezvous)** The *rendezvous problem* is to steer each agent to a common location inside the environment $Q_\epsilon$. This objective is to be achieved (1) with the limited information flow described in the model above, and (2) under the reasonable assumption that the initial position of the robots $\mathcal{P}[0] = \{p_1[0], \dots, p_n[0]\}$ gives rise to a connected robust visibility graph $\mathcal{G}_{\text{vis}, Q_\epsilon}$ at $\mathcal{P}[0]$. □

As one might imagine, the approach to solving the rendezvous problem involves two main ideas: first, the underlying proximity graph should not lose connectivity during the evolution of the group; second, while preserving the connectivity of the graph, the agents must move closer to each other. This discussion motivates a second complementary objective.

**Problem III.3 (Connectivity maintenance)** The *connectivity maintenance problem* is to design (state dependent) control constraints sets with the following property: if each agent's control takes values in the control constraint set, then the agents move in such a way that the number of connected components of $\mathcal{G}_{\text{vis}, Q_\epsilon}$ (evaluated at the agents' states) does not increase with time. □

## IV. The connectivity maintenance problem

In this section, we maintain the connectivity of the group of agents with visibility sensors by designing control constraint sets that guarantee that every edge of $\mathcal{G}_{r\text{-vis},\, Q_\epsilon}$ (i.e., every pair of mutually range-limited visible robots) is preserved. We have three objectives in doing so. First, the sets need to depend continuously on the position of the robots. Second, the sets need to be computed in a distributed way based only on the available sensory information. Third, the control constraint sets should be as "large" as possible so as to minimally constrain the motion of the robots. Because it appears difficult to formalize the notion of "largest continuous constraint set that can be computed in a distributed fashion," we instead propose a geometric strategy to compute appropriate constraint sets and we show in the next section that our proposed geometric strategy is sufficiently efficient for the rendezvous problem.

### A. *Preserving mutual visibility: The* `Constraint Set Generator Algorithm`

Consider a pair of robots in an environment $Q$ that are $\epsilon$-robustly visible to each other and separated by a distance not larger than $r$. To preserve this range-limited mutual visibility property, we restrict the motion of the robots to an appropriate subset of the environment. This idea is inspired by [5] and we begin by stating the result therein. Let the sensing region of robot $i$ located at $p_i$ be $\mathcal{S}(p_i) = B(p_i, r)$, for some $r > 0$. If at any time instant $t$, $\|p_i[t] - p_j[t]\| \le r$ then to ensure that at the next time instant $t+1$, $\|p_i[t+1] - p_j[t+1]\| \le r$, it suffices to impose the following constraints on the motion of robots $i$ and $j$:

$$p_i[t+1],\ p_j[t+1] \in B\Big(\frac{p_i[t] + p_j[t]}{2}, \frac{r}{2}\Big),$$

or, equivalently,

$$u_i[t],\ u_j[t] \in B\Big(\frac{p_j[t] - p_i[t]}{2}, \frac{r}{2}\Big).$$

In summary, $B(\frac{p_j - p_i}{2}, \frac{r}{2})$ is the control constraint set for robot $i$ and $j$. This constraint is illustrated in Figure 6 (left).

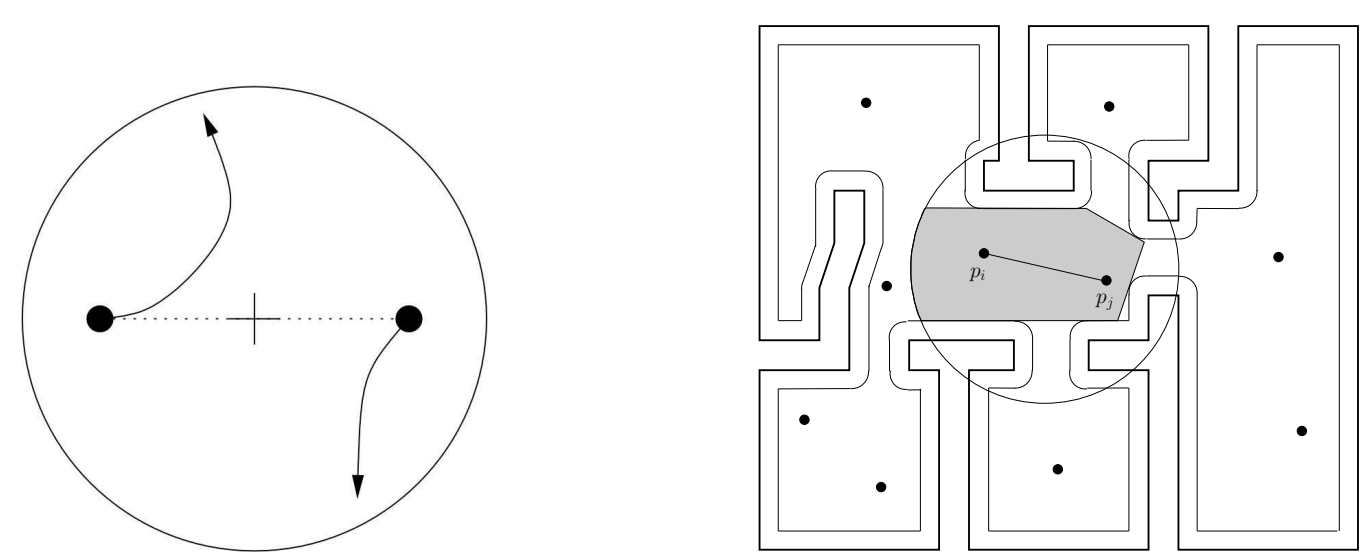

Fig. 6. In the figure on the left, starting from $p_i$ and $p_j$, the robots are restricted to move inside the disk centered at $\frac{p_i + p_j}{2}$ with radius $\frac{r}{2}$. In the figure on the right, the robots are constrained to move inside the shaded region which is a convex subset of $Q_\epsilon$ intersected with the disk centered at $\frac{p_i + p_j}{2}$ with radius $\frac{r}{2}$.

Let us now consider the case when a robot $i$ is located at $p_i$ in a nonconvex environment $Q$ with sensing region $\mathcal{S}(p_i) = \mathcal{V}(p_i, \epsilon) \cap B(p_i, r)$. If at any time instant $t$, we have that $\|p_i[t] - p_j[t]\| \le r$ and $[p_i[t], p_j[t]] \in Q_\epsilon$, then to ensure that $\|p_i[t+1] - p_j[t+1]\| \le r$ and $[p_i[t+1], p_j[t+1]] \in Q_\epsilon$, it suffices to require that:

$$p_i[t+1],\ p_j[t+1] \in \mathcal{C},$$

where $\mathcal{C}$ is any convex subset of $Q_\epsilon \cap B\big(\frac{p_i[t] + p_j[t]}{2}, \frac{r}{2}\big)$; see Figure 6 (right). Equivalently,

$$u_i[t] \in \mathcal{C} - p_i[t],\ u_j[t] \in \mathcal{C} - p_j[t],$$

where $\mathcal{C} - p_i[t]$ and $\mathcal{C} - p_i[t]$ are the sets $\{p - p_i[t] \mid p \in \mathcal{C}\}$ and $\{p - p_j[t] \mid p \in \mathcal{C}\}$, respectively. Note that both robots $i$ and $j$ must independently compute the same set $\mathcal{C}$. Given the positions $p_i, p_j$ in an environment $Q$, Table I describes the `Constraint Set Generator Algorithm`, a geometric strategy for each robot to compute a constraint set $\mathcal{C} = \mathcal{C}_Q(p_i, p_j)$ that changes continuously with $p_i$ and $p_j$. Figure 7 illustrates a step-by-step execution of the algorithm.

TABLE I

`Constraint Set Generator Algorithm`

**Goal:** Generate convex sets to act as constraints to preserve mutual visibility

**Given:** $(p_i, p_j) \in Q_\epsilon^2$ such that $[p_i, p_j] \subseteq Q_\epsilon$ and $p_j \in B(p_i, r)$

Robot $i \in \{1, \dots, n\}$ executes the following computations:

1: $\mathcal{C}_{\text{temp}} := \mathcal{V}(p_i, \epsilon) \cap B(\frac{p_i+p_j}{2}, \frac{r}{2})$
2: **while** $\partial \mathcal{C}_{\text{temp}}$ contains a concavity **do**
3: $\quad v :=$ a strictly concave point of $\partial \mathcal{C}_{\text{temp}}$ closest to the segment $[p_i, p_j]$
4: $\quad \mathcal{C}_{\text{temp}} := \mathcal{C}_{\text{temp}} \cap H_{Q_\epsilon}(v)$
5: **end while**
6: **return:** $\mathcal{C}_Q(p_i, p_j) := \mathcal{C}_{\text{temp}}$

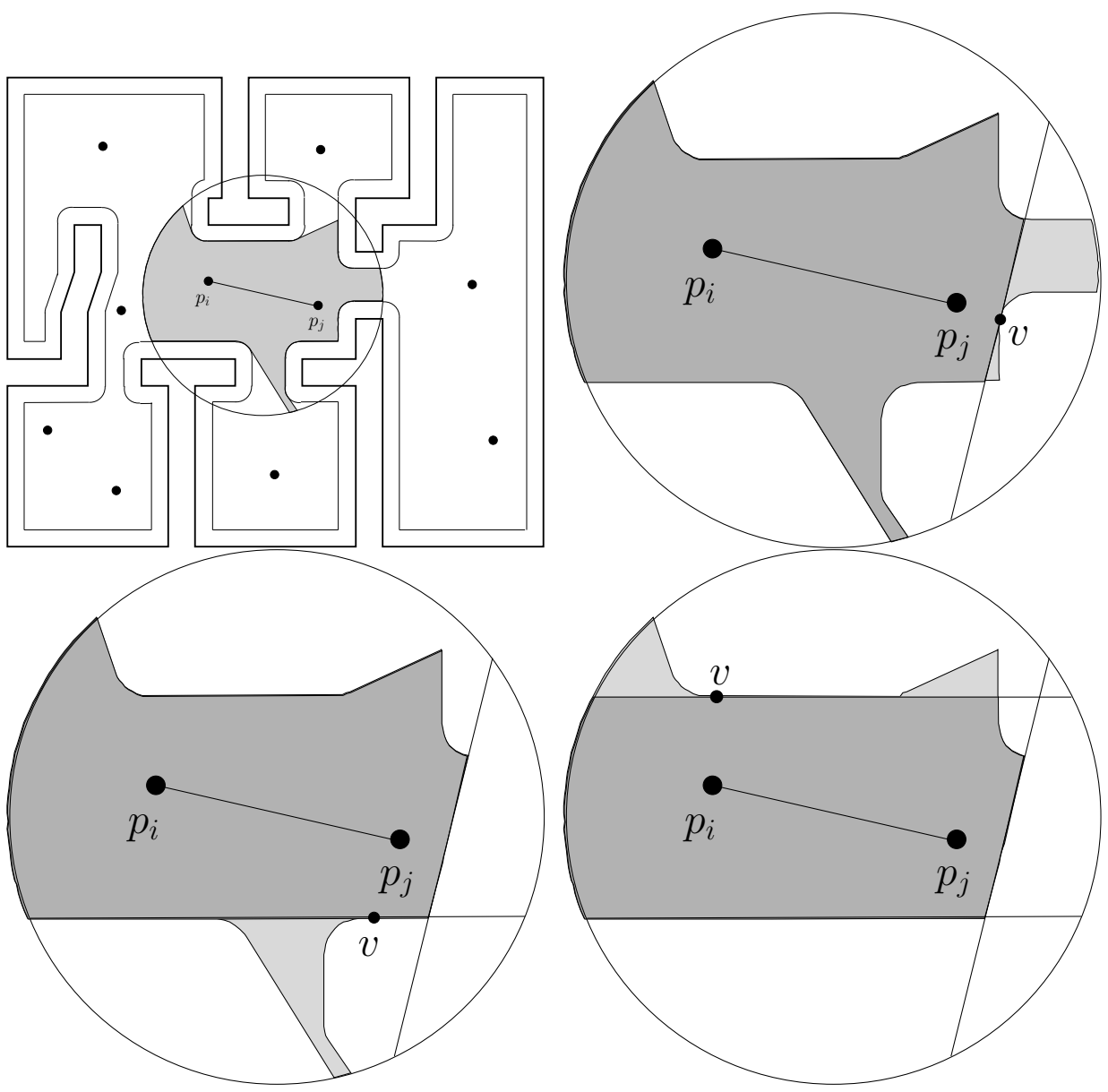

Fig. 7. From left to right and top to bottom, a sample incomplete run of the `Constraint Set Generator Algorithm` (cf. Table I). The top left figure shows $\mathcal{C}_{\text{temp}} := \mathcal{V}(p_i, \epsilon) \cap B(\frac{p_i+p_j}{2}, \frac{r}{2})$. In all the other figures, the lightly and darkly shaded regions together represent $\mathcal{C}_{\text{temp}}$. The darkly shaded region represents $\mathcal{C}_{\text{temp}} \cap H_Q(v)$, where $v$ is as described in step `3:`. The final outcome of the algorithm, $\mathcal{C}_Q(p_i, p_j)$, is shown in Figure 6 (right).

In step `3:` of the algorithm, note that there can be multiple distinct points belonging to distinct concavities satisfying the required property. In that case, $v$ can be chosen to be any one of them. The following lemma justifies this observation.

**Lemma IV.1** *Throughout the execution of the* `Constraint Set Generator Algorithm` *in Table I, let $v_1, v_2$ be two strictly concave points on $\partial \mathcal{C}_{\text{temp}}$ that are closest to $[p_i, p_j]$. Then $v_1 \in \mathcal{C}_{\text{temp}} \cap H_{Q_\epsilon}(v_2)$ and vice versa.*

Next, we characterize the main properties of the `Constraint Set Generator Algorithm` and the corresponding convex sets. Notice that the constraint set is defined at any point in the following set:

$$J = \{(p_i, p_j) \in Q_\epsilon^2 \mid [p_i, p_j] \in Q_\epsilon, \ \|p_i - p_j\| \le r\}.$$

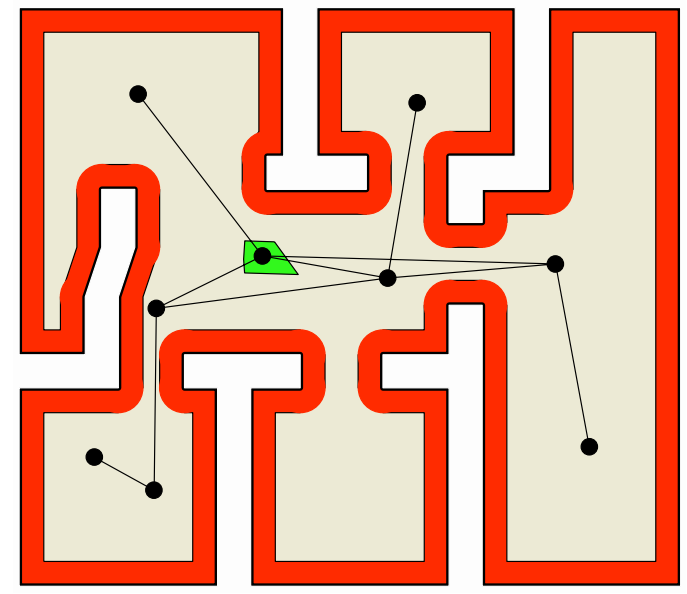

Fig. 8. The green convex set in the center represents $C_{p_i,Q}(\mathcal{N}_{i,\mathcal{G}_{r\text{-vis},Q_\epsilon}})$. The black disks represent the position of the robots. The straight line segments between pairs of robots represent edges of $\mathcal{G}_{r\text{-vis},Q_\epsilon}$. Here, $p_i$ is the black disk contained in the constraint set.

**Proposition IV.2 (Properties of the `Constraint Set Generator Algorithm`)** *Given an allowable environment $Q$ with $\kappa$ strict concavities, $\epsilon > 0$ and $(p_i, p_j) \in J$, the following statements hold:*

(i) *The* `Constraint Set Generator Algorithm` *terminates in at most $\kappa$ steps;*
(ii) *$\mathcal{C}_Q(p_i, p_j)$ is nonempty, compact and convex;*
(iii) *$\mathcal{C}_Q(p_i, p_j) = \mathcal{C}_Q(p_j, p_i)$; and*
(iv) *The set-valued map $\mathcal{C}_Q$ is closed[5] at every point of $J$.*

**Remark IV.3 (No common reference frame: continued)** Consider a group of robots with visibility sensors and no common reference frame. With the notation and assumptions described in Remark III.1, on can verify that the constraint sets transform under changes of coordinate frames according to:

$$\mathcal{C}_{Q^0}(p_i^0, p_j^0) = R_i^0 \mathcal{C}_{Q^i}(p_i^i, p_j^i) + p_i^0. \tag{3}$$

We omit the proof in the interest of brevity. □

For each pair of mutually visible robots, the execution of the `Constraint Set Generator Algorithm` outputs a control constraint set such that, if the robots' motions are constrained to it, then the robots remain mutually visible. Clearly, given a connected graph at time $t$, if every robot remains connected with all its neighbors at time $t+1$ (i.e., each pair of mutually visible robots remain mutually visible), then the connectivity of the graph is preserved. This can be accomplished as follows. For robot $i \in \{1, \dots, n\}$ at $p_i \in Q_\epsilon$, define the control constraint set

$$C_{p_i,Q}(\mathcal{N}_{i,\mathcal{G}_{r\text{-vis},\,Q_\epsilon}}) = \bigcap_{p_j \in \mathcal{N}_{i,\mathcal{G}_{r\text{-vis},\,Q_\epsilon}}} \mathcal{C}_Q(p_i, p_j). \tag{4}$$

Now, if $u_i \in C_{p_i,Q}(\mathcal{N}_{i,\mathcal{G}_{r\text{-vis},\,Q_\epsilon}}) - p_i$, for all $i \in \{1, \dots, n\}$, then all neighboring relationships in $\mathcal{G}_{r\text{-vis},\,Q_\epsilon}$ are preserved

[5]Let $\Omega$ map points in $X$ to all possible subsets of $Y$. Then the set-valued map, $\Omega$, is open at a point $x \in X$ if for any sequence $\{x_k\}$ in $X$, $x_k \to x$ and $y \in \Omega(x)$ implies the existence of a number $m$ and a sequence $\{y_k\}$ in $Y$ such that $y_k \in \Omega(x_k)$ for $k \ge m$ and $y_k \to y$. The map $\Omega$ is closed at a point $x \in X$ if for any sequence $\{x_k\}$ in $X$, $x_k \to x$, $y_k \to y$ and $y_k \in \Omega(x_k)$ imply that $y \in \Omega(x)$. $\Omega$ is continuous at any point $x \in X$ if it is both open and closed at x

at the next time instant. Using inputs that satisfy these constraints, the number of edges in $\mathcal{G}_{r\text{-vis},\,Q_\epsilon}$ is guaranteed to be nondecreasing.

### *B. The locally-cliqueless visibility graph*

In this section, we propose the construction of constraint sets that are, in general, larger than $C_{p_i,Q}(\mathcal{N}_{i,\mathcal{G}_{r\text{-vis},\,Q_\epsilon}})$. To do this, we define the notion of *locally-cliqueless graph*. The locally-cliqueless graph of a proximity graph $\mathcal{G}$ is a subgraph of $\mathcal{G}$, and therefore has generally fewer edges, but it has the same number of connected component as $\mathcal{G}$. This fundamental property will be very useful in the design of less conservative constraint sets.

Before proceeding with the definition of locally-cliqueless graph, let us recall that (i) a *clique* of a graph is a complete subgraph of it, and (ii) a *maximal clique of an edge* is a clique of the graph that contains the edge and is not a strict subgraph of any other clique of the graph that also contains the edge.

**Definition IV.4 (Locally-cliqueless graph of a proximity graph)** *Given a point set $\mathcal{P}$ and an allowable environment $Q$, the* locally-cliqueless graph $\mathcal{G}_{\text{lc},\mathcal{G}}$ *at $\mathcal{P}$ of a proximity graph $\mathcal{G}$ is the proximity graph with node set $\mathcal{P}$ and with edge set $\mathcal{E}_{\text{lc},\mathcal{G}}(\mathcal{P})$ defined by: $(p_i, p_j) \in \mathcal{E}_{\mathcal{G}_{\text{lc},\mathcal{G}}}(\mathcal{P})$ if and only if $(p_i, p_j) \in \mathcal{E}_{\mathcal{G}}(\mathcal{P})$ and $(p_i, p_j)$ belongs to a set $\mathcal{E}_{\mathcal{G}_{\text{EMST}}}(\mathcal{P}')$ for any maximal clique $\mathcal{P}'$ of the edge $(p_i, p_j)$ in $\mathcal{G}$.*

In combinatorial optimization, it is a well-known that finding the maximal clique of a graph is an NP complete problem. However, efficient polynomial time heuristics are detailed in [16].

For simplicity, we will refer to the locally-cliqueless graph of the proximity graphs $\mathcal{G}_{\text{vis},Q}$, $\mathcal{G}_{\text{vis},Q_\epsilon}$ or $\mathcal{G}_{\text{r-vis},Q_\epsilon}$ as *locally-cliqueless visibility graphs*. Figure 9 shows an example of a locally-cliqueless visibility graph.

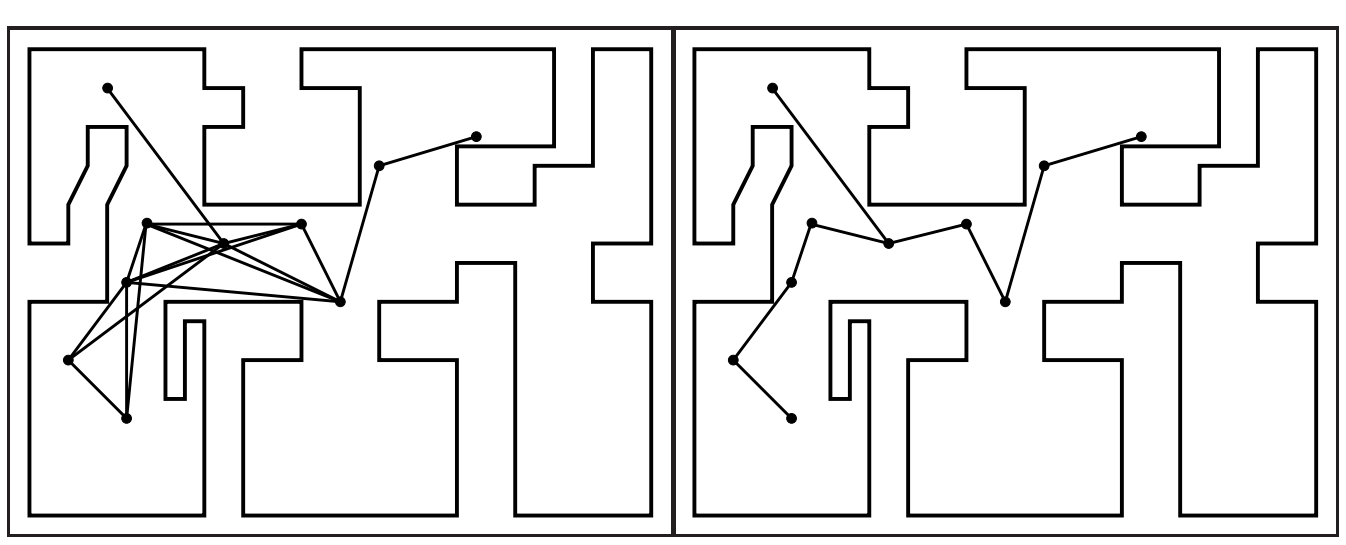

Fig. 9. Visibility graph (left) and locally-cliqueless visibility graph (right).

**Theorem IV.5 (Properties of a locally-cliqueless graph of a proximity graph)** *Let $\mathcal{G}$ be a proximity graph. Then, the following statements hold:*

(i) $\mathcal{G}_{\text{EMST},\mathcal{G}} \subseteq \mathcal{G}_{\text{lc},\mathcal{G}} \subseteq \mathcal{G}$;
(ii) *$\mathcal{G}_{\text{lc},\mathcal{G}}$ and $\mathcal{G}$ have the same connected components.*

In general, the inclusions in Theorem IV.5(i) are strict. Figure 10 shows an example where $\mathcal{G}_{\text{EMST},\mathcal{G}_{\text{vis},Q}} \subsetneq \mathcal{G}_{\text{lc},\mathcal{G}_{\text{vis},Q}} \subsetneq \mathcal{G}_{\text{vis},Q}$.

The next result follows directly from Theorem IV.5.

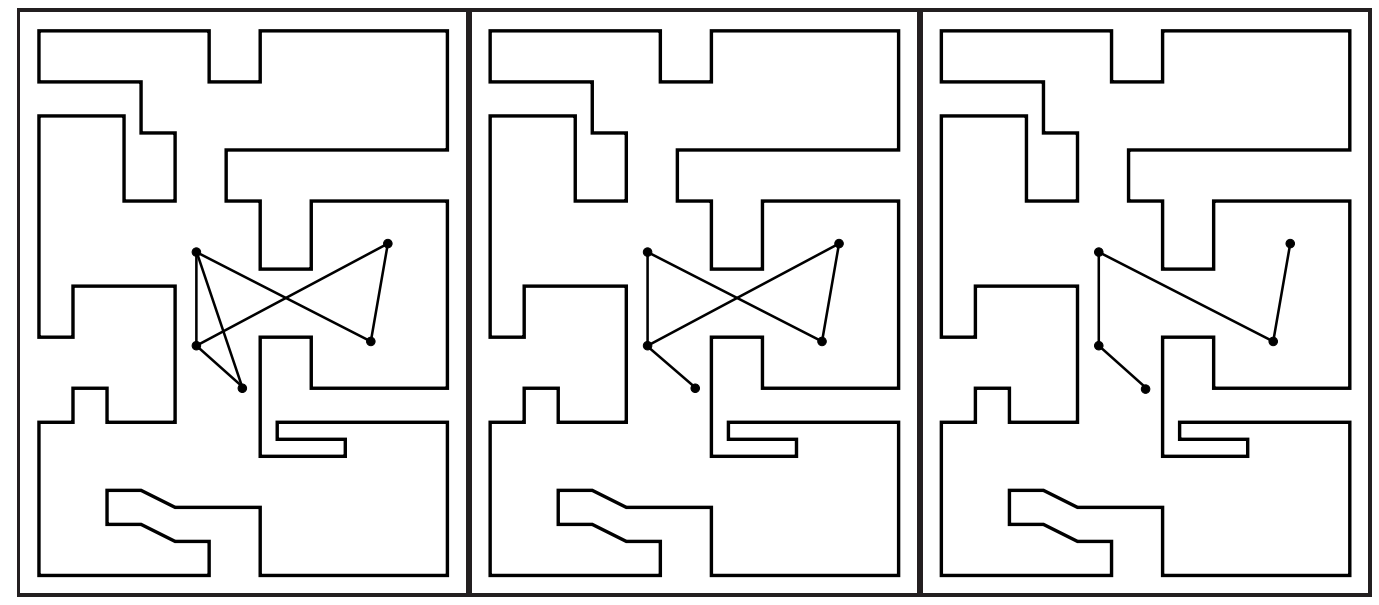

Fig. 10. From left to right, visibility graph, locally-cliqueless graph and Euclidean Minimum Spanning Tree of the visibility graph.

**Corollary IV.6 (Properties of locally-cliqueless visibility graphs)** *Let $Q$ and $Q_\epsilon$, $\epsilon > 0$, be allowable environments. Let $\mathcal{G}$ be either one of the graphs $\mathcal{G}_{\text{vis},Q}$, $\mathcal{G}_{\text{vis},Q_\epsilon}$ or $\mathcal{G}_{\text{r-vis},Q_\epsilon}$. Then, the following statements hold:*

(i) $\mathcal{G}_{\text{EMST},\mathcal{G}} \subseteq \mathcal{G}_{\text{lc},\mathcal{G}} \subseteq \mathcal{G}$;
(ii) *$\mathcal{G}_{\text{lc},\mathcal{G}}$ and $\mathcal{G}$ have the same connected components.*

Let us now proceed to define new constraint sets that are in general larger than the ones defined in (4). For simplicity, let $\mathcal{G} = \mathcal{G}_{\text{r-vis},Q_\epsilon}$, and consider its locally-cliqueless graph $\mathcal{G}_{\text{lc},\,\mathcal{G}}$. For robot $i \in \{1, \dots, n\}$ at position $p_i$, define the constraint set

$$C_{p_i,Q}(\mathcal{N}_{i,\mathcal{G}_{\text{lc},\,\mathcal{G}}}) = \bigcap_{p_j \in \mathcal{N}_{i,\mathcal{G}_{\text{lc},\,\mathcal{G}}}} \mathcal{C}_Q(p_i, p_j). \tag{5}$$

Since $\mathcal{G}_{\text{lc},\mathcal{G}}$ is a subgraph of $\mathcal{G}$ (cf. Corollary IV.6(i)), we have $\mathcal{N}_{i,\mathcal{G}_{\text{lc},\,\mathcal{G}}} \subseteq \mathcal{N}_{i,\mathcal{G}} = \mathcal{N}_{i,\mathcal{G}_{r\text{-vis},\,Q_\epsilon}}$, and therefore

$$C_{p_i,Q}(\mathcal{N}_{i,\mathcal{G}_{r\text{-vis},\,Q_\epsilon}}) \subseteq C_{p_i,Q}(\mathcal{N}_{i,\mathcal{G}_{\text{lc},\,\mathcal{G}}}).$$

In general, since $\mathcal{G}_{\text{lc},\mathcal{G}}$ is a strict subgraph of $\mathcal{G}$, the set $C_{p_i,Q}(\mathcal{N}_{i,\mathcal{G}_{\text{lc},\,\mathcal{G}}})$ is strictly larger that $C_{p_i,Q}(\mathcal{N}_{i,\mathcal{G}_{r\text{-vis},\,Q_\epsilon}})$.

Now, if $u_i \in C_{p_i,Q}(\mathcal{N}_{i,\mathcal{G}_{\text{lc},\,\mathcal{G}}}) - p_i$ for all $i \in \{1, \dots, n\}$, then all neighboring relationships in the graph $\mathcal{G}_{\text{lc},\,\mathcal{G}}$ are preserved at the next time instant. As a consequence, it follows from Corollary IV.6(ii) that the connected components of $\mathcal{G}_{r\text{-vis},\,Q_\epsilon}$ are also preserved at the next time instant. Thus, we have found constraint sets (5) for the input that are larger than the constraint sets (4), and are yet sufficient to preserve the connectivity of the overall group.

**Remark IV.7 (Distributed computation of locally-cliqueless visibility graphs)** According to the model specified in Section III, each robot can detect all other robots in its sensing region $\mathcal{S}(p_i) = \mathcal{V}(p_i, \epsilon) \cap B(p_i, r)$, i.e., its neighbors in the graph $\mathcal{G}_{r\text{-vis},\,Q_\epsilon}$. Given the construction of the constraint sets in this section, it is important to guarantee that the set of neighbors of robot $i$ in the locally-cliqueless graph $\mathcal{G}_{\text{lc},\mathcal{G}}$ can be computed locally by robot $i$. From the definition of the locally-cliqueless graph, this is indeed possible if a robot $i$ can detect whether another robot $j$ in its sensing region $\mathcal{S}(p_i)$ belongs to a clique of the graph $\mathcal{G}_{r\text{-vis},\,Q_\epsilon}$. This is equivalent to being able to check if two robots $p_k, p_l \in \mathcal{S}(p_i)$ satisfy the condition that $p_k \in \mathcal{S}(p_l)$ and vice versa. Note that $p_k \in \mathcal{S}(p_l)$ is equivalent to $\|p_k - p_l\| \leq r$ and $[p_k, p_l] \subseteq Q_\epsilon$. Given that $p_k - p_l = (p_k - p_i) - (p_l - p_i)$, the vector $p_k - p_l$ (and hence $\|p_k - p_l\|$) can be computed based on local sensing alone. Now, checking if $[p_k, p_l] \subseteq Q_\epsilon$ is possible only if $Q_\epsilon$ does not contain any hole; see Figure 11. In such a case, it suffices to check if the entire line segment $[p_k, p_l]$ is visible from $p_i$ or not.

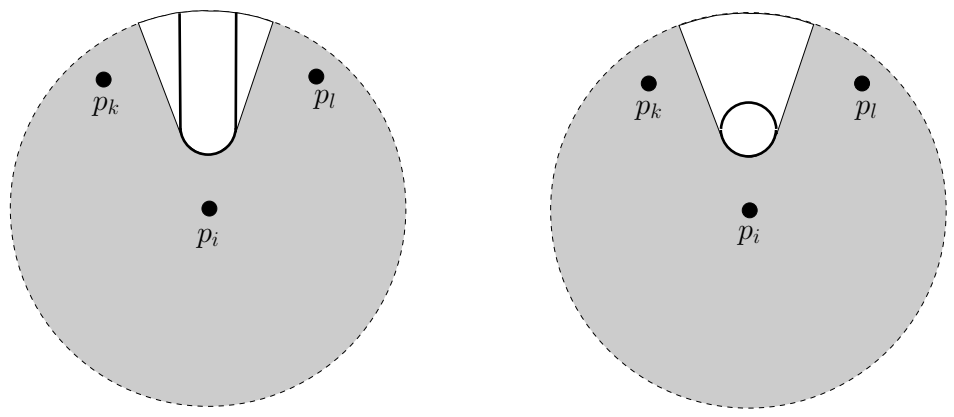

Fig. 11. The dashed circle is centered at $p_i$ and is of radius $r$. The thick curves represent the boundary of $Q_\epsilon$; the one on the left represents the outer boundary whereas the one on the right represents a hole in the environment.

Along these same lines it is possible to state that the locally-cliqueless visibility graph can be computed also under the "no common reference frame" model described in Remarks III.1 and IV.3. □

## V. The rendezvous problem: Algorithm design and analysis results

In this section, we solve the rendezvous problem through a novel `Perimeter Minimizing Algorithm`. The algorithm is inspired by the one introduced in [5] but is unique in many different ways. The rendezvous algorithm uses different graphs to maintain connectivity and to move closer to other robots. Instead of moving towards the circumcenter of the neighboring robots, the robots move towards the center of a suitably defined motion constraint set.

The section is organized as follows. We present the algorithm in Subsection V-A followed by its main convergence properties in Subsection V-B.

### A. *The* `Perimeter Minimizing Algorithm`

We begin with an informal description of the `Perimeter Minimizing Algorithm` over graphs $\mathcal{G}_{\text{sens}}$ and $\mathcal{G}_{\text{constr}}$. The sensing graph $\mathcal{G}_{\text{sens}}$ is $\mathcal{G}_{r\text{-vis},Q_\epsilon}$ while the constraint graph $\mathcal{G}_{\text{constr}}$ is either $\mathcal{G}_{\text{sens}}$ or $\mathcal{G}_{\text{lc},\mathcal{G}_{\text{sens}}}$:

> Every robot $i$ performs the following tasks: (i) it acquires the positions of other robots that are its neighbors according to $\mathcal{G}_{\text{sens}}$; (ii) it computes a point that is "closer" to the robots it senses, and (iii) it moves toward this point while maintaining connectivity with its neighbors according to $\mathcal{G}_{\text{constr}}$.

The algorithm is formally described in Table II; Figure 1 in the Introduction illustrates an example execution.

TABLE II

`Perimeter Minimizing Algorithm`

**Assumes:** (i) $s_{\max} > 0$ is the maximal step size
(ii) $Q, Q_\epsilon$ are allowable
(iii) $\mathcal{G}_{\text{sens}}$ is $\mathcal{G}_{r\text{-vis},\,Q_\epsilon}$; $\mathcal{G}_{\text{constr}}$ is either $\mathcal{G}_{\text{sens}}$ or $\mathcal{G}_{\text{lc},\mathcal{G}_{\text{sens}}}$

Each robot $i \in \{1,\dots,n\}$ executes the following steps at each time instant:

1: acquire $\{p_{i_1},\dots,p_{i_m}\}$ := positions of robots within $p_i$ sensing region
2: compute $\mathcal{N}_{i,\mathcal{G}_{\text{sens}}}$ and $\mathcal{N}_{i,\mathcal{G}_{\text{constr}}}$
3: compute $X_i := C_{p_i,Q}(\mathcal{N}_{i,\mathcal{G}_{\text{constr}}}) \cap \text{rco}(\mathcal{N}_{i,\mathcal{G}_{\text{sens}}}, \mathcal{V}(p_i,\epsilon))$
4: compute $p_i^* := \text{CC}(X_i)$
5: **return:** $u_i := \dfrac{\min(s_{\max}, \|p_i^* - p_i\|)}{\|p_i^* - p_i\|}(p_i^* - p_i)$

**Remarks V.1** (i) According to the algorithm proposed in [5] the robots move towards the circumcenter of their neighbors position. In the `Perimeter Minimizing Algorithm` the robots move towards the circumcenter of their constraint set.

(ii) We prove later in Lemma II.2 in Appendix II that $X_i$ is convex and that, in turn, $\text{CC}(X_i)$ is well-defined. Because $\text{CC}(X_i) \in X_i$, we know that $p_i^* \in X_i$. Therefore, $u_i \in X_i - p_i \subseteq C_{p_i,Q}(\mathcal{N}_{i,\mathcal{G}_{\text{constr}}}) - p_i$ and, in turn, $p_i$ at the next time instant belongs to $C_{p_i,Q}(\mathcal{N}_{i,\mathcal{G}_{\text{constr}}})$. From our discussion in Section IV, this implies that the graph $\mathcal{G}_{\text{constr}}$ remains connected (or, more generally, that the number of connected components of $\mathcal{G}_{\text{constr}}$ does not decrease). Therefore, by Corollary IV.6, the number of connected components of $\mathcal{G}_{\text{sens}}$ also does not decrease.

(iii) If the initial positions of the robots are in $Q_\epsilon$, then the robots will remain forever in $Q_\epsilon$ because $p_i^* \in X_i \subseteq Q_\epsilon$.

(iv) All information required to execute the steps in the algorithm is available to a robot through the sensing model described in Section III. The constraint on the input size, $\|u_i\| \le s_{\max}$, is enforced in step `5:`. □

Finally, we conclude this section by completing our treatment of robots without a common reference frame.

**Remark V.2 (No common reference frame: continued)** Consider a group of robots with visibility sensors and no common reference frame as discussed in Remarks III.1 and IV.3. Because the relative convex hull and the circumcenter of a set transform under changes of coordinate frames in the same way as the constraint set does in equation (3), one can verify that

$$u_i(p_1^0,\dots,p_n^0) = R_i^0 u_i(p_1^i,\dots,p_n^i),$$

where $u_i(p_1^0,\dots,p_n^0)$ is computed with environment $Q^0$ and $u_i(p_1^i,\dots,p_n^i)$ is computed with environment $Q^i$. This equality implies that the robot motion with control $u_i(p_1^0,\dots,p_n^0)$ in equation (1) is identical to the robot motion with control $u_i(p_1^i,\dots,p_n^i)$ in equation (2). □

### B. *Main convergence result*

To state the main results on the correctness of the `Perimeter Minimizing Algorithm`, we require some preliminary notation. First, note that given the positions of the robots $\{p_1,\dots,p_n\}$ at any time instant $t$, the algorithm computes the positions at time instant $t+1$. We can therefore think of the `Perimeter Minimizing Algorithm` as the map $T_{\mathcal{G}_{\text{sens}},\mathcal{G}_{\text{constr}}} : Q_\epsilon^n \to Q_\epsilon^n$. Second, in what follows we will work with tuple of points $P = (p_1,\dots,p_n) \in Q^n$. We let $\mathcal{G}(P)$ denote the proximity graph $\mathcal{G}(\mathcal{P})$ and $\text{rco}(P, Q)$ denote the relative convex hull of the set $\mathcal{P}$ inside $Q$, where $\mathcal{P}$ is the point set given by $\{p_i \mid i \in \{1, \ldots, n\}\}$. Third, we introduce a Lyapunov function that encodes the rendezvous objective. Given an allowable environment $Q$, we recall the notions of relative convex hull and of perimeter from Section II, and we define $V_{\text{perim},Q} : Q^n \to \mathbb{R}_{\geq 0}$ by

$$V_{\text{perim},Q}(P) = \text{perimeter}(\text{rco}(P, Q)).$$

**Lemma V.3 (Properties of Lyapunov function)** *The function $V_{\text{perim},Q}$ has the following properties:*

(i) *$V_{\text{perim},Q}$ is continuous and invariant under permutations of its arguments;*
(ii) *$V_{\text{perim},Q}(P) = 0$ for $P = (p_1, \ldots, p_n)$ if and only if $p_i = p_j$ for all $i, j \in \{1, \ldots, n\}$.*

The technical result on continuity plays an important role in the convergence analysis of the algorithm. Fact (ii) implies that achieving the rendezvous objective is equivalent to making $V_{\text{perim},Q_\epsilon}$ equal to zero. We are now ready to state the main result of the paper.

**Theorem V.4 (Rendezvous is achieved via the `Perimeter Minimizing Algorithm`)** *Let $Q$ and $Q_\epsilon$ be allowable environments. Let $p_1, \ldots, p_n$ be a group of robots with visibility sensors in $Q_\epsilon$. Any trajectory $\{P[t]\}_{t \in \mathbb{Z}_{\geq 0}} \subset Q_\epsilon$ generated by $P[t+1] = T_{\mathcal{G}_{\text{sens}}, \mathcal{G}_{\text{constr}}}(P[t])$ has the following properties:*

(i) *if the locations of two robots belong to the same connected component of $\mathcal{G}_{\text{sens}}$ at $P[t_0]$ for some $t_0$, then they remain in the same connected component of $\mathcal{G}_{\text{sens}}$ at $P[t]$ for all $t \geq t_0$;*
(ii) *$V_{\text{perim},Q_\epsilon}(P[t+1]) \leq V_{\text{perim},Q_\epsilon}(P[t])$; and*
(iii) *the trajectory $\{P[t]\}_{t \in \mathbb{Z}_{\geq 0}}$ converges to a point $P^* \in Q_\epsilon$ such that either $p_i^* = p_j^*$ or $p_i^* \notin \mathcal{S}(p_j^*)$ for all $i, j \in \{1, \ldots, n\}$.*

As a direct consequence of the theorem, note that if the graph $\mathcal{G}_{\text{sens}}$ is connected at any time during the evolution of the system, then all the robots converge to the same location in $Q_\epsilon$.

## VI. Practical implementation issues

In Section V, we designed a provably correct rendezvous algorithm for an ideal model of *point* robots with *perfect sensing and actuation capabilities*. Also, we assumed that it was possible for the robots to operate *synchronously*. However, such an ideal model is not realistic in practical situations. In this section, we investigate, via extensive computer simulations, the effects of deviations from this ideal scenario.

### *A. Nominal experimental set-up*

The computer simulation was written in C++ using the Computational Geometry Algorithmic Library (CGAL) (`http://www.cgal.org`). However, it was found that Boolean operations on polygons using the utilities present in CGAL were not adequate in terms of speed for the purpose of running extensive simulations. Hence, Boolean operations on polygons were performed using the General Polygon Clipping Library (GPC) (`http://www.cs.man.ac.uk/~toby/alan/software/gpc.h`

For the purpose of simulations, the environment considered is a typical floor plan; see Figure 1. Experiments were performed with $20$ robots starting from $10$ randomly generated initial conditions from a uniform distribution. For each initial condition, experiments were repeated $20$ times. The environment size is roughly $80 \times 70$, the step size of a robot is taken as $s_{\max} = 0.5$ and the sensing radius $r = 30$. For simplicity, $\mathcal{G}_{\text{constr}}$ is taken to be the same as $\mathcal{G}_{\text{sens}}$. To utilize the $\epsilon$-robust visibility notion in providing robustness to asynchronism and sensing and control errors, at each time instant, $\epsilon$ is set to be $0.97$ times the $\epsilon$ at the previous time instant. Initially, $\epsilon$ is set equal to $3$. In case a robot approaches a reflex vertex of the environment closely, it reduces its speed. This is done to reduce the risk of collision due to errors in sensing the exact location of the reflex vertex. We now describe the various assumptions we make on the model to simulate the actual implementation on physical robots followed by the respective simulation results. The algorithm performance is then evaluated based on the following three performance measures: (i) the average number of steps taken by the robots to achieve the rendezvous objective; (ii) the fraction of the edges of $\mathcal{G}_{\text{sens}}$ that are preserved at the end of the simulation; and (iii) the number of connected components of $\mathcal{G}_{\text{sens}}$ at the end of the simulation compared with the number of connected components at the initial time.

### *B. Robustness against asynchronism, sensing and control noise, and finite-size disk robot models*

(A1) *Asynchronism:* The robots operate asynchronously, i.e., do not share a common processor clock. All the robots start operating at the same time. Each robot's clock speed is a random number uniformly distributed on the interval $[0.9, 1]$. At integral multiples of its clock speed, a robot wakes up, senses the positions of other robots within its sensing region and takes a step according to the `Perimeter Minimizing Algorithm`. Note that at the time when any given robot wakes up, there may be other robots that are moving. No sensing and control errors were introduced in the model. It is observed that under this asynchronous implementation, the performance of the algorithm is very similar to the synchronous implementation; see Figure 12. In the subsequent implementations, we assume the robots to operate synchronously.

(A2) *Distance error in sensing and control:* The visibility sensors measure the relative distance of another object according to the following multiplicative noise model. If $d_{\text{act}}$ is the actual distance to the object, then the measured distance is given by $(1 + e_{\text{dist}})d_{\text{act}}$, where $e_{\text{dist}}$ is a random variable uniformly distributed in the interval $[-0.1, 0.1]$. The objects to which distances are measured are other robots in the sensing range and the boundary of the environment. For simplicity, instead of measuring a sequence of points along the boundary, as a real range sensor does, we assume that only the vertices of the environment are measured and the sensed region is reconstructed from that information. The sensor error, therefore, occurs in the measurement of other robots and environment vertices. The actuators moving the robots are also subject to a multiplicative noise distance model with the same error parameter. The results are shown in Figure 13. It is observed that only in 1 out of the 10 initial conditions does the number of connected components of $\mathcal{G}_{\text{sens}}$ increase as compared to the number of connected components of $\mathcal{G}_{\text{sens}}$ initially. In all cases, the fraction of edges preserved is almost equal to 1. Also, in 7 out of 10 cases, the performance is almost identical to the synchronized implementation with no noise. In the subsequent simulations, we assume the robots to operate synchronously with no distance error in sensing and control.

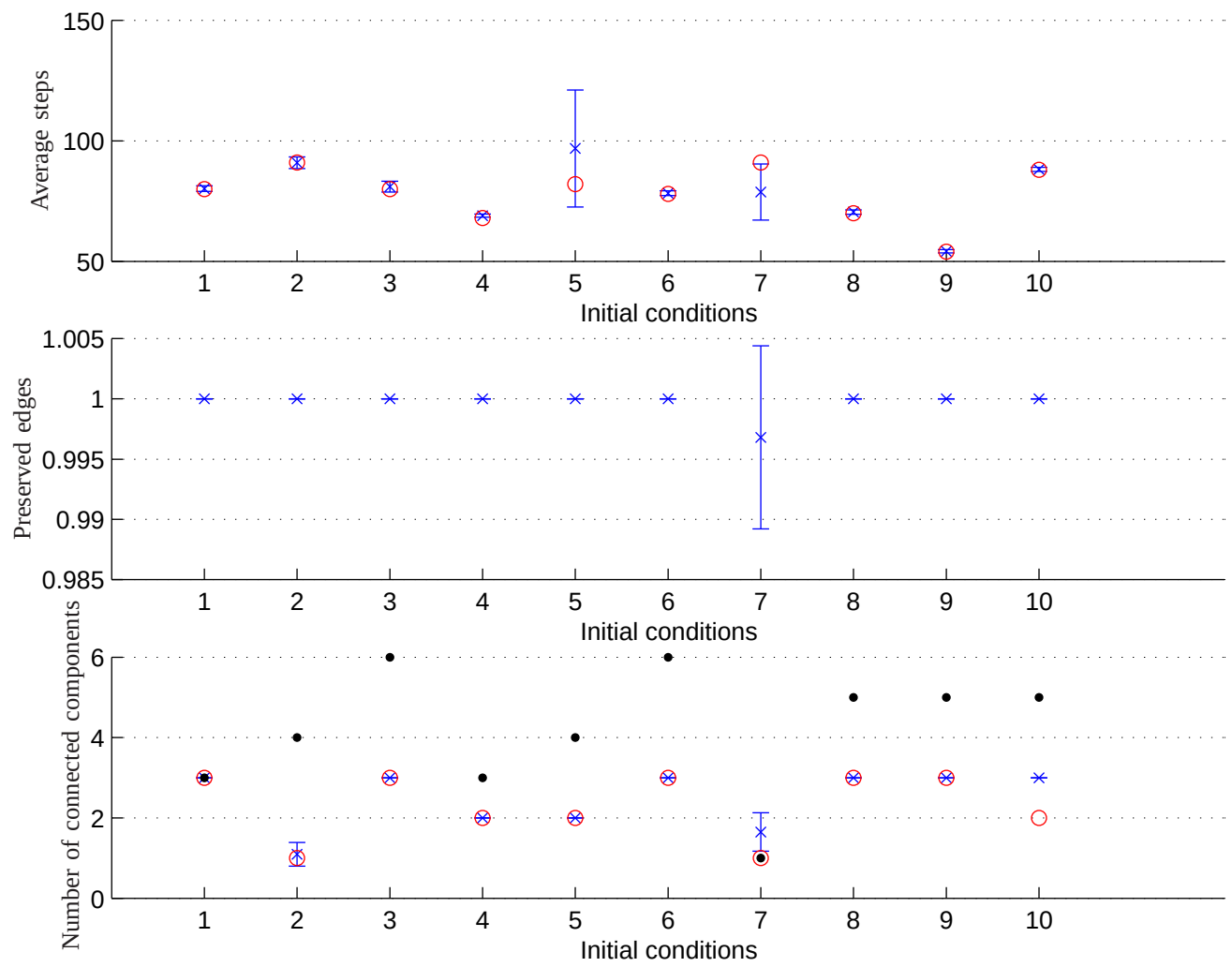

Fig. 12. Computer simulation results with asynchronism. The top figure represents the average number of steps taken per robot for convergence(crosses). Also shown is the number of steps taken by each robot in synchronous implementation (red circle). The center figure shows the fraction of the edges of the initial sensing graph that are preserved till the end. The bottom figure shows the number of connected components of the sensing graph initially (small black discs) and at the end of asynchronous (blue crosses) and synchronous (red circle) implementations. The blue crosses in the figure denote the mean of the observed quantities and the vertical bars denote standard deviation.

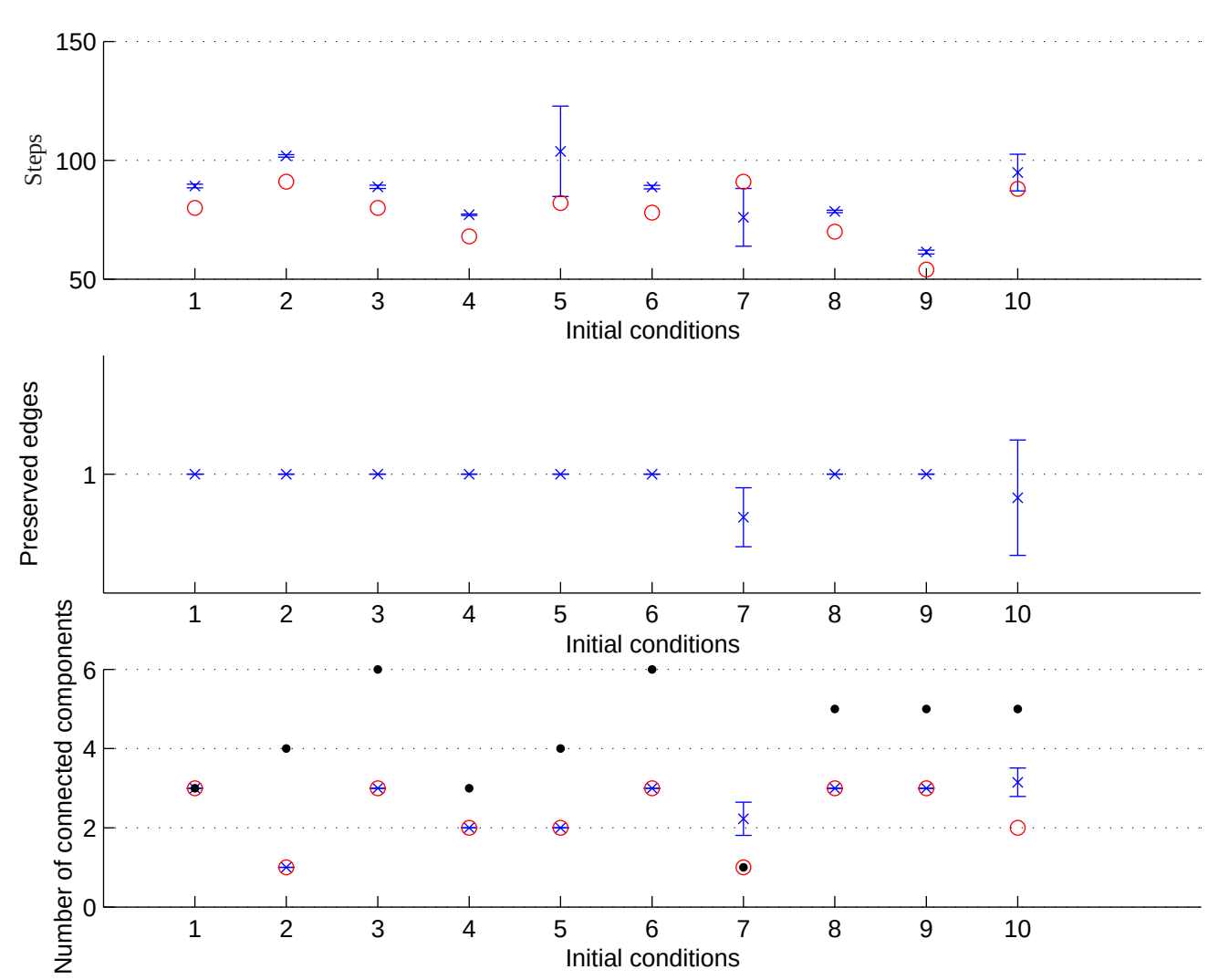

Fig. 13. Computer simulation results with distance error and no directional error in sensing and control.The meaning of the quantities is as in Figure 12.

(A3) *Direction error in sensing and control:* The visibility sensors measure the relative angular location of another object according to the following additive noise model. If $\theta_{\text{act}}$ is the actual angular location of any object in the local reference frame of a robot, the measured angular location is given by $\theta_{\text{act}} + e_\theta$, where $e_\theta$ is a random variable uniformly distributed in the interval $[-5, 5]$. As before, the actuators moving the robots are also subject to an additive noise directional model with the same error parameter. The results are shown in Figure 14.

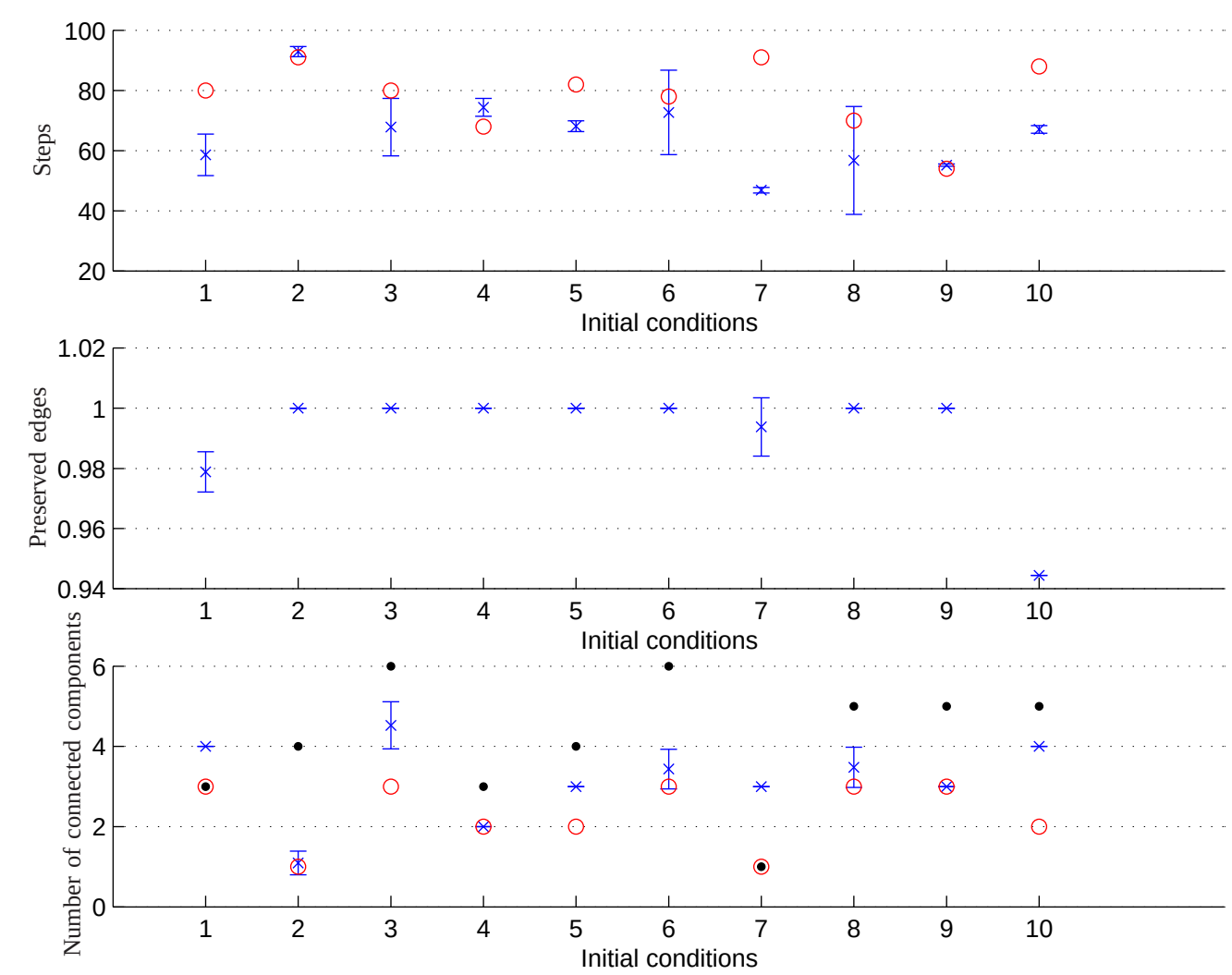

Fig. 14. Computer simulation results with direction error and no distance error in sensing and control. The meaning of the quantities is as in Figure 12.

It is observed that the algorithm no longer performs similar to the synchronized implementation with no noise. Thus, it is more sensitive to directional errors than distance errors in sensing and control. However, almost all of the edges of $\mathcal{G}_{\text{sens}}$ are preserved for all the initial conditions. Also, in n 9 out of the 10 initial conditions the number of connected components of $\mathcal{G}_{\text{sens}}$ corresponding decreases during the course of evolution of the group. In the following simulations, no asynchronism or sensing and control errors are assumed.

(A4) *Disk robot model:* The robots are assumed to be disks of radius 0.2, but do not obstruct the views of others. During any step, a robot moves a distance of at most $s_{\max} = 0.5$. The next position of the center of the robot, therefore, lies in a *motion disc* of radius 0.7 centered at the center of the robot; see the red and green discs in Figure 15. A *colliding neighbor* of a robot $i$ is any neighbor $j$ according to $\mathcal{G}_{\text{sens}}$ such that the motion discs of $i$ and $j$ intersect and that the motion disc of $j$ intersects the physical disc of $i$ on the path between $i$ and its next point. If a robot has no colliding neighbors, then its motion is according to Perimeter Minimizing Algorithm. If on the other hand a robot has exactly one colliding neighbor, then it tries to swerve around it while reducing its speed. Finally, if the robot has more than two colliding neighbors then it stays at the current location. To ensure free movement of the robots inside the environment, $\epsilon$ is not allowed to fall below $0.2$, i.e., the radius of a robot disc. Simulations were performed without any asynchronism or sensing and control errors for the 10 initial conditions of the robots as in the previous experiments.The terminating condition for the simulations is that robots belonging to each connected component of $\mathcal{G}_{\text{sens}}$ form a "cohesive" group[6]; see Figure 15. For all initial conditions, the number of cohesive groups in the final configurations is equal to the number of connected components of $\mathcal{G}_{\text{sens}}$ when the simulations are performed assuming point robots. Thus, the algorithm yields the same performance if a disk robot model is assumed instead of a point robot model.

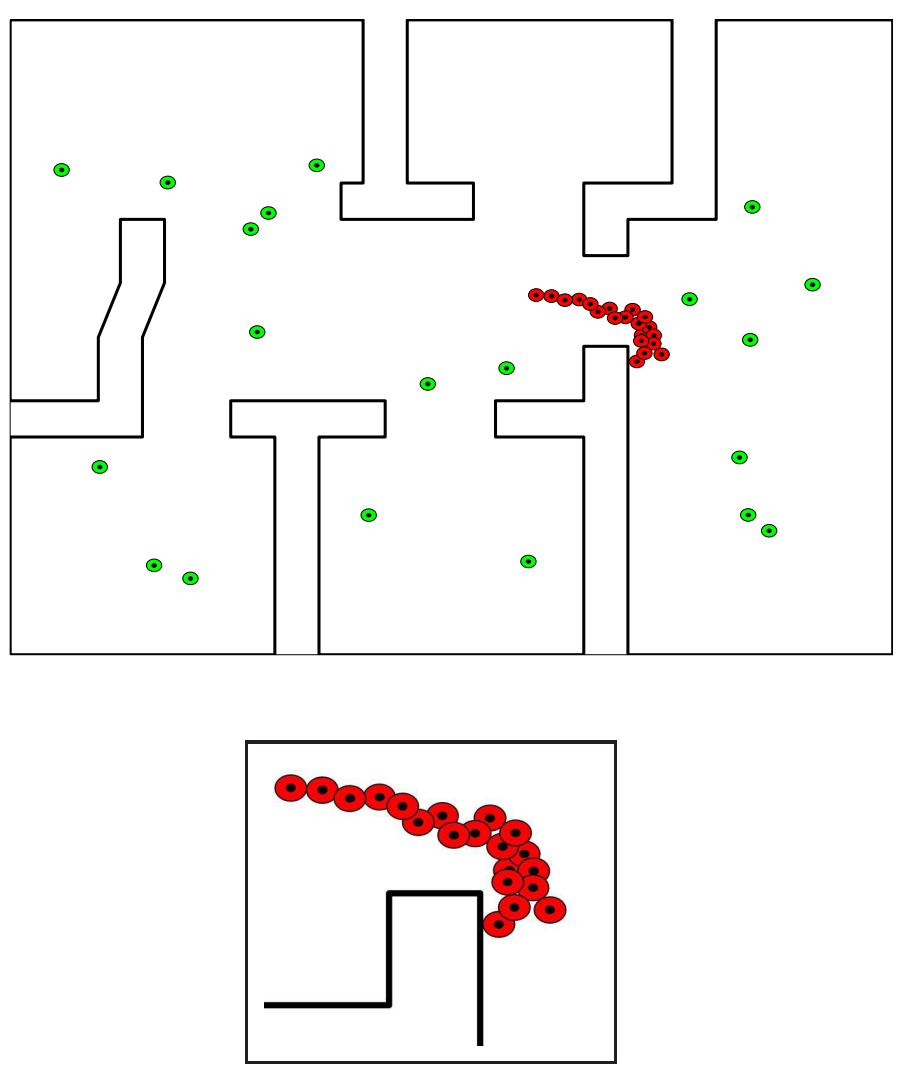

Fig. 15. Computer simulation results with no asynchronism and no sensing and control errors. The black and red discs denote the robots and their motion discs respectively. The initial position of the robots correspond to initial condition 2 in Figures 12, 13 and 14 and are shown by the small black discs scattered over the environment with the green discs denoting their motion discs. The robots converge to positions corresponding to a single cohesive group part of the same component of $\mathcal{G}_{\text{sens}}$.

Thus, we see that the Perimeter Minimizing Algorithm is robust to various deviations from the ideal scenario. The magnitudes of $e_{\text{dist}}$ and $e_\theta$ are in line with the state-of-the-art. Finally, in the next section we analyze the computation complexity of the algorithm.

### C. Computation complexity with finite resolution sensing

In addition to the issues that might arise in a practical implementation of the Perimeter Minimizing Algorithm, another important consideration is the time taken for a robot to complete each step of the algorithm. This is dependent on the computational complexity of the algorithm, that we characterize in the following.

A real visibility sensor, e.g., a range scanner, will sense the position of other robots and the boundary of the environment with some finite resolution; in particular, the boundary of the sensing region will be described by a set of points. It is reasonable to assume that the cardinality of this set of points is bounded, say by $M$, for all robots, irrespective of the shape of the environment and the location of the robot in it. For example, if a laser range sensor is used to measure the distance to the boundary and a measurement is taken at intervals of one degree, then $M$ is equal to $360$.

**Proposition VI.1 (Computational complexity)** *Let $Q$ be any allowable environment. Let $M$ be the resolution of the visibility sensor located at any robot in $Q$. Then the following statements are true:*

(i) *The computation complexity of the* Constraint Set Generator Algorithm *is $O(\kappa M)$;*
(ii) *The computation complexity of the* Perimeter Minimizing Algorithm *is $\tau(M) + O(M^3 \log M)$;*
(iii) *If $\mathcal{G}_{\text{constr}} = \mathcal{G}_{\text{sens}}$, then the computation complexity of the* Perimeter Minimizing Algorithm *is $O(M^2 \log M)$,*

*where $\tau(M)$ is time taken for the computation of $\mathcal{N}_{i,\mathcal{G}_{\text{constr}}}$ given the set $\mathcal{N}_{i,\mathcal{G}_{\text{sens}}}$ assuming $|\mathcal{N}_{i,\mathcal{G}_{\text{sens}}}| \leq M$, and $\kappa$ is the number of strict concavities of $Q$.*

As discussed in Section IV-B, if $\mathcal{G}_{\text{constr}} = \mathcal{G}_{\text{lc},\mathcal{G}_{\text{sens}}}$, then the computation of $\mathcal{G}_{\text{constr}}$ from $\mathcal{G}_{\text{sens}}$ can be performed using efficient polynomial time heuristics. The time $\tau(M)$ above depends on the specific heuristic used. Thus, we see that the running time of each step of the Perimeter Minimizing Algorithm is polynomial in the number of data points obtained by the visibility sensor.

This analysis helps us assess whether it is feasible to implement this algorithm on an actual robot without demanding an unreasonable computational power.

## VII. Conclusions

In this paper, we present a provably correct discrete-time synchronous Perimeter Minimizing Algorithm algorithm for rendezvous of robots equipped with visibility sensors in a nonconvex environment. The algorithm builds on a novel solution to the connectivity maintenance problem also proposed in this paper. The performance of the algorithm under asynchronous operation of the robots, presence of noise in sensing and control, and nontrivial robot dimension is investigated and found to be quite satisfactory. The computational complexity of the algorithm under the assumption of finite sensing resolution is also investigated.

## VIII. Acknowledgments

This material is based upon work supported in part by AFOSR MURI Award F49620-02-1-0325, NSF Award CMS-0626457, NSF Award IIS-0525543, and NSF CAREER Award

[6] For each connected component of $\mathcal{G}_{\text{sens}}$, the graph having nodes as the robot locations and with an edge between two nodes whenever the corresponding motion discs of the robots intersect is connected

ECS-0546871. The authors thank Prof. Jana Košecká for an early inspiring discussion.

## APPENDIX I
## PROOFS OF RESULTS IN SECTION IV

*Proof of Lemma IV.1:* Let $d = \mathrm{dist}(v_1, [p_i, p_j]) = \mathrm{dist}(v_2, [p_i, p_j])$ and let $L = \{p \in \mathbb{R}^2 \mid \mathrm{dist}(p, [p_i, p_j]) \leq d\}$. Then $v_1, v_2 \in \partial L$; see Figure 16. We now prove our result by contradiction. Let $v_1 \notin H_{Q_\epsilon}(v_2)$. Then $v_1 \in \partial L \setminus H_{Q_\epsilon}(v_2)$ as shown in Figure 16. Since $v_1$ is visible from $p_i$, the boundary $\partial Q_\epsilon$ must intersect $\partial L$ at a point $v$ in between $v_2$ and $v_1$, as illustrated in Figure 16. But this means that there exists a point $v^*$ belonging to the concavity containing $v_2$ that is strictly closer to the segment $[p_i, p_j]$ than $v_2$; this is a contradiction. ■

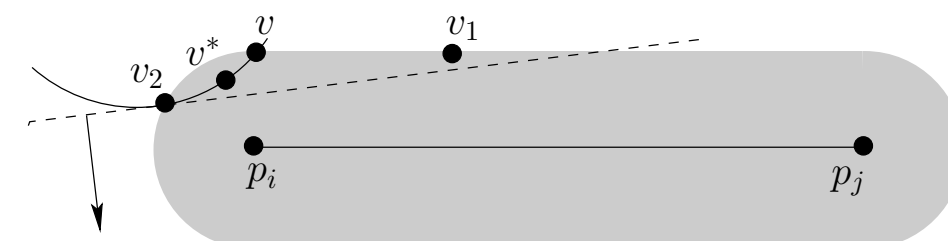

Fig. 16. The shaded region represents $L$. The solid curve passing through $v_2$ represents a portion of the boundary of $Q_\epsilon$. The dashed line is the boundary of $H_{Q_\epsilon}(v_2)$ which is along the tangent to $\partial Q_\epsilon$ at $v_2$. The interior of $H_{Q_\epsilon}(v_2)$ is in the direction of the arrow.

*Proof of Proposition IV.2:* We first prove statement (i). Note that initially $\mathcal{C}_{\text{temp}} = \mathcal{V}(p_i, \epsilon) \cap B(\frac{p_i+p_j}{2}, \frac{r}{2})$. Therefore $\partial\mathcal{C}_{\text{temp}}$ has at most $\kappa$ concavities. This is because the concavities of $\partial\mathcal{C}_{\text{temp}}$ are induced by the concavities of $\partial Q_\epsilon$, and the number of concavities of $\partial Q_\epsilon$ is the same as the number of concavities of $\partial Q$. Now, by construction, the number of concavities in $\partial(\mathcal{C}_{\text{temp}} \cap H_{Q_\epsilon}(v))$ is strictly less than the number of concavities of $\partial\mathcal{C}_{\text{temp}}$. It follows then that the `Constraint Set Generator Algorithm` terminates in at most $\kappa$ steps. This completes the proof of statement (i).

Now note that $[p_i, p_j] \subseteq \mathcal{C}_Q(p_i, p_j)$. Hence $\mathcal{C}_Q(p_i, p_j)$ is always non-empty. Notice that $\mathcal{C}_Q(p_i, p_j)$ can be written in the following way

$$\mathcal{C}_Q(p_i, p_j) = \mathcal{V}(p_i, \epsilon) \cap B(\frac{p_i + p_j}{2}, \frac{r}{2}) \\ \cap H_{Q_\epsilon}(v_{i,j,1}) \ldots \cap H_{Q_\epsilon}(v_{i,j,n_{ij}}), \quad (6)$$

where $\mathrm{dist}(v_{i,j,1}, [p_i, p_j]) \leq \ldots \leq \mathrm{dist}(v_{i,j,n_{ij}}, [p_i, p_j])$. The $v_{i,j,k}$'s are computed according to step `3:` in the `Constraint Set Generator Algorithm`. Therefore, $\mathcal{C}_Q(p_i, p_j)$ is the result of the intersection of a finite number of closed sets, and therefore is closed as well. Also $\mathcal{C}_Q(p_i, p_j) \subseteq Q_\epsilon$ and is thus bounded. Therefore, it is compact. Finally, the `Constraint Set Generator Algorithm` terminates when $\mathcal{C}_{\text{temp}}$ has no concavities, or in other words, when $\mathcal{C}_{\text{temp}}$ is convex. This proves statement (ii).

To prove statement (iii), let us write $\mathcal{C}_Q(p_j, p_i)$ as

$$\mathcal{C}_Q(p_j, p_i) = \mathcal{V}(p_i, \epsilon) \cap B(\frac{p_i + p_j}{2}, \frac{r}{2}) \\ \cap H_{Q_\epsilon}(v_{j,i,1}) \ldots \cap H_{Q_\epsilon}(v_{j,i,n_{ji}}),$$

in the same way as (6). We first claim that $n_{ij} = n_{ji}$ and $\{v_{i,j,1}, \ldots, v_{i,j,n_{ij}}\} = \{v_{j,i,1}, \ldots, v_{j,i,n_{ji}}\}$. Before proving this claim, let us assume it is true and see where it leads. Because of this assumption, the expressions for $\mathcal{C}_Q(p_i, p_j)$ and $\mathcal{C}_Q(p_j, p_i)$ would differ only due to the terms $\mathcal{V}(p_i, \epsilon)$ and $\mathcal{V}(p_j, \epsilon)$. Now, let $q \in \mathcal{C}_Q(p_i, p_j)$. Because $\mathcal{C}_Q(p_i, p_j)$ is a convex subset of $Q_\epsilon$ containing $p_j$ and $q$, the line segment $[p_j, q] \subseteq \mathcal{C}_Q(p_i, p_j) \subseteq Q_\epsilon$. Hence, $q \in \mathcal{V}(p_j, \epsilon)$. This in turn implies that $q \in \mathcal{C}_Q(p_j, p_i)$. Therefore, we have $\mathcal{C}_Q(p_i, p_j) \subseteq \mathcal{C}_Q(p_j, p_i)$. The opposite inclusion can be shown to be true in a similar fashion. Thus, we have $\mathcal{C}_Q(p_i, p_j) = \mathcal{C}_Q(p_j, p_i)$.

We now prove that $n_{ij} = n_{ji}$ and $\{v_{i,j,1}, \ldots, v_{i,j,n_{ij}}\} = \{v_{j,i,1}, \ldots, v_{j,i,n_{ji}}\}$. Note that $v_{i,j,1}$ is the strictly concave point nearest to $[p_i, p_j]$ belonging to $\mathcal{V}(p_i, \epsilon) \cap B(\frac{p_i+p_j}{2}, \frac{r}{2})$. In general, there can be more than one nearest strictly concave point, say $v_{i,j,1}, \dots, v_{i,j,s_{ij}}$. Let $d = \mathrm{dist}(v_{i,j,1}, [p_i, p_j])$ and $L = \{p' \in \mathcal{C}_{\text{temp}} | \mathrm{dist}(p', [p_i, p_j]) \leq d\}$, where $\mathcal{C}_{\text{temp}} = \mathcal{V}(p_i, \epsilon) \cap B(\frac{p_i+p_j}{2}, \frac{r}{2})$. Note that $L$ is convex since there cannot be any strictly concave points of $\partial\mathcal{C}_{\text{temp}}$ in the interior of $L$. Since $p_j, v_{i,j,1} \in L$ then $[v_{i,j,1}, p_j] \subseteq L \subseteq Q_\epsilon$. Therefore $v_{i,j,1} \in \mathcal{V}(p_j, \epsilon)$ and hence $v_{i,j,1} \in \mathcal{V}(p_j, \epsilon) \cap B(\frac{p_i+p_j}{2}, \frac{r}{2})$. Therefore, $v_{i,j,1}, \dots, v_{i,j,s_{ij}} \in \mathcal{V}(p_j, \epsilon) \cap B(\frac{p_i+p_j}{2}, \frac{r}{2})$. Similarly, if $v_{j,i,1}, \dots, v_{j,i,s_{ji}}$ are the strictly concave points nearest to $[p_i, p_j]$ belonging to $\mathcal{V}(p_j, \epsilon) \cap B(\frac{p_i+p_j}{2}, \frac{r}{2})$, then $v_{j,i,1}, \dots, v_{j,i,s_{ji}} \in \mathcal{V}(p_i, \epsilon) \cap B(\frac{p_i+p_j}{2}, \frac{r}{2})$. Therefore $\mathrm{dist}(v_{i,j,1}, [p_i, p_j]) = \mathrm{dist}(v_{j,i,1}, [p_i, p_j])$ and $s_{ij} = s_{ji}$. This implies that $v_{j,i,k} \in \{v_{i,j,1}, \dots, v_{i,j,s_{ij}}\}$ for all $k \in \{1, \dots, s_{ji}\}$. Proceeding recursively, we get that $n_{ij} = n_{ji}$ and $\{v_{i,j,1}, \dots, v_{i,j,n_{ij}}\} = \{v_{j,i,1}, \dots, v_{j,i,n_{ji}}\}$. This completes the proof of statement (iii).

We now prove statement (iv). First, let us write

$$\mathcal{C}_Q(p_i, p_j) = \mathcal{V}(p_i, \epsilon) \cap B(\frac{p_i + p_j}{2}, \frac{r}{2}) \bigcap \cap_{k=1}^{n_{ji}} H_{Q_\epsilon}(v_{j,i,k}).$$

Since $Q$ is bounded, there exists $p_0 \in Q$ such that $\mathcal{C}_Q(p_i, \overline{p_j}) \subseteq Q \subseteq B(p_0, \mathrm{diam}(Q))$. Therefore, we can write

$$\mathcal{C}_Q(p_i, p_j) = \mathcal{V}(p_i, \epsilon) \cap B(\frac{p_i + p_j}{2}, \frac{r}{2}) \bigcap \cap_{k=1}^{n_{ji}} \left(H_{Q_\epsilon}(v_{j,i,k}) \cap B(p_0, \mathrm{diam}(Q))\right).$$

The map $p_i \to \mathcal{V}(p_i, \epsilon)$ is upper semicontinuous at any point $p_i \in Q_\epsilon$ because there exists a neighborhood around any point $p_i$ where the visibility set can only shrink. Also, the map $p_i \to \mathcal{V}(p_i, \epsilon)$ has a compact range. Thus the notions of upper semicontinuity and closedness are equivalent and, hence, $p_i \to \mathcal{V}(p_i, \epsilon)$ is closed at any point $p_i \in Q_\epsilon$. The second term is the closed ball of radius $\frac{r}{2}$ centered at the point $\frac{p_i+p_j}{2}$ and clearly depends continuously on $p_i$.

Let $a_1, \dots, a_\kappa$ be the strict concavities of $Q_\epsilon$. Now $v_{i,j,k}$ belongs to exactly one strict concavity, say $a_{i_k}$. Given $a_{i_k}$, $v_{i,j,k}$ is a function of $(p_i, p_j)$. We can write $v_{i,j,k} := v_{i,j,k}(p_i, p_j) = \mathrm{argmin}\{\mathrm{dist}(v, [p_i, p_j]) \mid v \in a_{i_k}\}$. Since $a_{i_k}$ is a continuous curve, it is clear that $v_{i,j,k}$ is a continuous function of $(p_i, p_j)$. We now show that $H_{Q_\epsilon}(v_{i,j,k}(p_i, p_j)) \cap B(p_0, \mathrm{diam}(Q))$ is in turn a set-valued map continuous with respect to $(p_i, p_j)$. Since $v_{i,j,k}$ is a continuous function of $(p_i, p_j)$, from Propositions 1 and 8 in [17], it suffices to show that $H_{Q_\epsilon}(v) \cap B(p_0, \mathrm{diam}(Q))$ is continuous at any point $v$ over the domain $a_{i_k}$. This is equivalent to showing that $H_{Q_\epsilon}(v) \cap B(p_0, \mathrm{diam}(Q))$ is closed as well as open at any point $v \in a_{i_k}$. Without loss of generality, let us assume that $\partial H_{Q_\epsilon}(v)$ is parallel to the $x$ axis as shown in Figure 17. Let us first show that $H_{Q_\epsilon}(v) \cap B(p_0, \mathrm{diam}(Q))$ is open at any point $v \in a_{i_k}$. Let $\{v_l\}$ be a sequence in $a_{i_k}$ such that $v_l \to v$, and let $q$ be any point on $\partial H_{Q_\epsilon}(v)$. Let $q_l$ be the point on $\partial H_{Q_\epsilon}(v_l)$ that is either vertically above or below $q$. Now $\|q - q_l\| \leq |s(v_l)| \,\mathrm{diam}(Q)$, where $s(v_l)$ is the slope of the line defining the boundary of $H_{Q_\epsilon}(v_l)$. Since $a_{i_k}$ is continuously differentiable, as $v_l \to v$, we have that $s(v_l) \to s(v) = 0$. Thus, $q_l \to q$. Hence, $H_{Q_\epsilon} \cap B(p_0, \mathrm{diam}(Q))$ is open at any point $v \in a_{i_k}$. We now show that $H_{Q_\epsilon} \cap B(p_0, \mathrm{diam}(Q))$ is closed at any point $v \in a_{i_k}$. As before, let $\{v_l\}$ be a sequence in $a_{i_k}$ such that $v_l \to v$. Let $q_l \in H_{Q_\epsilon}(v_l) \cap B(p_0, \mathrm{diam}(Q))$ and let $q_l \to q$. But $\mathrm{dist}(q_l, H_{Q_\epsilon}(v) \cap B(p_0, \mathrm{diam}(Q))) \leq |s(v_l)| \,\mathrm{diam}(Q)$ and $s(v_l) \to s(v) = 0$ as $v_l \to v$. Therefore, $\mathrm{dist}(q_l, H_{Q_\epsilon}(v) \cap B(p_0, \mathrm{diam}(Q))) \to 0$. Since $q_l \to q$ and the distance of a point to the convex set $H_{Q_\epsilon}(v) \cap B(p_0, \mathrm{diam}(Q))$ is a continuous function, we deduce $\mathrm{dist}(q_l, H_{Q_\epsilon}(v) \cap B(p_0, \mathrm{diam}(Q))) \to \mathrm{dist}(q, H_{Q_\epsilon}(v) \cap B(p_0, \mathrm{diam}(Q))) = 0$. Hence $q \in H_{Q_\epsilon}(v) \cap B(p_0, \mathrm{diam}(Q)))$. Therefore $H_{Q_\epsilon}(v) \cap B(p_0, \mathrm{diam}(Q)))$ is a closed map and, in turn, is continuous over the domain $a_{i_k}$. This implies that $H_{Q_\epsilon}(v_{i,j,k}(p_i, p_j)) \cap B(p_0, \mathrm{diam}(Q)))$ is continuous at $p_i$ for fixed $a_{i_k}$.

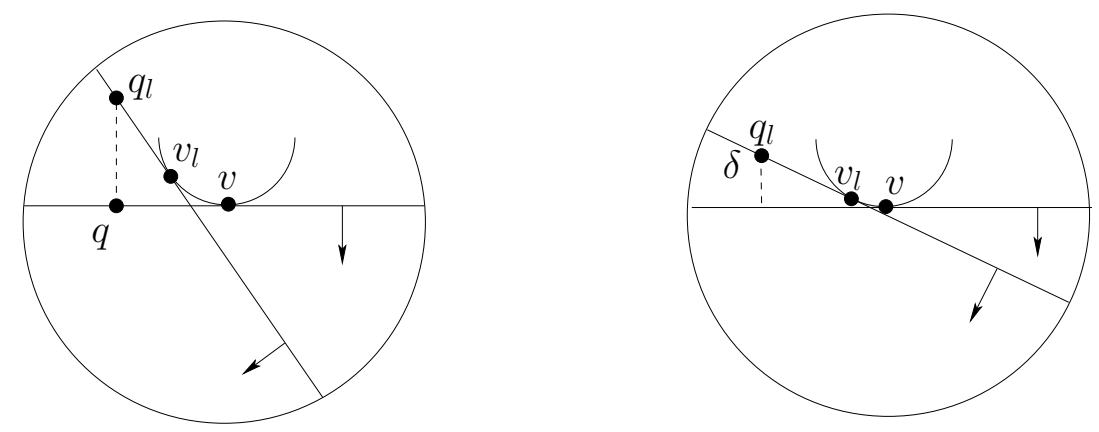

Fig. 17. Illustration of various notions used in proving that $H_{Q_\epsilon}(v) \cap B(p_0, \mathrm{diam}(Q))$ is continuous at any point $v$ over the domain $a_{i_k}$. The circle represents the boundary of the set $B(p_0, \mathrm{diam}(Q))$. The arc represents $a_{i_k}$. The lines tangent to $a_{i_k}$ at $v$ and $v_l$ are the boundaries of the half-planes $H_{Q_\epsilon}(v)$ and $H_{Q_\epsilon}(v_l)$ respectively. The interior of the half-planes is in the direction of the arrows. In the figure on the right, $\delta$ represents $\mathrm{dist}(q_l, H_{Q_\epsilon}(v) \cap B(p_0, \mathrm{diam}(Q)))$.

Now, let $(p_i^k, p_j^k) \to p_i$ and let $q^k \to q$ where $q^k \in \mathcal{C}_Q(p_i^k, p_j^k)$. We need to show that $q \in \mathcal{C}_Q(p_i, p_j)$. Since $q^k \in \mathcal{C}_Q(p_i^k, p_j^k)$, we have that $q^k \in \mathcal{V}(p_i^k, \epsilon)$. Since the map $p_i \to \mathcal{V}(p_i, \epsilon)$ is closed, we have that $q \in \mathcal{V}(p_i, \epsilon)$. Also, $q^k \in B(\frac{p_i^k + p_j^k}{2}, \frac{r}{2})$. Since the map $(p_i, p_j) \to B(\frac{p_i+p_j}{2}, \frac{r}{2})$ is continuous, it follows that $q \in B(\frac{p_i+p_j}{2}, \frac{r}{2})$. Therefore, what remains to be shown is that $q \in H_{Q_\epsilon}(v_{i,j,l}(p_i, p_j))$ for all $l \in \{1, \dots, n_{ij}\}$. In what follows, for any points $(p_i, p_j)$, $(p_i^k, p_j^k)$ and $(p_i', p_j')$, let $v_{i,j,l}(p_i, p_j)$, $v_{i,j,l}(p_i^k, p_j^k)$ and $v_{i,j,l}(p_i', p_j')$ belong to the concavities $a_{i_l}$, $a_{i_l}^k$ and $a_{i_l}'$ respectively and with $l$ belonging to $\{1, \cdots, n_{ij}\}$, $\{1, \cdots, n_{ij}^k\}$ and $\{1, \cdots, n_{ij}'\}$ respectively. Since $q^k \in H_{Q_\epsilon}(v_{i,j,l}(p_i^k, p_j^k))$, if there exists $k_0$ such that for all $k \geq k_0$, we have that $a_{i_l}^k \in \{a_{i_1}, \cdots, a_{i_{n_{ij}}}\}$, then we are done. Equivalently, it suffices to show that if $a$ is any strict concavity such that $a \notin \{a_{i_1}, \cdots, a_{i_{n_{ij}}}\}$ then there exists a neighborhood $\mathcal{B}$ around $p_i$ such that for all $p_i' \in \mathcal{B}$, $a \notin \{a_{i_1}', \cdots, a_{i_{n_{ij}'}}'\}$. But if $a \notin \{a_{i_1}, \cdots, a_{i_{n_{ij}}}\}$, then $a$ is in the exterior of at least one of $\mathcal{V}(p_i, \epsilon)$, $B(\frac{p_i+p_j}{2}, \frac{r}{2})$ and $H_{Q_\epsilon}(v_{i,j,l}(p_i, p_j))$. The existence of $\mathcal{B}$ now follows from the fact that the maps $p_i \to \mathcal{V}(p_i, \epsilon)$, $p_i \to B(\frac{p_i+p_j}{2}, \frac{r}{2})$ and $p_i \to H_{Q_\epsilon}(v_{i,j,l}(p_i)) \cap B(p_0, \mathrm{diam}\, Q)$ are all closed. ■

*Proof of Lemma IV.5:* The second inclusion in fact (i) is a direct consequence of the definition of $\mathcal{G}_{\text{lc},\mathcal{G}}$. We prove now that $\mathcal{G}_{\text{EMST},\mathcal{G}} \subseteq \mathcal{G}_{\text{lc},\mathcal{G}}$ by contradiction. Let $\mathcal{P}$ be any point set and assume, without loss of generality, that $\mathcal{G}(\mathcal{P})$ is connected (otherwise, the same reasoning carries over for each connected component of $\mathcal{G}(\mathcal{P})$). For simplicity, further assume that the distances $\|p_k - p_l\|$, $k, l \in \{1, \dots, n\}$, $k \neq l$ are all distinct. This ensures that there is a unique Euclidean Minimum Spanning Tree of $\mathcal{G}(\mathcal{P})$. If this is not the case, the same reasoning exposed here carries through for each Euclidean Minimum Spanning Tree associated with $\mathcal{G}(\mathcal{P})$. Let $(p_i, p_j) \in \mathcal{E}_{\mathcal{G}_{\text{EMST},\mathcal{G}}}(\mathcal{P})$ and $(p_i, p_j) \notin \mathcal{E}_{\mathcal{G}_{\text{lc},\mathcal{G}}}(\mathcal{P})$. Since necessarily $(p_i, p_j) \in \mathcal{E}_{\mathcal{G}}(\mathcal{P})$, the latter implies that there exists a maximal clique $\mathcal{P}'$ of the edge $(p_i, p_j)$ in $\mathcal{G}$ such that $(p_i, p_j) \notin \mathcal{E}_{\mathcal{G}_{\text{EMST}}}(\mathcal{P}')$. If we remove the edge $(p_i, p_j)$ from $\mathcal{G}_{\text{EMST},\mathcal{G}}(\mathcal{P})$, the tree becomes disconnected into two connected components $T_1$ and $T_2$, with $p_i \in T_1$ and $p_j \in T_2$. Now, there must exist an edge $e \in \mathcal{E}_{\mathcal{G}_{\text{EMST}}}(\mathcal{P}')$ with one vertex in $T_1$ and the other vertex in $T_2$ and with length strictly less than $\|p_i - p_j\|$. To see this, let $\{e_1, \dots, e_d\}$ be the edges of $\mathcal{G}_{\text{EMST}}(\mathcal{P}')$ obtained in incremental order by running Prim's algorithm (e.g., see [18]) starting from the vertex $p_i$. Because $p_i$ is in $T_1$ and $p_j$ is in $T_2$, there must exist at least an edge in $\{e_1, \dots, e_d\}$ with one vertex in $T_1$ and the other vertex in $T_2$. Let $s \in \{1, \dots, d\}$ be such that $e_s$ is the first edge having one vertex in $T_1$ and another vertex in $T_2$. Since $(p_i, p_j) \notin \mathcal{E}_{\mathcal{G}_{\text{EMST}}}(\mathcal{P}')$, according to Prim's algorithm, the length of $e_s$ must be strictly less than $\|p_i - p_j\|$ (otherwise, the edge $(p_i, p_j)$ will be part of the Euclidean Minimum Spanning Tree of $\mathcal{P}'$). If we add the edge $e_s$ to the set of edges of $T_1 \cup T_2$, the resulting graph $G$ is acyclic, connected and contains all the vertices $\mathcal{P}$, i.e., $G$ is a spanning tree. Moreover, since the length of $e_s$ is strictly less than $\|p_i - p_j\|$ and $T_1$ and $T_2$ are induced subgraphs of $\mathcal{G}_{\text{EMST},\mathcal{G}}(\mathcal{P})$, we conclude that $G$ has shorter length than $\mathcal{G}_{\text{EMST},\mathcal{G}}(\mathcal{P})$, which is a contradiction. Fact (ii) is a consequence of fact (i). ■

## Appendix II
## Additional analysis results and proof of Theorem V.4

In this section we provide some key technical results that help establish Theorem V.4. We first present the technical results on which the proof depends.

**Lemma II.1 (Properties of the relative convex hull)** *For any allowable $Q, Q_\epsilon$, the following statements hold:*

(i) *if $\mathcal{P}_1, \mathcal{P}_2 \subseteq Q$, then $\text{rco}(\mathcal{P}_1, Q) \subseteq \text{rco}(\mathcal{P}_1 \cup \mathcal{P}_2, Q)$;*

(ii) *if $\text{rco}(P', Q)$ is a strict subset of $\text{rco}(P'', Q)$, then $V_{\text{perim},Q}(P') < V_{\text{perim},Q}(P'')$;*

(iii) *if $\mathcal{P} \subset Q_\epsilon$ and $\mathcal{G}(\mathcal{P}) \subseteq \mathcal{G}_{\text{sens}}(\mathcal{P})$, then for $p_i \in \mathcal{P}$, we have that $\{p_i\} \cap \text{Ve}(\text{rco}(\mathcal{P}, Q_\epsilon)) \subseteq \text{Ve}(\text{co}(\mathcal{N}_{i,\mathcal{G}}))$.*

*Proof:* Statements (i) and (ii) are obvious from the definitions of the relative convex hull and its perimeter. To prove statement (iii), assume $p_i \notin \text{Ve}(\text{co}(\mathcal{N}_{i,\mathcal{G}}))$. Then, $p_i$ belongs either to $\partial\,\text{co}(\mathcal{N}_{i,\mathcal{G}}) \setminus \text{Ve}(\text{co}(\mathcal{N}_{i,\mathcal{G}}))$ or to the interior of $\text{co}(\mathcal{N}_{i,\mathcal{G}})$. In the former case, $p_i \in [p_k, p_l]$ where $p_k, p_l \in \text{Ve}(\text{co}(\mathcal{N}_{i,\mathcal{G}}))$; see Figure 18(left). Note that since $\mathcal{G} \subseteq \mathcal{G}_{\text{sens}}$ and $p_k$, $p_l$ are neighbors of $p_i$ in $\mathcal{G}$, we have $[p_i, p_k] \subseteq Q_\epsilon$ and $[p_i, p_l] \subseteq Q_\epsilon$. Since $p_i \in [p_k, p_l]$, this implies that $[p_k, p_l] \subseteq Q_\epsilon$. Therefore $[p_k, p_l] \subseteq \text{rco}(\mathcal{P}, Q_\epsilon)$ because the shortest path in $Q_\epsilon$ between $p_k, p_l$ is contained in $\text{rco}(\mathcal{P}, Q_\epsilon)$. Thus, $p_i$ cannot be a vertex of $\text{rco}(\mathcal{P}, Q_\epsilon)$. Let us now look at the case when $p_i$ belongs to the interior of $\text{co}(\mathcal{N}_{i,\mathcal{G}})$; see Figure 18(right). Then, since the strict concavities of $Q_\epsilon$ are all continuously differentiable curves, the shortest path between any two points $p_k, p_l \in \text{Ve}(\text{co}(\mathcal{N}_{i,\mathcal{G}}))$ will be a sequence of straight lines and curves as shown in Figure 18(right). By the definition of the relative convex hull, the shortest paths between two points are contained in the relative convex hull. Thus the region bounded by the shortest paths between adjacent vertices of $\text{co}(\mathcal{N}_{i,\mathcal{G}})$ is contained in the relative convex hull. Clearly, since $p_i$ is the interior of this region, it cannot be a vertex of $\text{rco}(\mathcal{P}, Q_\epsilon)$. This concludes the proof of statement (iii). ■

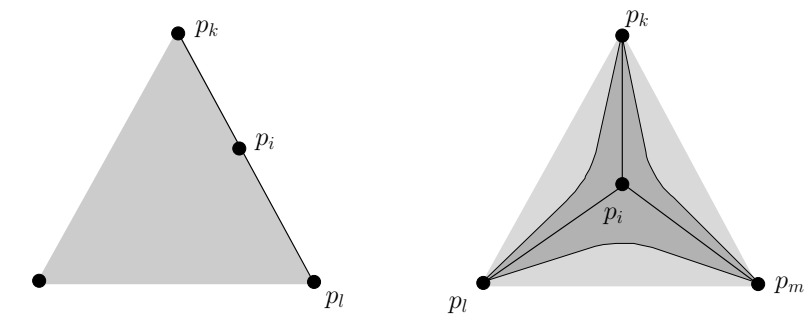

Fig. 18. Illustration of the cases when $p_i \notin \text{Ve}(\text{co}(\mathcal{N}_{i,\mathcal{G}}))$. The point set represented by the black disks is $\mathcal{N}_{i,\mathcal{G}}$. The shaded region is $\text{co}(\mathcal{N}_{i,\mathcal{G}})$. The straight line segments between any two points represent the fact that the two points are visible in $Q_\epsilon$. The curved lines between pairs of points in the figure on the right represent shortest paths in $Q_\epsilon$ between the points.

See Figure 19 for a graphical explanation of Lemma II.1.

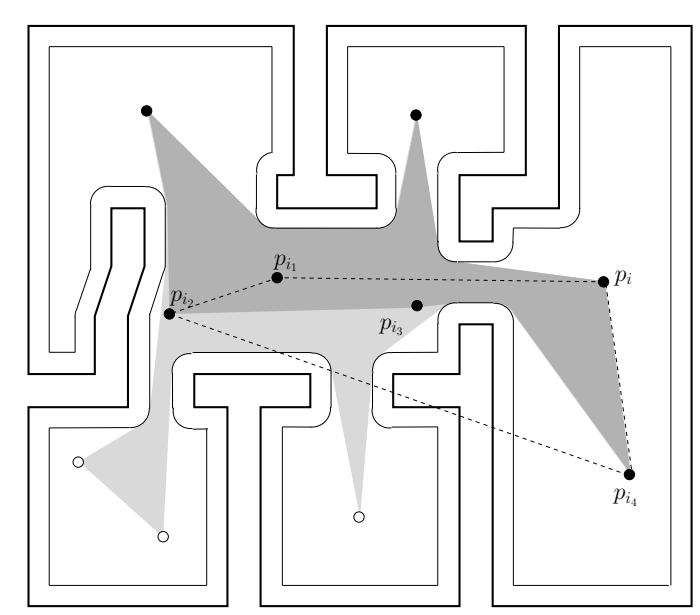

Fig. 19. Illustration of Lemma II.1. The outer polygonal environment is $Q$ and the inner curved environment is $Q_\epsilon$. The set of points represented by black and white disks are $\mathcal{P}_1$ and $\mathcal{P}_2$ respectively. Note that $\text{rco}(\mathcal{P}_1, Q_\epsilon)$, represented by the dark shaded region, is a *subset* of $\text{rco}(\mathcal{P}_1 \cup \mathcal{P}_2, Q_\epsilon)$, represented by the union of the dark and light shaded regions (c.f. Lemma II.1 (i)). In fact $\text{rco}(\mathcal{P}_1, Q_\epsilon)$ is a *strict subset* of $\text{rco}(\mathcal{P}_1 \cup \mathcal{P}_2, Q_\epsilon)$ and hence $\text{rco}(\mathcal{P}_1, Q_\epsilon)$ has a strictly smaller perimeter than $\text{rco}(\mathcal{P}_1 \cup \mathcal{P}_2, Q_\epsilon)$ (c.f. Lemma II.1 (ii)). Now, $p_i \in \text{Ve}(\text{rco}(\mathcal{P}_1 \cup \mathcal{P}_2))$. The set $\{p_i\} \cup \{p_{i_1}, \dots, p_{i_4}\}$ is equal to the set $\mathcal{N}_{i,\mathcal{G}_{\text{sens}}}$ where $\mathcal{G}_{\text{sens}}$ is such that $r = +\infty$ . Note that $p_i \in \text{Ve}(\text{co}(\mathcal{N}_{i,\mathcal{G}_{\text{sens}}}))$ where $\text{co}(\mathcal{N}_{i,\mathcal{G}_{\text{sens}}})$ is the region bounded by the dashed lines (c.f. Lemma II.1 (iii)).

**Lemma II.2 (Properties of the constraint set)** *For any allowable $Q, Q_\epsilon$, if $P \in Q_\epsilon^n$ and $\mathcal{G}_2(P) \subseteq \mathcal{G}_1(P) \subseteq \mathcal{G}_{\text{sens}}(P)$, then for all $p_i$, the following statements hold:*

(i) *The set $C_{p_i,Q}(\mathcal{N}_{i,\mathcal{G}_2}) \cap \text{rco}(\mathcal{N}_{i,\mathcal{G}_1}, \mathcal{V}(p_i, \epsilon))$ is convex; and*

(ii) $C_{p_i,Q}(\mathcal{N}_{i,\mathcal{G}_1}) \cap \text{rco}(\mathcal{N}_{i,\mathcal{G}_1}, \mathcal{V}(p_i, \epsilon)) = C_{p_i,Q}(\mathcal{N}_{i,\mathcal{G}_1}) \cap \text{co}(\mathcal{N}_{i,\mathcal{G}_1})$.

*Proof:* Let $p, q \in C_{p_i,Q}(\mathcal{N}_{i,\mathcal{G}_2}) \cap \text{rco}(\mathcal{N}_{i,\mathcal{G}_1}, \mathcal{V}(p_i, \epsilon))$. Since $C_{p_i,Q}(\mathcal{N}_{i,\mathcal{G}_2})$ is an intersection of convex sets, it is convex as well. This implies that $[p, q] \subseteq C_{p_i,Q}(\mathcal{N}_{i,\mathcal{G}_2})$. Since $C_{p_i,Q}(\mathcal{N}_{i,\mathcal{G}_2}) \subseteq Q_\epsilon$, we deduce that $[p, q] \subseteq Q_\epsilon$. Now, $p, q \in \text{rco}(\mathcal{N}_{i,\mathcal{G}_1}, \mathcal{V}(p_i, \epsilon))$. Therefore, by the definition of the relative convex hull, the shortest path between $p$ and $q$ in $Q_\epsilon$ is also contained in $\text{rco}(\mathcal{N}_{i,\mathcal{G}_1}, \mathcal{V}(p_i, \epsilon))$. But the shortest path is the segment $[p, q]$ since $[p, q] \subseteq Q_\epsilon$. Thus, $[p, q] \subseteq \text{rco}(\mathcal{N}_{i,\mathcal{G}_1}, \mathcal{V}(p_i, \epsilon))$. This completes the proof of statement (i).

To prove statement (ii), note that $\mathrm{rco}(\mathcal{N}_{i,\mathcal{G}_1}, \mathcal{V}(p_i,\epsilon)) \subseteq \mathrm{co}(\mathcal{N}_{i,\mathcal{G}_1})$. Hence $C_{p_i,Q}(\mathcal{N}_{i,\mathcal{G}_1}) \cap \mathrm{rco}(\mathcal{N}_{i,\mathcal{G}_1}, \mathcal{V}(p_i,\epsilon)) \subseteq C_{p_i,Q}(\mathcal{N}_{i,\mathcal{G}_1}) \cap \mathrm{co}(\mathcal{N}_{i,\mathcal{G}_1})$. To prove the other inclusion, let $q \in C_{p_i,Q}(\mathcal{N}_{i,\mathcal{G}_1}) \cap \mathrm{co}(\mathcal{N}_{i,\mathcal{G}_1})$. Since $q \in \mathrm{co}(\mathcal{N}_{i,\mathcal{G}_1})$, then $q$ belongs to the closed triangle formed by $p_i$ and two other points belonging to $\mathrm{Ve}(\mathrm{co}(\mathcal{N}_{i,\mathcal{G}_1}))$, say $p_k, p_l$; see Figure 20. Also $q \in C_{p_i,Q}(\mathcal{N}_{i,\mathcal{G}_1})$. From the construction of $C_{p_i,Q}(\mathcal{N}_{i,\mathcal{G}_1})$, it follows that $q \in \mathcal{V}(p_j,\epsilon)$ for all $p_j \in \mathcal{N}_{i,\mathcal{G}_1}$. In particular, this means that $[q,p_l] \subset Q_\epsilon$ and $[q,p_k] \subset Q_\epsilon$.

Since $p_i, p_l \in \mathcal{N}_{i,\mathcal{G}_1}$ and $[p_i,p_l] \subseteq \mathcal{V}(p_i,\epsilon)$, we deduce that $[p_i,p_l] \subseteq \mathrm{rco}(\mathcal{N}_{i,\mathcal{G}_1}, \mathcal{V}(p_i,\epsilon))$ because the relative convex hull contains the shortest path in $\mathcal{V}(p_i,\epsilon)$ between any two points contained in it. Let the line containing the segment $[q,p_k]$ intersect the segment $[p_i,p_l]$ at $q'$. Note that $q \in [q',p_k]$ because by choice of $p_k$, $p_l$, we have that $q$ belongs to the closed triangle with vertices $p_i$, $p_k$, $p_l$. Now, since $q' \in [p_i,p_l]$ and $[p_i,p_l] \subseteq \mathrm{rco}(\mathcal{N}_{i,\mathcal{G}_1}, \mathcal{V}(p_i,\epsilon))$, then $q' \in \mathrm{rco}(\mathcal{N}_{i,\mathcal{G}_1}, \mathcal{V}(p_i,\epsilon))$. We now claim that the segment $[q',p_k] \subseteq \mathcal{V}(p_i,\epsilon)$. To see this, note that $[q',p_k]$ is a subset of the polygon given by the ordered set of vertices $(p_i,p_k,q,p_l)$. Since all the edges of the polygon, $[p_i,p_k]$, $[p_k,q]$, $[q,p_l]$, $[p_l,p_k]$, are a subset of $Q_\epsilon$, we have that the interior of $Q_\epsilon$ and the interior of the polygon $(p_i,p_k,q,p_l)$ do not intersect. Thus $[q',p_k] \subseteq \mathcal{V}(p_i,\epsilon)$. Now, since $q',p_k \in \mathrm{rco}(\mathcal{N}_{i,\mathcal{G}_1}, \mathcal{V}(p_i,\epsilon))$, the shortest path between $q',p_k$ in $\mathcal{V}(p_i,\epsilon)$ is also contained in $\mathrm{rco}(\mathcal{N}_{i,\mathcal{G}_1}, \mathcal{V}(p_i,\epsilon))$. Thus $q \in [q',p_k] \subseteq \mathrm{rco}(\mathcal{N}_{i,\mathcal{G}_1}, \mathcal{V}(p_i,\epsilon))$. This completes the proof of statement (ii). ■

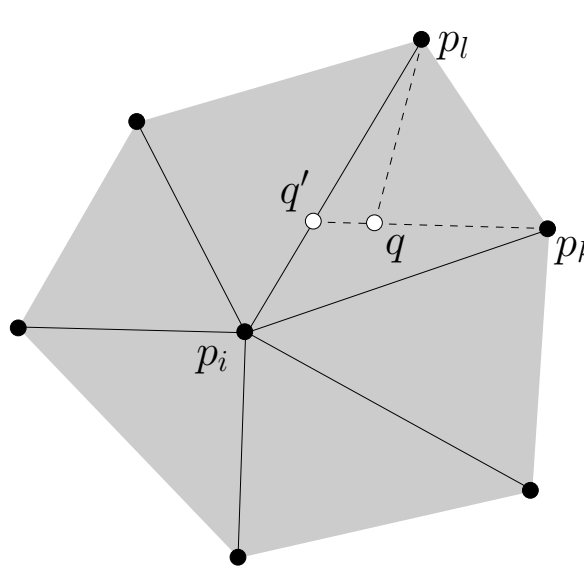

Fig. 20. The set of points represented by black disks is $\mathcal{N}_i$. The shaded region represents $\mathrm{co}(\mathcal{N}_i)$. $q$, represented by the white disk, is any point in $\mathrm{co}(\mathcal{N}_i)$. Since $q$ is $\epsilon$-robustly visible from every point in $\mathcal{N}_i$, the dashed line segments are a subset of $Q_\epsilon$. Also every point in $\mathcal{N}_i \setminus \{p_i\}$ is visible from $p_i$ by the definition of $\mathcal{N}_i$. Hence the solid line segments are also a subset of $Q_\epsilon$.

Statement (i) in the above lemma states that the constraint set is convex. Statement (ii) gives an alternate way of computing the set $X_i$ in step `3:` of the `Perimeter Minimizing Algorithm` when $\mathcal{G}_{\text{constr}}$ and $\mathcal{G}_{\text{sens}}$ are identical.

In what follows, we analyze the `Perimeter Minimizing Algorithm` by means of the *closed perimeter minimizing algorithm* over any proximity graphs $\mathcal{G}_2 \subseteq \mathcal{G}_1 \subseteq \mathcal{G}_{\text{sens}}$, defined as follows:

$$\begin{aligned} &T^c_{\mathcal{G}_1,\mathcal{G}_2}(P)_i \in p_i + u_i, \quad \text{where} \\ &X_i(P) = C_{p_i,Q}(\mathcal{N}_{i,\mathcal{G}_2}) \cap \mathrm{rco}(\mathcal{N}_{i,\mathcal{G}_1}, \mathcal{V}(p_i,\epsilon)), \\ &K_i(P) = \Big\{ \lim_{k_m \to +\infty} \mathrm{CC}(X_i(P_{k_m})) \mid P_k \to P, \\ &\{\mathrm{CC}(X_i(P_{k_m})\} \text{ is a convergent subseq. of } \{\mathrm{CC}(X_i(P_k)\}\Big\}, \\ &u_i \in \left\{ \frac{\min(s_{\max}, \|p_i^* - p_i\|)}{\|p_i^* - p_i\|}(p_i^* - p_i) \;\Big|\; p_i^* \in K_i(P) \right\}, \end{aligned} \tag{7}$$

and where $T^c_{\mathcal{G}_1,\mathcal{G}_2}(P)_i$ is the $i$th component of $T^c_{\mathcal{G}_1,\mathcal{G}_2}(P)$.

Note that $T_{\mathcal{G}_{\text{sens}},\mathcal{G}_{\text{constr}}}(P) \subseteq T^c_{\mathcal{G}_1,\mathcal{G}_2}(P)$ if $\mathcal{G}_1 = \mathcal{G}_{\text{sens}}$ and $\mathcal{G}_2 = \mathcal{G}_{\text{constr}}$. Therefore, the evolution under the `Perimeter Minimizing Algorithm` given by the trajectory $\{P[t]\}_{t\in\mathbb{Z}_{\geq 0}}$ with $P[t+1] \in T_{\mathcal{G}_{\text{sens}},\mathcal{G}_{\text{constr}}}(P[t])$ is just one of the possible evolutions under $T^c_{\mathcal{G}_1,\mathcal{G}_2}$ given by the trajectory $\{P^c[t]\}_{t\in\mathbb{Z}_{\geq 0}}$ with $P^c[t+1] \in T^c_{\mathcal{G}_1,\mathcal{G}_2}(P^c[t])$. We now begin stating the intermediate results for $T^c_{\mathcal{G}_1,\mathcal{G}_2}$. The first of these results essentially states that for at least one robot located at a vertex of the relative convex hull, the set $X_i$ is large enough for it to move under the application of $T^c_{\mathcal{G}_1,\mathcal{G}_2}$. This, in turn, implies that the relative convex hull shrinks.

**Lemma II.3 (Nontrivial constraints)** *Let $Q$, $Q_\epsilon$ be allowable environments. Assume that $P \in Q^n_\epsilon$ and the proximity graphs $\mathcal{G}_1$ and $\mathcal{G}_2$ satisfy*

A) *not all points in $P$ are equal; and*
B) *$\mathcal{G}_2(P) \subseteq \mathcal{G}_1(P) \subseteq \mathcal{G}_{\text{sens}}(P)$ and $\mathcal{G}_2(P)$ is connected.*

*Then there exists $p_i \in \mathrm{Ve}(\mathrm{rco}(P,Q_\epsilon))$ such that $\mathrm{diam}(X_i) > 0$, where $X_i = C_{p_i,\mathcal{N}_{i,\mathcal{G}_2}} \cap \mathrm{rco}(\mathcal{N}_{i,\mathcal{G}_1}, \mathcal{V}(p_i,\epsilon))$.*

*Proof:* Note that $p_i \subseteq X_i$ and $X_i$ is convex. Thus, $\mathrm{diam}(X_i) = 0$ if and only if $X_i = \{p_i\}$. The remainder of the proof is organized as follows. We first establish some necessary conditions for $X_i = \{p_i\}$. We then show that under the assumptions in the lemma, those necessary conditions cannot hold for every $p_i \in \mathrm{Ve}(\mathrm{rco}(P,Q_\epsilon))$. Note that $X_i = C_{p_i,Q}(\mathcal{N}_{i,\mathcal{G}_2}) \cap \mathrm{rco}(\mathcal{N}_{i,\mathcal{G}_1}, \mathcal{V}(p_i,Q_\epsilon))$ and $C_{p_i,Q}(\mathcal{N}_{i,\mathcal{G}_2}) = \bigcap_{p_j \in \mathcal{N}_{i,\mathcal{G}_2}} \mathcal{C}_Q(p_i,p_j)$. For any $p_j \in \mathcal{N}_{i,\mathcal{G}_2}$, $\mathcal{C}_Q(p_i,p_j) = \mathcal{V}(p_i,\epsilon) \cap B(\frac{p_i+p_j}{2}, \frac{r}{2}) \bigcap_{k=1}^{n_{ij}} H_{Q_\epsilon}(v_{i,j,k})$ where $v_{i,j,k}$'s are computed at step `3:` of the `Constraint Set Generator Algorithm`. Thus, we can write

$$X_i = \mathcal{V}(p_i,\epsilon) \cap \left( \bigcap_{p_j \in \mathcal{N}_{i,\mathcal{G}_2}} \left( B(\frac{p_i+p_j}{2}, \frac{r}{2}) \cap \left( \bigcap_{k=1}^{n_{ij}} H_{Q_\epsilon}(v_{i,j,k}) \right) \right) \right) \cap \mathrm{rco}(\mathcal{N}_{i,\mathcal{G}_1}, \mathcal{V}(p_i,Q_\epsilon)),$$

which can be rewritten as

$$X_i = \mathcal{V}(p_i,\epsilon) \cap \mathrm{rco}(\mathcal{N}_{i,\mathcal{G}_1}, \mathcal{V}(p_i,Q_\epsilon)) \cap \left( \bigcap_{p_j \in \mathcal{N}_{i,\mathcal{G}_2}} B(\frac{p_i+p_j}{2}, \frac{r}{2}) \right) \cap \left( \bigcap_{p_j \in \mathcal{N}_{i,\mathcal{G}_2}} \bigcap_{k=1}^{n_{ij}} H_{Q_\epsilon}(v_{i,j,k}) \right). \tag{8}$$

Since $\mathrm{rco}(\mathcal{N}_{i,\mathcal{G}_1}, \mathcal{V}(p_i,\epsilon)) \subseteq \mathcal{V}(p_i,\epsilon)$, we have that $\mathcal{V}(p_i,\epsilon) \cap \mathrm{rco}(\mathcal{N}_{i,\mathcal{G}_1}, \mathcal{V}(p_i, Q_\epsilon)) = \mathrm{rco}(\mathcal{N}_{i,\mathcal{G}_1}, \mathcal{V}(p_i,\epsilon))$. From assumptions (A) and (B), any robot $i$ will have a neighbor in $\mathcal{G}_1(P)$ and $\mathcal{G}_2(P)$ that is not placed at $p_i$. Hence, $\mathcal{N}_{i,\mathcal{G}_1}$ and $\mathcal{N}_{i,\mathcal{G}_2}$ are point sets strictly larger than $\{p_i\}$. Therefore, $\mathrm{rco}(\mathcal{N}_{i,\mathcal{G}_1}, \mathcal{V}(p_i,\epsilon))$ is strictly larger than $\{p_i\}$. Therefore, there exists a neighborhood around $p_i$ in $\mathrm{rco}(\mathcal{N}_{i,\mathcal{G}_1}, \mathcal{V}(p_i,\epsilon))$, as shown by the shaded region in Figure 23. Let this neighborhood be $\mathcal{B}$.

Therefore, if $p_i$ belongs to the interior of $\left(\bigcap_{p_j\in\mathcal{N}_{i,\mathcal{G}_2}} B(\frac{p_i+p_j}{2},\frac{r}{2})\right) \cap \left(\bigcap_{p_j\in\mathcal{N}_{i,\mathcal{G}_2}} \bigcap_{k=1}^{n_{ij}} H_{Q_\epsilon}(v_{i,j,k})\right)$, then $X_i$ will be strictly larger than $\{p_i\}$. Hence, if $X_i = \{p_i\}$ then $p_i$ belongs to the boundary of one or more of the sets $B(\frac{p_i+p_j}{2},\frac{r}{2})$ and $H_{Q_\epsilon}(v_{i,j,k})$, where $p_j \in \mathcal{N}_{i,\mathcal{G}_2}$ and $k \in \{1,\dots,n_{ij}\}$.

If $\|p_i - p_j\| < r$, then $p_i$ belongs to the interior of $B(\frac{p_i+p_j}{2},\frac{r}{2})$; see Figure 21 (left). Also, if $v_{i,j,k} \notin [p_i,p_j]$ then $p_i$ belongs to the interior of $H_{Q_\epsilon}(v_{i,j,k})$; see Figure 21 (right). Therefore, if $p_i$ belongs to the boundary of $H_{Q_\epsilon}(v_{i,j,k})$, then $v_{i,j,k} \in [p_i,p_j]$ and if $p_i$ belongs to the boundary of $B(\frac{p_i+p_j}{2},\frac{r}{2})$, then $\|p_i - p_j\| = r$.

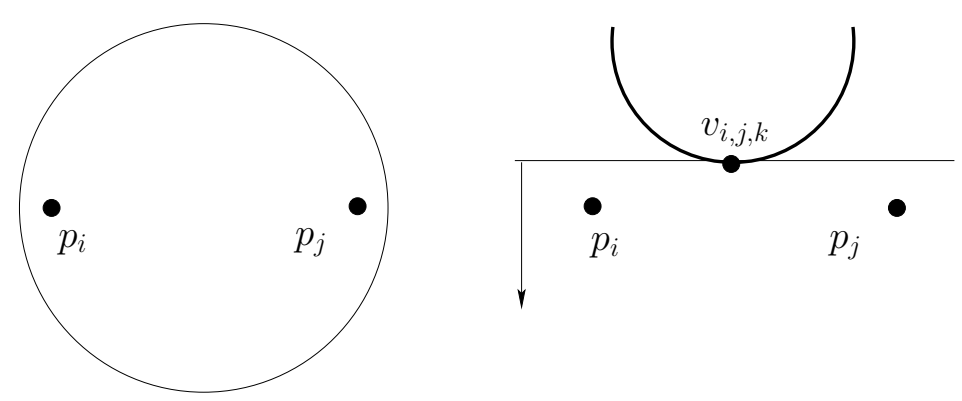

Fig. 21. In the figure on the left, $\|p_i - p_j\| < R$. The circle represents $B(\frac{p_i+p_j}{2},\frac{r}{2})$. In the figure on the curved line represents a strict concavity. $v_{i,j,k}$ represents the point on the concavity nearest to $[p_i,p_j]$. It can be seen that $\mathrm{dist}(v_{i,j,k},[p_i,p_j]) > 0$. The half-plane tangent to the strict concavity at $v_{i,j,k}$ and with interior in the direction of the arrow is $H_{Q_\epsilon}(v_{i,j,k})$.

Since $p_i \in \mathrm{Ve}(\mathrm{rco}(P,Q_\epsilon))$, from Lemma II.1 (iii) we have that $p_i \in \mathrm{Ve}(\mathrm{co}(\mathcal{N}_{i,\mathcal{G}_2}))$. This implies that $\bigcap_{p_j\in\mathcal{N}_{i,\mathcal{G}_2}} B(\frac{p_i+p_j}{2},\frac{r}{2})$ is a convex set strictly containing $\{p_i\}$; see Figure 22.

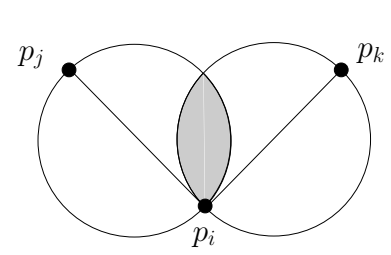

Fig. 22. $p_i$ is a vertex of the convex hull of the set $\{p_i,p_j,p_k\}$. Here $\|p_i - p_j\| = \|p_i - p_k\| = r$. Note that the angle between the vectors $p_j - p_i$ and $p_k - p_i$ is strictly less than $\pi$. The circles represent $B(\frac{p_i+p_j}{2},\frac{r}{2})$ and $B(\frac{p_i+p_j}{2},\frac{k}{2})$ and the shaded region represents their intersection. If $\|p_i - p_j\| < r$ or $\|p_i - p_k\| < r$, the region of intersection will still strictly contain $\{p_i\}$.

Therefore $\bigcap_{p_j\in\mathcal{N}_{i,\mathcal{G}_2}} B(\frac{p_i+p_j}{2},\frac{r}{2}) \cap \mathcal{B}$ strictly contains $\{p_i\}$. Therefore, $p_i$ must belong to the boundary of some $H_{Q_\epsilon}(v_{i,j,k})$. It is easy to see that $H_{Q_\epsilon}(v_{i,j,k}) \cap \mathcal{B}$ is strictly larger than $\{p_i\}$. Therefore, $p_i$ must belong to the boundary of another set $H_{Q_\epsilon}(v_{i,k,l})$ or $B(\frac{p_i+p_j}{2},\frac{r}{2})$ as explained in Figure 23.

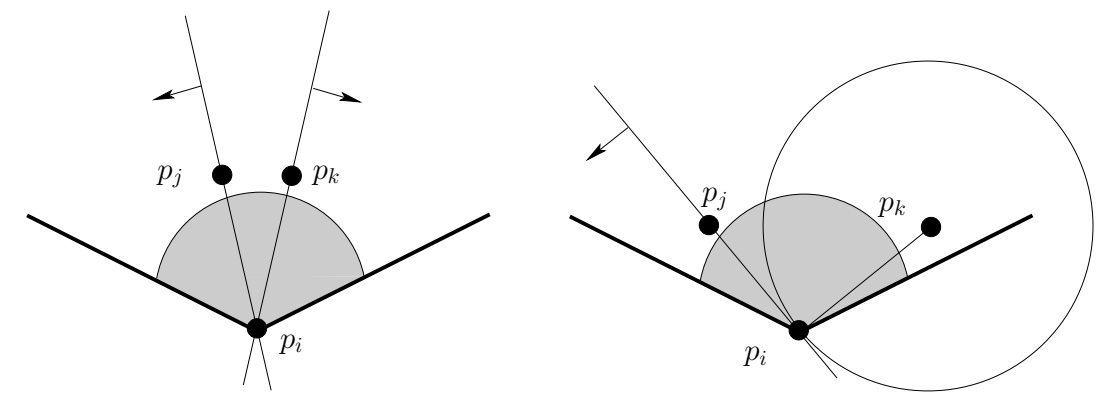

Fig. 23. The thick lines represent the boundary of $\mathrm{rco}(\mathcal{N}_{i,\mathcal{G}_1}, \mathcal{V}(p_i,\epsilon))$. The shaded region is a subset of $\mathrm{rco}(\mathcal{N}_{i,\mathcal{G}_1}, \mathcal{V}(p_i,\epsilon))$ and forms an entire neighborhood around $p_i$. $H_{Q_\epsilon}(v_{i,j,l})$ and $H_{Q_\epsilon}(v_{i,k,m})$ are the half-planes with boundary being the lines passing through $[p_i,p_j]$ and $[p_i,p_k]$ respectively and interior in the direction of the arrows. In the figure on the right, $\|p_j - p_i\| = r$ and the angles between the vectors $p_j - p_i$ and $p_k - p_i$ is greater than $\frac{\pi}{2}$. The circle represents $B(\frac{p_i+p_k}{2},\frac{r}{2})$.

Therefore, for $p_i \in \mathrm{Ve}(\mathrm{rco}(P,Q_\epsilon))$, we obtain the following necessary condition for $\mathrm{diam}(X_i) = 0$. There exist at least two neighbors of $p_i$ located at $p_j, p_k$ such that one or more than one of following cases hold:

(A) $\|p_j - p_i\| = r$, $v_{i,k,l} \in [p_i,p_k]$ for some $l \in \{1,\dots,n_{ik}\}$, the inner product $\langle p_j - p_i, p_k - p_i\rangle \le 0$ and $p_j \notin H_{Q_\epsilon}(v_{i,k,l})$;

(B) $v_{i,j,m} \in [p_i,p_j]$, $v_{i,k,l} \in [p_i,p_k]$ and $p_k \notin H_{Q_\epsilon}(v_{i,j,m})$, $p_j \notin H_{Q_\epsilon}(v_{i,k,l})$ for some $m \in \{1,\dots,n_{ij}\}$ and some $l \in \{1,\dots,n_{ik}\}$.

Cases (A) and (B) correspond to Figure 24 left and right respectively.

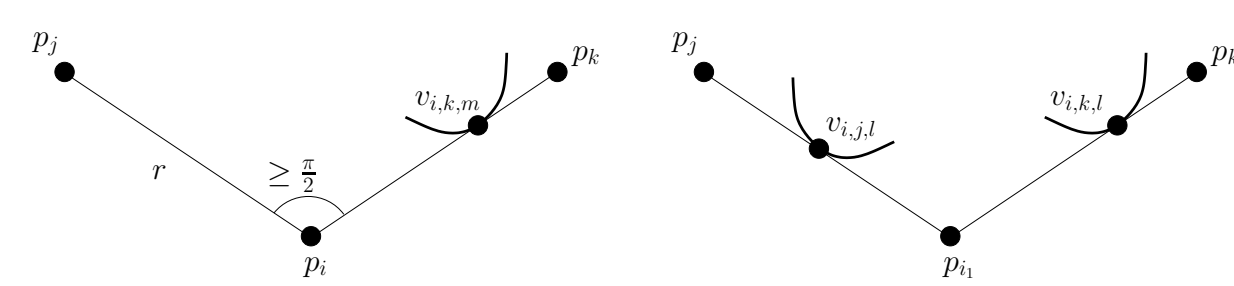

Fig. 24. Illustration of cases (A) and (B) in the proof of Lemma II.3. These cases illustrate the necessary conditions in order for the constraint set $X_i$ to be a point.

We now show that the necessary condition cannot hold for every $p_i \in \mathrm{Ve}(\mathrm{rco}(P,Q_\epsilon))$. Let $p_{i_1} \in \mathrm{Ve}(\mathrm{rco}(P,Q_\epsilon))$. Then $p_{i_1}$ has at least two neighbors in $\mathcal{G}_2(P)$ located at $p_j, p_k$ such that conditions (A) and/or (B) are satisfied. Now $p_{i_1} \in \partial\,\mathrm{rco}(P,Q_\epsilon)$ because $p_{i_1} \in \mathrm{Ve}(\mathrm{rco}(P,Q_\epsilon))$. Also, $v_{i,k,l} \in \partial Q_\epsilon \cap \mathrm{rco}(P,Q_\epsilon)$. Hence $v_{i,k,l} \in \partial\,\mathrm{rco}(P,Q_\epsilon)$. Now, since $Q_\epsilon$ does not contain any hole, $\mathrm{rco}(P,Q_\epsilon)$ is simply connected. Thus the segment $[p_{i_1}, v_{i,k,l}]$ partitions $\mathrm{rco}(P,Q_\epsilon)$ into two simply connected sets closed sets, $Q'_{i_1}$ and $Q''_{i_1}$ such that $Q'_{i_1} \cap Q''_{i_1} = [p_{i_1}, v_{i,k,l}]$. Let $p_k \in Q'_{i_1}$.

Let us construct a segment $[v_{i,k,l}, d]$ by extending the segment $[p_{i_1}, v_{i,k,l}]$ till it intersects the boundary of $\mathrm{rco}(P,Q_\epsilon)$ at $d$. We refer to Figure 25 for an illustration of this argument. Then the segment $[v_{i,k,l}, d]$ partitions $\mathrm{rco}(P,Q_\epsilon)$ into two components such that again there exists $p'_{i_1} \in \mathrm{Ve}(P,Q_\epsilon)$ in the component not containing $p_{i_1}$. We point that $p'_{i_1} = v_{i,k,l}$, $p'_{i_1} = p_k$ and $v_{i,k,l} = d$ are possible. Now, $p'_{i_1}$ can only contain one neighbor in $Q''_{i_1}$ and that can only be $p_{i_1}$ since all other points in $Q''_{i_1}$ are not $\epsilon$-robust visible from $p'_{i_1}$.

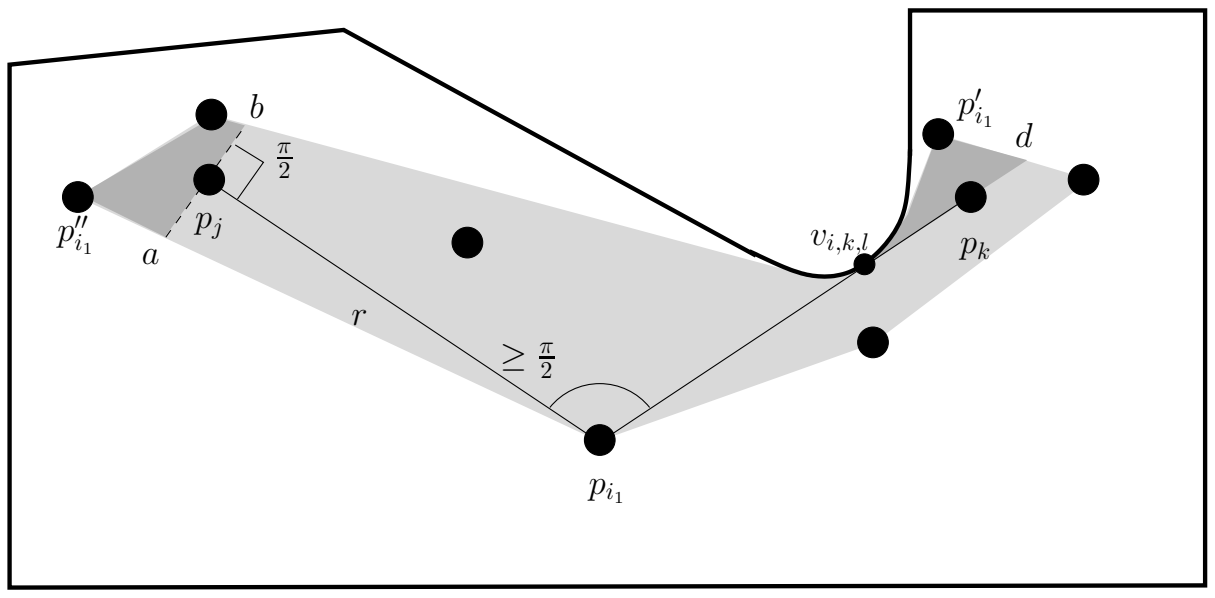

Fig. 25. Illustration of the construction of various segments and the partitions induced by them in the proof of Lemma II.3.

Let $p_j$ be such that $\|p_j - p_{i_1}\| = r$ and the inner product $\langle p_j - p_{i_1}, p_k - p_{i_1}\rangle \leq 0$. Let us construct a line segment $[a, b]$ perpendicular to the segment $[p_{i_1}, p_j]$, passing through $p_j$ and intersecting the boundary of $\mathrm{rco}(P, Q_\epsilon)$ at points $a$ and $b$. The segment $[a, b]$ partitions $\mathrm{rco}(P, Q_\epsilon)$ into two components such that there exists $p''_{i_1} \in \mathrm{Ve}(P, Q_\epsilon)$ in the component not containing $p_{i_1}$. Here we point that $p''_{i_1} = p_j$ is possible. Now, $p''_{i_1}$ can only contain one neighbor in $Q'_{i_1}$ and that can only be $p_{i_1}$ since all other points in $Q'_{i_1}$ are at a distance strictly greater than $r$.

If, on the other hand, $p_j$ is such that there exists $v_{i,j,m} \in [p_i, p_j]$ and $p_k \notin H_{Q_\epsilon}(v_{i,j,m})$, $p_j \notin H_{Q_\epsilon}(v_{i,k,l})$, then we proceed in a way similar to the case of $p_k$. Let us construct a line segment $[v_{i,j,m}, e]$ by extending the line segment $[p_i, v_{i,j,m}]$ till it intersects the boundary of $\mathrm{rco}(P, Q_\epsilon)$ at $e$. Then the segment $[v_{i,j,m}, e]$ partitions $\mathrm{rco}(P, Q_\epsilon)$ into two components such that again there exists $p''_{i_1} \in \mathrm{Ve}(P, Q_\epsilon)$ in the component not containing $p_{i_1}$. We point that $p''_{i_1} = v_{i,j,m}$, $p''_{i_1} = p_j$ and $v_{i,j,m} = e$ are possible. Now, as before $p''_{i_1}$ can only contain one neighbor in $Q'_{i_1}$ and that can only be $p_{i_1}$ since all other points in $Q'_{i_1}$ are not $\epsilon$-robust visible from $p''_{i_1}$.

Thus, $X_{i_1} = \{p_{i_1}\}$ implies that there exists a partition of $\mathrm{rco}(P, Q_\epsilon)$ into $Q'_{i_1}$ and $Q''_{i_1}$ containing $p'_{i_1} \in \mathrm{Ve}(\mathrm{rco}(P, Q_\epsilon))$ and $p''_{i_1} \in \mathrm{Ve}(\mathrm{rco}(P, Q_\epsilon))$ respectively such that $p'_{i_1}$ has only one neighbor in $Q''_{i_1}$ and vice versa. Now, let $p_{i_2} = p''_{i_1}$. Therefore, $p_{i_2} \neq p_{i_1}$. Again $X_{i_2} = \{p_{i_2}\}$ implies that $\mathrm{rco}(P, Q_\epsilon)$ can be partitioned into $Q'_{i_2}$ and $Q''_{i_2}$. Let $p_{i_1} \in Q'_{i_2}$. Therefore, there exists $p''_{i_2} \in \mathrm{Ve}(\mathrm{rco}(P, Q_\epsilon))$ such that $p_{i_1}$ cannot be a neighbor of $p''_{i_2}$. Let $p_{i_3} = p''_{i_2}$. Therefore, $X_{i_2} = \{p_{i_2}\}$ implies the existence of $p_{i_3} \in \mathrm{Ve}(\mathrm{rco}(P, Q_\epsilon))$ such that $p_{i_3} \notin \{p_{i_1}, p_{i_2}\}$. Let the cardinality of $\mathrm{Ve}(\mathrm{rco}(P, Q_\epsilon))$ be $n_v$. Proceeding recursively, $X_{i_{n_v}} = \{p_{i_{n_v}}\}$ implies the existence of $p_{i_{n_v+1}} \notin \{p_{i_1}, \dots, p_{i_{n_v}}\}$. This is a contradiction. ∎

Now, let $G_1(P)$ and $G_2(P)$ be graphs with fixed topologies and let $G_2$ be a subgraph of $G_1$. Let $Y_{G_1,G_2} = \{P \in Q_\epsilon^n \mid G_2(P) \subseteq G_1(P) \subseteq \mathcal{G}_{\text{sens}}(P)\}$.

We now present some results on smoothness of $T^c_{G_1,G_2}$ on $Y_{G_1,G_2}$ and contraction of $V_{\text{perim},Q_\epsilon}$ under the action of $T^c_{G_1,G_2}$. The technical approach in what follows is similar to the one in [7].

**Lemma II.4 (Properties of the `Perimeter Minimizing Algorithm`)** *The set $Y_{G_1,G_2}$ and the map $T^c_{G_1,G_2}$ have the following properties:*

(i) *$Y_{G_1,G_2}$ is compact, and $T^c_{G_1,G_2}$ is closed on $Y_{G_1,G_2}$;*
(ii) *for all $i \in \{1, \dots, n\}$, $T^c_{G_1,G_2}(P)_i \subseteq (\mathrm{rco}(P, Q_\epsilon) \setminus \mathrm{Ve}(\mathrm{rco}(P, Q_\epsilon))) \cup \{p_i\}$;*
(iii) *if $G_2$ is connected, then there exists $p_i \in \mathrm{Ve}(\mathrm{rco}(P, Q_\epsilon))$ such that $T^c_{G_1,G_2}(P)_i \subseteq \mathrm{rco}(P, Q_\epsilon) \setminus \mathrm{Ve}(\mathrm{rco}(P, Q_\epsilon))$; and*
(iv) *$\mathrm{rco}(T^c_{G_1,G_2}(P), Q_\epsilon)) \subseteq \mathrm{rco}(P, Q_\epsilon)$ at any $P \in Y_{G_1,G_2}$,*

*where $T^c_{G_1,G_2}(P)_i$ is the $i$th component of $T^c_{G_1,G_2}(P)$.*

*Proof:* We begin by proving statement (i). Showing that $Y_{G_1,G_2}$ is compact is equivalent to showing that $Y_{G_1,G_2}$ is closed and bounded. Since $Y_{G_1,G_2} \subseteq Q_\epsilon$, then $Y_{G_1,G_2}$ is bounded. Now, let $\{P_k\}$ be any sequence in $Y_{G_1,G_2}$ such that $P_k \to P$. Let us show that $P \in Y_{G_1,G_2}$. By definition, $G_2(P) \subseteq G_1(P)$. It remains to be shown that $G_1(P) \subseteq \mathcal{G}_{\text{sens}}(P)$. To see this, let $(p_i, p_j)$ be an edge that does not belong to $\mathcal{G}_{\text{sens}}(P)$. We now show that it cannot belong to $G_1(P)$. Then, either $[p_i, p_j] \not\subset Q_\epsilon$ or $\|p_i - p_j\| > R$. If $[p_i, p_j] \not\subset Q_\epsilon$, we can construct neighborhoods around $p_i$ and $p_j$ such that for all $p'_i$ and $p'_j$ belonging to the neighborhoods, we have $[p'_i, p'_j] \not\subset Q_\epsilon$; see Figure 26. If, on the other hand, $\|p_i - p_j\| = R + \delta$, then let $p'_i, p'_j$ be any points such that $\|p'_i - p_i\| < \frac{\delta}{2}$ and $\|p'_j - p_j\| < \frac{\delta}{2}$. Then $\|p'_i - p'_j\| = \|p'_i - p_i + p_i - p_j + p_j - p'_j\| \geq \|p_i - p_j\| - \|p'_i - p_i\| - \|p'_j - p_j\| > R + \delta - \frac{\delta}{2} - \frac{\delta}{2} = R$. Thus there exist neighborhoods around $p_i$ and $p_j$ such that for all $p'_i$ and $p'_j$ belonging to the neighborhoods, we have $\|p'_i - p'_j\| > R$. Therefore, we can construct a neighborhood around $P$ in $Q_\epsilon$ such that for all $P'$ in the neighborhood, $(p'_i, p'_j)$ is not an edge of $\mathcal{G}_{\text{sens}}(P')$ if $(p_i, p_j)$ is not an edge of $\mathcal{G}_{\text{sens}}(P)$. Since $P_k \to P$, there exists $k_0$ such that for all $k \geq k_0$, $P_k$ belongs to this neighborhood. Therefore, for all $k \geq k_0$, $(p^k_i, p^k_j)$ is not an edge of $\mathcal{G}_{\text{sens}}(P_k)$. Since $G_1(P_k) \subseteq \mathcal{G}_{\text{sens}}(P_k)$, for all $k \geq k_0$, $(p^k_i, p^k_j)$ is not an edge of $G_1(P_k)$. Then, since $G_1(P)$ has the same topology as $G_1(P_k)$, we have that $(p_i, p_j)$ is not an edge of $G_1(P)$. This completes the proof that $Y_{G_1,G_2}$ is compact.

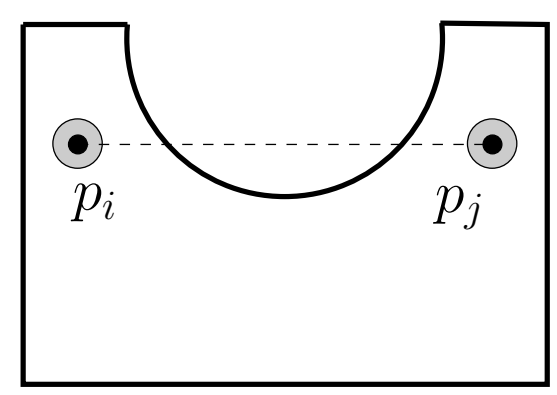

Fig. 26. If $[p_i, p_j] \not\subset Q_\epsilon$, there exist neighborhoods of $p_i$ and $p_j$ such that $[p'_i, p'_j] \not\subset Q_\epsilon$, for all $p'_i$ and $p'_j$ belonging to the neighborhoods.

Showing that $T^c_{G_1,G_2}$ is closed over $Y_{G_1,G_2}$ is equivalent to proving that the map $P \to K_i(P)$ is closed over $Y_{G_1,G_2}$ for each $i \in \{1, \dots, n\}$. Now, let $P_l \to P$ and let $q_l \to q$ where $q_l \in K_i(P_l)$. We need to show that $q \in K_i(P)$.Since $q_l \in K_i(P_l)$, we have that $q_l = \lim_{m\to\infty} \mathrm{CC}(X_i(P'_{k_m}))$ for some sequence $P'_k \to P_l$. Hence, given $\delta > 0$, there exists $m_0(l)$ such that for all $m \geq m_0(l)$, we have $\|q_l - \mathrm{CC}(X_i(P'_{k_m}))\| \leq \delta$. Also, since $P'_k \to P_l$, there exists $k_l$ such that for all $k \geq k_l$, we have $\|P'_k - P_l\| \leq \delta$. Without loss of generality, let us assume that $k_{m_0(l)} > k_l$. Now, let $P''_l = P'_{k_{m_0(l)}}$ and $q''_l = \mathrm{CC}(X_i(P''_l))$. Therefore, $\|P''_l - P_l\| \leq \delta$ and $\|q''_l - q_l\| \leq \delta$.

Since $P_l \to P$ and $q_l \to q$, there exists $l_0$ such that for all $l \geq l_0$, we have that $\|P_l - P\| \leq \delta$ and $\|q_l - q\| \leq \delta$. But $\|P_l - P\| = \|P_l - P_l'' + P_l'' - P\| \geq \|P_l'' - P\| - \|P_l - P"_l\|$. Therefore, $\|P_l'' - P\| \leq \delta + \|P_l - P"_l\| \leq 2\delta$. Hence $P_l'' \to P$. Similarly, it follows that for $l \geq l_0$. we have $\|q_l'' - q\| \leq 2\delta$. Thus, $q_l'' \to q$. It then follows from the definition of $K_i(P)$ that $q \in K_i(P)$.

We now prove statement (ii). It suffices to prove that $K_i(P) \subseteq (\mathrm{rco}(P, Q_\epsilon) \setminus \mathrm{Ve}(\mathrm{rco}(P, Q_\epsilon))) \cup \{p_i\}$. Let $q \in K_i(P)$. Therefore, $q = \lim_{k_m \to \infty} \mathrm{CC}(X_i(P_{k_m}))$ where $\{P_{k_m}\}$ is some subsequence of $P_k \to P$. We begin by showing that the map $P \to X_i(P)$ is closed at $P$. Note that $X_i(P) = C_{p_i,Q}(\mathcal{N}_{i,G2}) \cap \mathrm{rco}(\mathcal{N}_{i,G_1}, \mathcal{V}(p_i, \epsilon))$ and that $C_{p_i,Q}(\mathcal{N}_{i,G2}) = \cap_{p_j \in \mathcal{N}_{i,G_2}} \mathcal{C}_Q(p_i, p_j)$. If $p_j \in \mathcal{N}_{i,G_2}$, then $(p_i, p_j)$ is an edge of $G_2(P) \subseteq \mathcal{G}_{\text{sens}}(P)$. This implies that $(p_i, p_j) \in J$. But from Lemma IV.2 (iv), the map $(p_i, p_j) \to \mathcal{C}_Q(p_i, p_j)$ is closed over $J$. Therefore, the map $P \to \mathcal{C}_Q(p_i, p_j)$ is closed over $Y_{G_1,G_2}$ for each $p_j \in \mathcal{N}_{i,G_2}$. Since $G_2$ has fixed topology, we have that the set of indices $\{j \mid p_j \in \mathcal{N}_{i,G_2}\}$ is constant over $Y_{G_1,G_2}$. Thus, the map $P \to C_{p_i,Q}(\mathcal{N}_{i,G2})$ is closed over $Y_{G_1,G_2}$ since by Proposition 4 in [17], the intersection of closed maps is closed.

We now show that $\mathrm{rco}(\mathcal{N}_{i,G_1}, \mathcal{V}(p_i, \epsilon)) = \mathrm{rco}(\mathcal{N}_{i,G_1}, Q_\epsilon)$. Since $\mathcal{V}(p_i, \epsilon) \subseteq Q_\epsilon$, the inclusion $\mathrm{rco}(\mathcal{N}_{i,G_1}, \mathcal{V}(p_i, \epsilon)) \subseteq \mathrm{rco}(\mathcal{N}_{i,G_1}, Q_\epsilon)$ follows directly. Let $q_1,\ q_2 \in \mathrm{rco}(\mathcal{N}_{i,G_1}, \mathcal{V}(p_i, \epsilon))$. Therefore, $q_1,\ q_2 \in \mathcal{V}(p_i, \epsilon)$. By definition the shortest path in $\mathcal{V}(p_i, \epsilon)$ between $q_1$ and $q_2$ is contained in $\mathrm{rco}(\mathcal{N}_{i,G_1}, \mathcal{V}(p_i, \epsilon))$. Since $Q_\epsilon$ does not contain any holes, the shortest path between $q_1$ and $q_2$ in $\mathcal{V}(p_i, \epsilon)$ is also the shortest path between $q_1$ and $q_2$ in $Q_\epsilon$. Therefore, by definition of a relative convex set, $\mathrm{rco}(\mathcal{N}_{i,G_1}, \mathcal{V}(p_i, \epsilon))$ is convex relative to $Q_\epsilon$. Since $\mathcal{N}_{i,G_1} \subseteq \mathrm{rco}(\mathcal{N}_{i,G_1}, \mathcal{V}(p_i, \epsilon))$ and $\mathrm{rco}(\mathcal{N}_{i,G_1}, Q_\epsilon)$ is the intersection of all relative convex set containing $\mathcal{N}_{i,G_1}$, we have that $\mathrm{rco}(\mathcal{N}_{i,G_1}, Q_\epsilon) \subseteq \mathrm{rco}(\mathcal{N}_{i,G_1}, \mathcal{V}(p_i, \epsilon))$. This proves that $\mathrm{rco}(\mathcal{N}_{i,G_1}, \mathcal{V}(p_i, \epsilon)) = \mathrm{rco}(\mathcal{N}_{i,G_1}, Q_\epsilon)$.

Now, since the shortest distances between two points in $Q_\epsilon$ change continuously as the position of the points, the relative convex hull of the points also changes continuously as a function of the points. Again the set $\{j \mid p_j \in \mathcal{N}_{i,G_1}\}$ is constant over $Y_{G_1,G_2}$. Thus, the map $P \to \mathrm{rco}(\mathcal{N}_{i,G_1}, Q_\epsilon)$ is continuous over $Y_{G_1,G_2}$ and hence, also closed. Finally, since $X_i(P)$ is the intersection of closed maps, we have that $P \to X_i(P)$ is closed.

Then from the definition of a closed map $q \in X_i(P)$. But $X_i(P) \subseteq \mathrm{rco}(\mathcal{N}_{i,G_1}, \mathcal{V}(p_i, \epsilon)) = \mathrm{rco}(\mathcal{N}_{i,G_1}, Q_\epsilon)$ and since $\mathcal{N}_{i,G_1} \subseteq \{p_i \mid \{1, \ldots, n\}\}$, we have that $\mathrm{rco}(\mathcal{N}_{i,G_1}, Q_\epsilon) \subseteq \mathrm{rco}(P, Q_\epsilon)$. Thus, $X_i(P) \subseteq \mathrm{rco}(P, Q_\epsilon)$ and hence $q \in \mathrm{rco}(P, Q_\epsilon)$. Now, let $q \in \mathrm{Ve}(\mathrm{rco}(P, Q_\epsilon))$. We show that then $\mathrm{diam}(X_i(P)) = 0$. Our strategy is as follows. We first show that if $q \in \mathrm{Ve}(\mathrm{rco}(P, Q_\epsilon))$, then $\mathrm{diam}(X_i(P_{k_m})) \to 0$. But that in turn implies that $q = p_i$ and hence $p_i \in \mathrm{Ve}(\mathrm{rco}(P, Q_\epsilon))$. But $p_i \in \mathrm{Ve}(\mathrm{rco}(P, Q_\epsilon))$ and $\mathrm{diam}(X_i(P)) > 0$ together imply that $\mathrm{diam}(X_i(P_{k_m}))$ does not tend to zero. This is a contradiction. Therefore, $\mathrm{diam}(X_i(P)) = 0$.

We now start by showing that $\mathrm{diam}(X_i(P_{k_m})) \to 0$. The range of the map $P \to X_i(P)$ is compact, the notions of upper semicontinuity and closedness are identical. Therefore, $P \to X_i(P)$ is upper semicontinuous at any point $P \in Y_{G_1,G_2}$. It follows from the definition of upper semicontinuity that given any $\delta > 0$, there exists $m_0$ such that for all $m \geq m_0$, we have $X_i(P_{k_m}) \subseteq \cup_{x \in X_i(P)} B(x, \delta)$.

Since $X_i(P) \subseteq \mathrm{rco}(P, Q_\epsilon)$, $q$ is a strictly convex point on the boundary of $X_i(P)$. By a strictly convex point, we mean that there exists a tangent to $\partial X_i(P)$ at $q$ which intersects $\partial X_i(P)$ at exactly one point. Let $q_1$ be the point on the boundary of $\cup_{x \in X_i(P)} B(x, \delta)$ such that $\|q_1 - q\| = \delta$; see Figure 27. It follows that $q_1$ is a strictly convex point on the boundary of $\cup_{x \in X_i(P)} B(x, \delta)$. Let $l_1$ be the tangent at $q_1$. Now, let $l_2$ be the tangent to $B(q, \delta)$ parallel to $l_1$. It is easy to see that as $\delta \to 0$, the region of $\cup_{x \in X_i(P)} B(x, \delta)$ between the lines $l_1$ and $l_2$ tends to the point set $\{q\}$. Now, since $\mathrm{CC}(X_i(P_{k_m})) \to q$, there exists $m_1$ such that for all $m \geq m_1$, $\mathrm{CC}(X_i(P_{k_m})) \in B(q, \delta)$. For $m \geq \max\{m_0, m_1\}$, let $y$ be a point on the opposite side of $l_2$ as $\mathrm{CC}(X_i(P_{k_m}))$ and equidistant to $l_2$ as $\mathrm{CC}(X_i(P_{k_m}))$; see Figure 27. All points on the right of $l_2$ are nearer to $y$ than to $\mathrm{CC}(X_i(P_{k_m}))$. Also, the distance of $y$ to any point in $\cup_{x \in X_i(P)} B(x, \delta)$ on or to the left of $l_2$ tends to 0 as $\delta \to 0$.

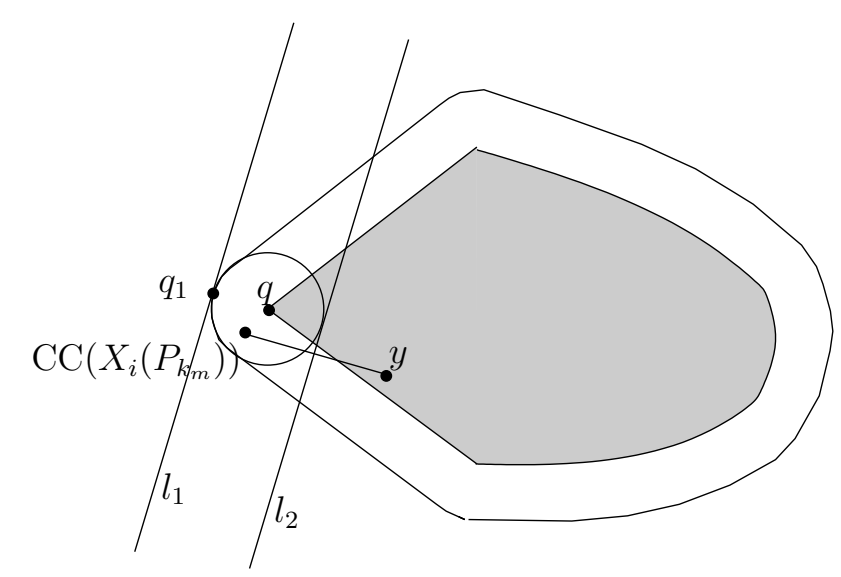

Fig. 27. Illustration of notions used in the proof of Lemma II.4. The shaded region represents $X_i(P)$. The outermost boundary represents the boundary of $\cup_{x \in X_i(P)} B(x, \delta)$. The small circle represents $B(q, \delta)$.

Let $x_1 = \arg\max_x\{\|x - \mathrm{CC}(X_i(P_{k_m}))\| \mid x \in X_i(P_{k_m})\}$ and let $x_2 = \arg\max_x\{\|x - y\| \mid x \in X_i(P_{k_m})\}$. By definition of the circumcenter and the fact that it is unique, $\|x_1 - \mathrm{CC}(X_i(P_{k_m}))\| < \|x_2 - y\|$. Also, by construction $\|x_1 - \mathrm{CC}(X_i(P_{k_m}))\| \geq \|x_2 - \mathrm{CC}(X_i(P_{k_m}))\|$. It follows then that $\|x_2 - y\| > \|x_2 - \mathrm{CC}(X_i(P_{k_m}))\|$. Therefore, $x_2$ belongs to the left of $l_2$. But as pointed earlier, $\|x_2 - y\| \to 0$ as $\delta \to 0$. But the circumradius of $X_i(P_{k_m})$ is equal to $\|x_1 - \mathrm{CC}(X_i(P_{k_m}))\|$ and $\|x_1 - \mathrm{CC}(X_i(P_{k_m}))\| < \|x_2 - y\|$. Since $\delta$ can be chosen arbitrarily small, it follows that $\|x_1 - \mathrm{CC}(X_i(P_{k_m}))\| \to 0$ as $m \to \infty$. This proves that if $q \in \mathrm{Ve}(\mathrm{rco}(P, Q_\epsilon))$, then $\mathrm{diam}(X_i(P_{k_m})) \to 0$.

Now let $p_i^{k_m}$ be the $i$th component of $P_{k_m}$. Since $p_i^{k_m}$, $\mathrm{CC}(X_i(P_{k_m})) \in X_i(P_{k_m})$ and $\mathrm{diam}(X_i(P_{k_m})) \to 0$, it follows that $\|p_i^{k_m} - \mathrm{CC}(X_i(P_{k_m}))\| \to 0$. But $P_{k_m} \to P$, and therefore, $p_i^{k_m} \to p_i$. This implies that $\mathrm{CC}(X_i(P_{k_m})) \to p_i$. Thus $p_i = q$ and $p_i \in \mathrm{Ve}(\mathrm{rco}(P, Q_\epsilon))$. We now claim that if $\mathrm{diam}(X_i(P)) > 0$ and $p_i \in \mathrm{Ve}(\mathrm{rco}(P, Q_\epsilon))$, then there exists a neighborhood of $P$ in $Y_{G_1,G_2}$ such that for all points $P'$ in the neighborhood, we have that $\mathrm{diam}(X_i(P')) \geq d$ for some $d > 0$. But this is a contradiction since we have shown that $\mathrm{diam}(X_i(P_{k_m})) \to 0$ for some sequence $P_{k_m} \to P$.

To prove the above claim, note that since $\mathcal{V}(p_i,\epsilon) \subseteq \mathrm{rco}(\mathcal{N}_{i,G_1},\mathcal{V}(p_i,Q_\epsilon))$ and $\mathrm{rco}(\mathcal{N}_{i,G_1},\mathcal{V}(p_i,Q_\epsilon)) = \mathrm{rco}(\mathcal{N}_{i,G_1},\mathcal{V}(p_i,Q_\epsilon))$, we can write

$$X_i = \mathrm{rco}(\mathcal{N}_{i,G_1},Q_\epsilon) \cap \left(\bigcap_{p_j\in\mathcal{N}_{i,G_2}} B(\frac{p_i+p_j}{2},\frac{r}{2})\right) \cap \left(\bigcap_{p_j\in\mathcal{N}_{i,G_2}} \bigcap_{k=1}^{n_{ij}} H_{Q_\epsilon}(v_{i,j,k})\right). \quad (9)$$

Now, let us first consider the case when $X_i(P)$ has non-empty interior. Then one can choose a segment $[p',p'']$ in the interior of $X_i(P)$ whose length is arbitrarily close to the length of $\mathrm{diam}(X_i(P))$. Therefore, $[p',p'']$ belongs in the interior of $\mathrm{rco}(\mathcal{N}_{i,G_1},Q_\epsilon)$, $B(\frac{p_i+p_j}{2},\frac{r}{2})$ and $H_{Q_\epsilon}(v_{i,j,k})$ for all $i$, $j$ and $k$. As shown earlier in the proof of statement (ii), the map $p_i \to \mathrm{rco}(\mathcal{N}_{i,G_1}(P),Q_\epsilon)$ is continuous for every $i$. Therefore, there exists a neighborhood $\mathcal{B}_1$ of $Y_{G_1.G_2}$ around $P$ such that for all $P'\in\mathcal{B}_1$, the segment $[p',p'']\subseteq \mathrm{rco}(\mathcal{N}_{i,G_1}(P'),Q_\epsilon)$.

From the proof of Proposition IV.2 (iv), we know that the map $(p_i,p_j) \to B(\frac{p_i+p_j}{2},\frac{r}{2})$ is continuous over $J$. It follows then that for fixed $i$ and $j$, there exists a neighborhood of $J$ around $(p_i,p_j)$ such that for all $(p_i',p_j')$ in the neighborhood, $[p',p'']\subseteq B(\frac{p_i'+p_j'}{2},\frac{r}{2})$. Since $G_2$ is fixed the set $\{j \mid j\in\mathcal{N}_{i,G_2}\}$ is fixed. Thus, there exists a neighborhood, $\mathcal{B}_2$ of $Y_{G_1,G_2}$ of $P$ such that for all $P'\in\mathcal{B}_2$, we have $[p',p'']\subseteq \left(\bigcap_{p_j\in\mathcal{N}_{i,G_2}} B(\frac{p_i+p_j}{2},\frac{r}{2})\right)$. Also, from the proof of Proposition IV.2 (iv), for fixed $j$ and a fixed strict concavity $a$, the map $(p_i,p_j)\to H_{Q_\epsilon}(v(p_i,p_j))$ is continuous over $J$, where $v(p_i,p_j)$ is the point on the strict concavity $a$ nearest to the segment $[p_i,p_j]$. Again, from the proof of Proposition IV.2 (iv), there exists a neighborhood of $J$ around $(p_i,p_j)$ such that for all points $(p_i',p_j')$ in the neighborhood, we have $a_{il}'\in\{a_{i_1},\cdots,a_{i_{n_{ij}}}\}$ for any $l\in\{1,\ldots,n_{ij}'\}$ where the notation is as in the proof of Proposition IV.2 (iv). Let $a_{il}' = a_{i_s}$. Since $[p',p'']$ belongs to the interior of $H_{Q_\epsilon}(v_{i,j,s}(p_i,p_j))$, which in turn varies continuously as $(p_i,p_j)$ for fixed $a_{i_s}$, there exists a neighborhood around $(p_i,p_j)$ such that for all $(p_i',p_j')$, we have $[p',p'']\subset H_{Q_\epsilon}(v_{i,j,l})(p_i',p_j')$. It then follows that there exists a neighborhood $\mathcal{B}_3$ of $Y_{G_1,G_2}$ around $P$ such that for all $P'=(p_1',\cdots,p_n')\in\mathcal{B}_3$, $[p',p'']\subset \left(\bigcap_{p_j\in\mathcal{N}_{i,G_2}(P')}\bigcap_{k=1}^{n_{ij}'} H_{Q_\epsilon}(v_{i,j,k}(p_i',p_j'))\right)$. Thus, for any $P'\in\mathcal{B}_1\cap\mathcal{B}_2\cap\mathcal{B}_3$, we have that $[p',p'']\subseteq X_i(P')$ or that $\mathrm{diam}(X_i(P'))\geq d$ where $d$ is equal to the length of the segment $[p',p'']$.

Second, let $X_i(P)$ have empty interior. Therefore, $X_i(P)$ is a line segment and since $p_i\in\mathrm{Ve}(\mathrm{rco}(P,Q_\epsilon))$ we have that $p_i$ is one of the end points of $X_i(P)$. Also, since $X_i(P)$ is a line segment, it follows from equation (9) that $X_i(P)$ must belong to the intersection of the boundaries of at least two of the sets $\mathrm{rco}(\mathcal{N}_{i,G_1},Q_\epsilon)$ and the half-planes $H_{Q_\epsilon}(v_{i,j,k})$. The set $X_i(P)$ cannot belong to the boundary of a ball $B(\frac{p_i+p_j}{2},\frac{r}{2})$ since its boundary is curved. Therefore, $X_i(P)$ belongs to the intersection of the boundaries of either (i) $\mathrm{rco}(\mathcal{N}_{i,G_1},Q_\epsilon)$ and a half-plane $H_{Q_\epsilon}(v_{i,j,k})$ or (ii) two half-planes $H_{Q_\epsilon}(v_{i,j,k})$ and $H_{Q_\epsilon}(v_{i,l,m})$. Let us assume without loss of generality that $X_i(P)$ belongs to the interior of the remaining sets. Then, as shown earlier, there exists a neighborhood, $\mathcal{B}_4$, of $Y_{G_1,G_2}$ around $P$ such that for all $P'$ in the neighborhood, $X_i(P)$ belongs to the interior of the remaining sets. Now cases (i) and (ii) correspond to Figure 28 top and bottom respectively. It follows immediately from the figure that we can choose a neighborhood, $\mathcal{B}_5$, of $P$ in $Y_{G_1,G_2}$ such that for $P'$ in the neighborhood, we have that the $\mathrm{rco}(\mathcal{N}_{i,G_1},Q_\epsilon)\cap H_{Q_\epsilon}(v_{i,j,k})$ and $H_{Q_\epsilon}(v_{i,j,k})\cap H_{Q_\epsilon}(v_{i,l,m})$ contain a segment of length arbitrarily close to the length of $X_i(P)$. Therefore, for all $P'\in\mathcal{B}_4\cap\mathcal{B}_5$, we have that $X_i(P')$ contains a segment of length $d$ where $d>0$.

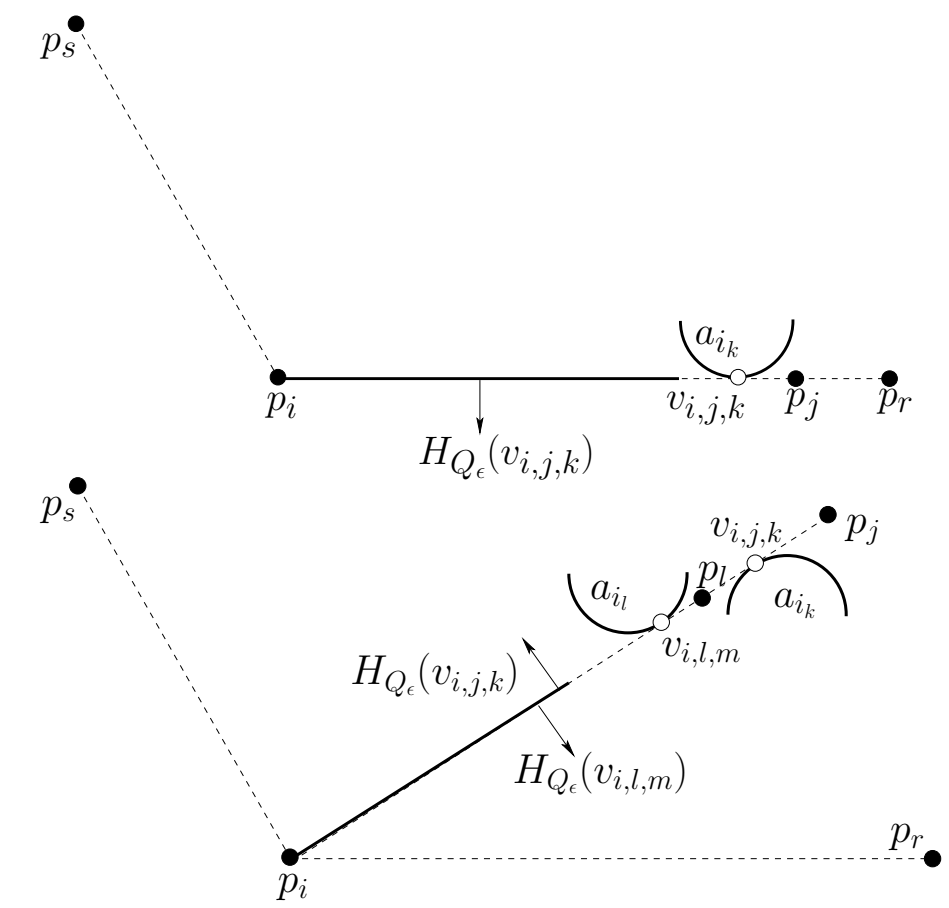

Fig. 28. Illustration of cases (i) and (ii) in the proof of Lemma II.4. The curved arcs represent strict concavities. The solid line segment represents $X_i(P)$. The dashed lines represent the boundary of $\mathrm{rco}(P,Q_\epsilon)$. Note that there must exist robots $p_r, p_s\in\mathcal{N}_{i,G_1}$ for the dashed lines to be the boundary of $\mathrm{rco}(P,Q_\epsilon)$. Note that $p_i\in\mathrm{Ve}(RCH(P,Q_\epsilon))$ and also $p_i$ is an end point of the segment $X_i(P)$. In the top figure $v_{i,j,k}$, denoted by the white disc, is the point on the strict concavity $a_{i_k}$ nearest to the segment $[p_i,p_j]$. Here $p_j\in\mathcal{N}_{i,G_2}$. The half-plane $H_{Q_\epsilon}(v_{i,j,k})$ has interior in the direction of the arrow. In the bottom figure, $v_{i,l,m}$ and $v_{i,j,k}$ are points on $a_{i_l}$ and $a_{i_k}$ nearest to the segments $[p_i,p_l]$ and $[p_i,p_j]$ respectively. The direction of the arrows points towards the interior of the half-planes $H_{Q_\epsilon}(v_{i,l,m})$ and $H_{Q_\epsilon}(v_{i,j,k})$. Here $p_j$, $p_l\in\mathcal{N}_{i,G_2}$.

This completes the proof of the fact that if $q\in\mathrm{Ve}(\mathrm{rco}(P,Q_\epsilon))$, then $\mathrm{diam}(X_i(P))=0$. But then this implies that $q=p_i$. Thus, this completes the proof of statement (ii).

To prove statement (iii), note that in proving statement (ii) we have shown that if $\mathrm{diam}(X_i(P))>0$, then $q=\lim_{k_m\to\infty}\mathrm{CC}(X_i(P_{k_m}))\notin\mathrm{Ve}(\mathrm{rco}(P,Q_\epsilon))$. From Lemma II.3, there exists at least one $p_i\in\mathrm{Ve}(\mathrm{rco}(P,Q_\epsilon))$ such that $\mathrm{diam}(X_i(P))>0$. Statement (iii) now follows directly.

We now prove statement (iv). It follows from Statement (ii), that $T^c_{G_1,G_2}(P)\subseteq\mathrm{rco}(P,Q_\epsilon)$. This implies that $\mathrm{rco}(T^c_{G_1,G_2}(P),Q_\epsilon)\subseteq\mathrm{rco}(P,Q_\epsilon)$. ■

From Lemma II.4(i), we know that, in particular, if the graphs $\mathcal{G}_{\text{sens}},\mathcal{G}_{\text{constr}}$ have fixed topology, then the closed perimeter minimizing algorithm is a closed map. However, during the evolution of the system we expect the graphs $\mathcal{G}_{\text{sens}},\mathcal{G}_{\text{constr}}$ to switch. To study the convergence properties of $T_{\mathcal{G}_{\text{sens}},\mathcal{G}_{\text{constr}}}$ over such a dynamic topology, we define a set-valued algorithm which essentially embeds all possible evolutions that can happen due to switching. Given any allowable $Q,Q_\epsilon$, define the *perimeter minimizing algorithm over all* *allowable graphs* to be the set-valued function $T$ that maps points in $Q_\epsilon^n$ to all possible subsets of $Q_\epsilon$ by:

$$\begin{aligned} T(P) = \{P' \in T^c_{G_1,G_2}(P) \mid G_2 \subseteq G_1 \subseteq \mathcal{G}_{\text{sens}}(P), \\ G_1, G_2 \text{ have fixed topologies} \\ \text{and same connected components as } \mathcal{G}_{\text{sens}}\}. \end{aligned}$$

Also, let $\mathcal{Y} = \{P \in Q_\epsilon^n \mid \mathcal{G}_{\text{sens}}(P) \text{ is connected}\}$ denote the set of all configurations where $\mathcal{G}_{\text{sens}}$ is connected.

**Proposition II.5 (Perimeter minimizing algorithm over all allowable graphs)** *The set $\mathcal{Y}$ and set-valued map $T$ have the following properties:*

(i) *$\mathcal{Y}$ is compact, $T(\mathcal{Y}) \subseteq \mathcal{Y}$, and $T$ is closed on $\mathcal{Y}$;*
(ii) *$\operatorname{rco}(T(P), Q_\epsilon) \subseteq \operatorname{rco}(P, Q_\epsilon)$, for $P \in Q_\epsilon^n$; and*
(iii) *if $P' \in T(P)$ and $P \in Q_\epsilon^n$, then $V_{\text{perim},Q_\epsilon}(P') \leq V_{\text{perim},Q_\epsilon}(P)$.*

*Proof:* Since $\mathcal{Y} \subseteq Q_\epsilon^n$, we have that $\mathcal{Y}$ is bounded. Now, let $\{P_k\}$ be a sequence in $\mathcal{Y}$ such that $P_k \to P$. Then, from the proof of Lemma II.4 (i), we know that there exists $k_0$ such that for all $k \geq k_0$, $(p_i^k, p_j^k)$ is an edge of $\mathcal{G}_{\text{sens}}(P_k)$ implies that $(p_i, p_j)$ is an edge of $\mathcal{G}_{\text{sens}}(P)$. But $\mathcal{G}_{\text{sens}}(P_k)$ is connected. Therefore, $\mathcal{G}_{\text{sens}}(P)$ is also connected. Hence, $P \in \mathcal{Y}$ and, therefore, $\mathcal{Y}$ is closed. This proves that $\mathcal{Y}$ is compact.

Now let $P' \in T(P)$ where $P \in \mathcal{Y}$. Then $P' \in T^c_{G_1,G_2}(P)$ for some $G_1$ and $G_2$ with fixed topologies having the same connected components as $\mathcal{G}_{\text{sens}}(P)$. Since $P \in \mathcal{Y}$, $\mathcal{G}_{\text{sens}}(P)$ is connected and hence, so are $G_1, G_2$. The fact that $P' \in \mathcal{Y}$ or that $\mathcal{G}_{\text{sens}}(P')$ has one connected component can be verified exactly along the lines of the proof of Theorem V.4 (i). This proves that $T(\mathcal{Y}) \subseteq \mathcal{Y}$.

Now, let $\{\overline{P}_k\}$ be any sequence with $\overline{P}_k \in T(P_k)$ such that $\overline{P}_k \to \overline{P}$. Since $\overline{P}_k \to \overline{P}$, and the number of distinct graphs possible over $n$ nodes is finite, there exists a subsequence $\overline{P}_{k_m} \to \overline{P}$ such that $\overline{P}_{k_m} \in T^c_{G_1,G_2}(P_{k_m})$ where $G_1(P_{k_m})$ and $G_2(P_{k_m})$ have fixed topologies for all $m$. Now, since $P_{k_m} \to P$, $\overline{P}_{k_m} \to \overline{P}$ and $T^c_{G_1,G_2}$ is closed over $Y_{G_1,G_2}$ from Lemma II.4 (i), we have that $\overline{P} \in T^c_{G_1,G_2}(P)$. Now, to finish the proof, we have to show that $G_2(P) \subseteq G_1(P) \subseteq \mathcal{G}_{\text{sens}}(P)$. The first inclusion follows from the fact the by definition of $T(P)$ where $G_2(P) \subseteq G_1(P)$. Now, notice that we have shown in proof of Lemma II.4 (i) that there exists $k_0$ such that for all $k \geq k_0$, $(p_i^k, p_j^k)$ being an edge of $\mathcal{G}_{\text{sens}}(P_k)$ implies that $(p_i, p_j)$ is an edge of $\mathcal{G}_{\text{sens}}(P)$. Also, by definition, $G_1(P_{k_m}) \subseteq \mathcal{G}_{\text{sens}}(P_{k_m})$. Therefore, for all $k_m \geq k_0$, we have that $(p_i^{k_m}, p_j^{k_m})$ being an edge of $G_1(P_{k_m})$ implies that $(p_i, p_j)$ is an edge of $\mathcal{G}_{\text{sens}}(P)$. Therefore, $G_1(P) \subseteq \mathcal{G}_{\text{sens}}(P)$. This completes the proof of statement (i).

Statement (ii) follows directly from Lemma II.4 (ii). Finally, statement (iii) is a consequence of statement (ii) and Lemma II.1 (ii). ∎

We are now ready to present the proof of the main result in the paper. The proof uses the analysis results presented in this section and the discrete time LaSalle Invariance Principle for set-valued maps [7].

*Proof of Theorem V.4:*

We begin by proving statement (i). Let $p_i[t_0], p_j[t_0]$ be two robots belonging to the same connected component of $\mathcal{G}_{\text{sens}}(P[t_0])$. Then from Corollary IV.6(ii), they must belong to the same connected component of $\mathcal{G}_{\text{constr}}(P[t_0])$. Therefore, there exists a path between $p_i[t_0]$ and $p_j[t_0]$ in $\mathcal{G}_{\text{constr}}(P[t_0])$. Let $(p_k[t_0], p_l[t_0])$ be any edge of $\mathcal{G}_{\text{constr}}(P[t_0])$ that belongs to this path. Note that $u_k[t_0]$ and $u_l[t_0]$ in the `Perimeter Minimizing Algorithm` algorithm are constrained to belong to the sets $C_{p_k,Q}(\mathcal{N}_{j,\mathcal{G}_{\text{constr}}}) - p_k[t_0]$ and $C_{p_l,Q}(\mathcal{N}_{j,\mathcal{G}_{\text{constr}}}) - p_l[t_0]$, respectively. Therefore, $p_k[t_0 + 1], p_l[t_0 + 1] \in \mathcal{C}_Q(p_k[t_0], p_l[t_0]) \subseteq Q_\epsilon \cap B(\frac{p_k[t_0]+p_l[t_0]}{2}, \frac{r}{2})$. Hence, the edge $(p_j[t_0 + 1], p_k[t_0 + 1])$ is an edge of $\mathcal{G}_{\text{sens}}(P[t_0 + 1])$. Thus, there exists a path between $p_i[t_0 + 1]$ and $p_j[t_0 + 1]$ in $\mathcal{G}_{\text{sens}}(P[t_0 + 1])$, which proves statement (i).

For statements (ii) and (iii), it suffices to prove the results for the system evolving under the action of $T$, since the evolution under $T_{\mathcal{G}_{\text{sens}},\mathcal{G}_{\text{constr}}}$ is just one of the possible evolutions under $T$. Statement (ii), therefore, follows from Proposition II.5(iii).

We now prove statement (iii). From statement (i), it follows that the number of connected components of $\mathcal{G}_{\text{sens}}(P[t])$ is nondecreasing. Since the number of possible connected components of $\mathcal{G}_{\text{sens}}$ is finite, this implies that, after a finite time $\bar{t}$, the number of connected components of $\mathcal{G}_{\text{sens}}$ does not change and robots belonging to different connected components never detect each other's presence. To prove statement (iii), it now suffices to show that any two robots belonging to the same connected component of $\mathcal{G}_{\text{sens}}$ converge to the same point. Since the evolution of any group of connected robots from time $\bar{t}$ on is independent from the existence of other connected components, we can then assume, without loss of generality, that $\mathcal{G}_{\text{sens}}(P[\bar{t}])$ has just one connected component.

According to Lemma V.3 and Proposition II.5(iii), $V_{\text{perim},Q_\epsilon}$ is continuous and non-increasing along $T$ on $Q_\epsilon^n$. From Proposition II.5(i), $T$ is a closed map on the compact set $\mathcal{Y} \subset Q_\epsilon^n$. Then, by the LaSalle Invariance Principle for closed set-valued maps [7], $P[t] \to \mathcal{M}$, where $\mathcal{M}$ is the largest weakly positively invariant[7] set contained in

$$\{P \in \mathcal{Y} \mid \exists P' \in T(P) \text{ s.t. } V_{\text{perim},Q_\epsilon}(P') = V_{\text{perim},Q_\epsilon}(P)\}.$$

Let us define the set $\operatorname{diag}(Q_\epsilon^n) = \{(p, \dots, p) \in Q_\epsilon^n \mid p \in Q_\epsilon\} \subset \mathcal{Y}$, and show that $\mathcal{M} = \operatorname{diag}(Q_\epsilon^n)$. Clearly, $\operatorname{diag}(Q_\epsilon^n) \subseteq \mathcal{M}$. To prove the other inclusion, we reason by contradiction. Let $P \in \mathcal{M} \setminus \operatorname{diag}(Q_\epsilon^n)$. Then, Lemma II.4 (iii) implies that for any connected $G_2(P)$, $G_1(P)$ satisfying $G_2(P) \subseteq G_1(P) \subseteq \mathcal{G}_{\text{sens}}(P)$, there exists $p_i \in \operatorname{Ve}(\operatorname{rco}(P, Q_\epsilon))$ such that $T^c_{\mathcal{G}_1,\mathcal{G}_2}(P)_i \subseteq \operatorname{rco}(P, Q_\epsilon) \setminus \operatorname{Ve}(\operatorname{rco}(P, Q_\epsilon))$. Therefore, if $P' \in T(P)$, then $P'$ contains a number of robots $N(P')$ belonging to $\operatorname{Ve}(\operatorname{rco}(P, Q_\epsilon))$ that is strictly smaller than $N(P)$. In turn, this implies that after at most $N(P)$ steps, at least one vertex of $\operatorname{rco}(P, Q_\epsilon)$ will contain any robot. By the definition of a vertex of $\operatorname{rco}(P, Q_\epsilon)$, it follows that the relative convex hull of the configuration after at most $N(P)$ steps is a strict subset of $\operatorname{rco}(P, Q_\epsilon)$. From Lemma II.1(ii) in Appendix II, $V_{\text{perim},Q_\epsilon}$ has strictly decreased. This is a contradiction with the fact that $\mathcal{M}$ is weakly positively invariant.

[7] A set $\mathcal{M}$ is weakly positively invariant with respect to a map $T$ on $W$ if, for any $w_0 \in W$, there exists $w \in T(w_0)$ such that $w \in W$.

To finish the proof, we need to show that every trajectory converges to an individual point, i.e., $P[t] \to P^* \in \mathcal{M}$ for all initial conditions $P[0] \in \mathcal{Y}$. Note that $\{P[t]\}$ is a sequence in a compact set. Hence, it has a convergent subsequence $\{P[t_k]\}$, say to $P^*$. Since $P[t] \to \mathcal{M}$, it follows that $P^* \in \mathcal{M}$. Using again Lemma V.3, we know that $V_{\text{perim},Q_\epsilon}$ is continuous and $V_{\text{perim},Q_\epsilon}(P^*) = 0$. Hence, $V_{\text{perim},Q_\epsilon}(P[t_k]) \to V_{\text{perim},Q_\epsilon}(P^*) = 0$. This implies that, given any $\delta > 0$, there exists $k_0$ such that for all $k \geq k_0$, $|V_{\text{perim},Q_\epsilon}(P[t_k]) - V_{\text{perim},Q_\epsilon}(P^*)| = V_{\text{perim},Q_\epsilon}(P[t_k]) < \delta$. From Proposition II.5(iii), we know that $V_{\text{perim},Q_\epsilon}(P[t]) \leq V_{\text{perim},Q_\epsilon}(P[t_{k_0}])$ for all $t > t_{k_0}$. In turn this implies that $V_{\text{perim},Q_\epsilon}(P[t]) \leq V_{\text{perim},Q_\epsilon}(P[t_{k_0}]) < \delta$ for all $t > t_{k_0}$. However, $\|P[t] - P^*\| \leq n \max\{\|p_i[t] - p_i^*\| \mid i \in \{1,\dots,n\}\} \leq nV_{\text{perim},Q_\epsilon}(P[t])$, where the second inequality holds because $p_i^* \in \operatorname{rco}(P[t], Q_\epsilon)$, for $i \in \{1,\dots,n\}$, and because the length of the shortest measurable curve enclosing a set of points is greater than the distance between any two of the points. Thus, for all $t \geq t_{k_0}$, we have $\|P[t] - P^*\| \leq n\delta$. Hence $P[t] \to P^* \in \mathcal{M}$. ■

## Appendix III
## Proofs of results in Section VI

*Proof of Proposition VI.1:* We begin by proving statement (i). Note that step `1:` of the `Constraint Set Generator Algorithm` is in fact the sensing operation that takes $O(M)$ time and the result is a star-shaped polygon with at most $M$ vertices. It is easy to see that the intersection of a star-shaped polygon with a half-plane takes time linear in the number of vertices of the polygon. Therefore, step `4:` of the algorithm takes at most $M$ time. If $Q$ has at most $\kappa$ concavities, then the computational complexity of `Constraint Set Generator Algorithm` is $O(M)+\kappa M$, which is $O(\kappa M)$.

We now prove statement (ii). Step `1:` of the `Perimeter Minimizing Algorithm` is the result of the sensing operation taking $O(M)$ time. Since the sensor takes at most $M$ readings, we can assume that the cardinality of $\mathcal{N}_{i,\mathcal{G}_{\text{sens}}}$ is at most $M$. In step `2:`, $\mathcal{N}_{i,\mathcal{G}_{\text{sens}}}$ is the set of all robot positions obtained in step `1:`. Let the computation of $\mathcal{N}_{i,\mathcal{G}_{\text{constr}}}$ given the set $\mathcal{N}_{i,\mathcal{G}_{\text{sens}}}$ take $\tau(M)$ time. Here $\tau(M)$ depends on the specific heuristic used to compute the maximal cliques of a graph. In step `3:`, $C_{p_i,Q}(\mathcal{N}_{i,\mathcal{G}_{\text{constr}}}) = \bigcap_{p_j \in \mathcal{N}_{i,\mathcal{G}_{\text{lc},\,\mathcal{G}}}}$ is the intersection of at most $M$ convex constraint sets. It is clear from the proof of statement (i) that each of the convex sets $\mathcal{C}_Q(p_i,p_j)$ has at most $M$ vertices and can be computed in $O(\kappa M)$ time. The intersection of $M$ convex polygons of $M$ vertices takes $O(M^2 \log M)$ time [19]. Thus, the computation of $C_{p_i,Q}(\mathcal{N}_{i,\mathcal{G}_{\text{constr}}})$ takes $MO(\kappa M) + O(M^2 \log M)$ time and $C_{p_i,Q}(\mathcal{N}_{i,\mathcal{G}_{\text{constr}}})$ has at most $M^2$ vertices. The computation of $\operatorname{rco}(\mathcal{N}_{i,\mathcal{G}_{\text{sens}}}, \mathcal{V}(p_i,\epsilon))$ takes $O(M \log M)$ time (cf. [20]), since the cardinality of both $\mathcal{N}_{i,\mathcal{G}_{\text{sens}}}$ and $\operatorname{Ve}(\mathcal{V}(p_i,\epsilon))$ is at most $M$. Now, finally step `3:` involves the intersection of two star-shaped polygons $C_{p_i,Q}(\mathcal{N}_{i,\mathcal{G}_{\text{constr}}})$ and $\operatorname{rco}(\mathcal{N}_{i,\mathcal{G}_{\text{sens}}}, \mathcal{V}(p_i,\epsilon))$ having at most $M^2$ and $M$ vertices respectively. Therefore the total number of vertices of the two polygons is at most $M^2 + M$. The computational complexity of computing the intersection of the two star-shaped polygons is $O((M^2 + M)\log(M^2+M) + M^3)$ or $O(M^3)$, where $M^3$ is due to the fact that the maximum number of possible intersection points is $M^3$ [21]. Thus the computational complexity of step `3:` is $MO(\kappa M) + O(M^2 \log M) + O(M) + O((M^2+M)\log(M^2 + M) + M^3)$. Since $\kappa \in O(M)$, we have that the computational complexity of step `3:` is $O(M^3)$. In step `4:`, the computation of the circumcenter of $X_i$, which is a convex polygon of at most $M^3$ vertices, takes $O(M^3 \log M^3)$ or $O(M^3 \log M)$ time [19]. Step `5:` can be performed in constant time. Thus, the complexity of the entire algorithm is $O(M) + \tau(M) + O(M^3) + O(M^3 \log M)$ or $\tau(M) + O(M^3 \log M)$.

To prove statement (iii), note that if $\mathcal{G} = \mathcal{G}_{\text{constr}} = \mathcal{G}_{\text{sens}}$, then $\tau(M) \in O(1)$. Also, from Lemma II.2 (ii), we have that $X_i = C_{p_i,Q}(\mathcal{N}_{i,\mathcal{G}}) \cap \operatorname{co}(\mathcal{N}_{i,\mathcal{G}})$. The convex hull of $M$ points can be computed in $O(M \log M)$ time (cf. [19]) and hence $\operatorname{co}(\mathcal{N}_{i,\mathcal{G}})$ can be computed in $O(M \log M)$ time. The set $X_i$ is an intersection of convex sets and can be computed in $O(M^2 + M)$ or $O(M^2)$ time, and contains at most $O(M^2)$ vertices. Step `4:` now takes $O(M^2 \log M^2)$ or $O(M^2 \log M)$ time. Thus, the complexity of the entire algorithm is $O(M^2 \log M)$. ■